\documentclass[final]{cvpr}

\usepackage{amsmath}
\usepackage{amssymb}
\usepackage{times}
\usepackage{mathtools}
\usepackage{bm}
\usepackage{bbm}
\usepackage{mmstyles}

\usepackage{enumitem}
\setlist[itemize]{topsep={0pt},partopsep={0pt}}

\usepackage{booktabs}
\usepackage{multirow}
\usepackage{rotating}
\usepackage{diagbox}

\usepackage{epsfig}
\usepackage{graphicx}
\usepackage[percent]{overpic}

\usepackage{caption}
\renewcommand{\arraystretch}{1.1}

\usepackage{xcolor}
\usepackage{colortbl}
\definecolor{remark}{rgb}{1,.5,0} 
\definecolor{citecolor}{rgb}{0,0.443,0.737} 
\definecolor{linkcolor}{rgb}{0.956,0.298,0.235} 
\definecolor{gray}{gray}{0.95}
\definecolor{cyan}{rgb}{0.831,0.901,0.945}

\usepackage[linesnumbered,ruled,vlined]{algorithm2e}
\SetAlCapNameFnt{\small}
\SetAlCapFnt{\small}

\usepackage[numbers,sort&compress]{natbib}
\usepackage[pagebackref=true,breaklinks=true,colorlinks=true,bookmarks=false,linkcolor=linkcolor,citecolor=citecolor]{hyperref}

\def\cvprPaperID{717} % *** Enter the CVPR Paper ID here
\def\confYear{CVPR 2021}

\begin{document}

%%%%%%%%% TITLE
% \title{\LaTeX\ Author Guidelines for CVPR Proceedings}
\title{Prototypical  Pseudo Label Denoising and Target Structure Learning for \\ Domain Adaptive Semantic Segmentation
}

\author{Pan Zhang$^1$
	\thanks{This work is done during the first author's internship at Microsoft Research Asia.}
	, Bo Zhang$^2$, Ting Zhang$^2$, Dong Chen$^2$, Yong Wang$^1$, Fang Wen$^2$
	\\ 
	$^1$University of Science and Technology of China \quad
	$^2$Microsoft Research Asia
}

\maketitle

%%%%%%%%% ABSTRACT
\begin{abstract}
    Self-training is a competitive approach in domain adaptive segmentation, which trains the network with the pseudo labels on the target domain. 
    %We claim that denoising the pseudo labels and enforcing compact target features significantly improve the self-training.
    However inevitably, the pseudo labels are noisy and the target features are dispersed due to the discrepancy between source and target domains.
    In this paper, we rely on representative prototypes, the feature centroids of classes, to address the two issues for unsupervised domain adaptation.
    In particular, we take one step further and exploit the feature distances from prototypes that provide richer information than mere prototypes.
    Specifically,
    we use it to estimate the likelihood of pseudo labels to facilitate online correction in the course of training. Meanwhile, we 
    align the prototypical assignments based on relative feature distances for two different views of the same target, producing a more compact target feature space.
    %We claim that denoising the pseudo labels and enforcing compact target features significantly improve the self-training. We compute prototypes, the feature centroids of classes, to achieve the goals. We estimate the likelihood of false pseudo labels according to the feature distance to prototypes, and base on this measure to online correct the pseudo labels in the course of the training. Meanwhile, we compute the prototypical assignment for target images, which is further used to guide the learning for a augmented view. In this way, the network can produce more compact target feature space, as the self-supervised Deepcluster work. 
    Moreover, we find that distilling the already learned knowledge to a self-supervised pretrained model further boosts the performance. Our method shows tremendous performance advantage over state-of-the-art methods. We will make the code publicly available.

%   Keywords: self-training; pseudo-label denoising; structure learning; self-supervised pretraining
\end{abstract}

%%%%%%%%% BODY TEXT
\section{Introduction}
Despite the remarkable success of deep learning in computer vision, attaining high performance requires vast quantities of data. It is usually expensive to obtain labels for dense prediction tasks, \eg, semantic segmentation.
%as labeling such pixel-wise annotations is prohibitively labor-intensive.
Therefore, people think of leveraging abundant photo-realistic synthetic images with freely generated labels~\cite{richter2016playing,ros2016synthia}. 
However, deep neural networks are notoriously sensitive to the domain misalignment that any nuanced unrealism in rendered images will induce poor generalization to real data. 
To address this issue, domain adaption techniques aim to transfer the knowledge learned from the synthetic images (source domain) to real ones (target domain) with minimal performance loss. In this work, we focus on the challenging case, \emph{unsupervised domain adaptation} (UDA), where there are no accessible labels in the target domain. Specifically, we solve the UDA problem for semantic segmentation.

Rather than explicitly aligning the distributions of the source and target domains as most predominant solutions~\cite{hong2018conditional,saito2018maximum,chang2019all,wan2020bringing,tsai2018learning}, self-training~\cite{zou2018unsupervised,li2019bidirectional,zou2019confidence,zhang2019category} has recently emerged as a simple yet competitive approach in the UDA task. This is achieved by iteratively generating a set of pseudo labels based on the most confident predictions on the target data and then relying on these pseudo labels to retrain the network. In this way, the network 
%becomes more confident to easier target samples and 
gradually learns the adaptation in the self-paced curriculum learning. However, the performance 
%of the domain adaptation using this method 
still lags far behind the supervised learning or semi-supervised learning using a few labeled samples, making unsupervised domain adaptation impractical in real scenarios.

After dissecting the self-training, we find two key ingredients are lacking in previous works. First, typical practice~\cite{zou2018unsupervised,zou2019confidence} suggests selecting the pseudo labels according to a strict confidence threshold, 
while high scores are not necessarily correct,
%the labels are inevitably noisy, 
making the network fail to learn reliable knowledge in the target domain. 
%Recent work~\cite{zheng2020rectifying} attempts to address this issue by estimating the uncertainty of pseudo labels, but the prediction variance the paper used cannot fully reflect the false predictions. 
%Hence, learning from noisy pseudo labels is under-explored in domain adaptation. 
Second, due to the domain gap, the network is prone to produce dispersed features in the target domain. 
It is likely that for target data, the closer to the source distribution, the higher the confidence score.
As a result, data lying far from the source distribution (\ie low scores)
%the selected pseudo labels, mostly close to the source domain, cannot fully cover the target feature space, so the feature points lying in the far end of the target distribution 
will never be considered during the training. 
%Therefore, learning compact feature structures in the target domain is of great importance to transfer the knowledge based on pseudo labels.

In this paper, we propose to online denoise the pseudo labels and learn a compact target structure to address the above two issues respectively. We resort to prototypes, \ie, the class-wise feature centroids, to accomplish the two tasks. 
(1) We rectify the pseudo labels by estimating the class-wise likelihoods according to its relative feature distances to all class prototypes. 
This depends on a practical assumption that the prototype lies closer to the true centroid of the underlying cluster,
implying that false pseudo labels are in the minority.
It is worth noting that the prototypes are computed on-the-fly, and thus the pseudo
labels are progressively corrected throughout the training.
(2) We draw inspiration from the Deepcluster~\cite{caron2018deep} to learn the intrinsic structure of the target domain.
Instead of directly learning from the cluster assignment, we propose to align soft prototypical assignments for different views of the same target, which produces a more compact target feature space. We refer to our method \emph{ProDA} as we heavily rely on prototypes for domain adaption. 
%On one hand, we assume the false pseudo labels are in the minority, so the prototypes computed on the predicted pseudo labels are less sensitive to those outliers and lie closer to the true centroids of the underlying clusters. Once the network is capable to induce well-separated target features, we can measure the likelihood of false labels according to the relative feature distance to different prototypes, so the labels can be rectified accordingly. Note that we compute the prototypes on-the-fly, and the pseudo labels can be progressively corrected throughout the training. The network is bootstrapped by learning from the denoised labels and is able to learn the transferred source domain knowledge far better than using the pre-computed psuedo labels. 

%On the other hand, we draw inspiration from the Deepcluster~\cite{caron2018deep} to learn the intrinsic structure of the target domain. The difference is that there is not need to learn the prototypes from scratch, but they can be initialized using the knowledge transferred from the source domain. Besides, instead of directly learning from the prototypical assignment, we propose to use this assignment to guide the learning for a strong augmented view. More compact feature space forms in the target domain, which eases the adaptation based on the pseudo labeling.

Supercharged with the above techniques, our ProDA can demonstrate clear superiority over prior works. Moreover, we find that the domain adaptation can also benefit from the task-agnostic pretraining ---  distilling the knowledge to a self-supervised model~\cite{he2020momentum,chen2020simple} further boosts the performance to a record high. Our contributions can be summarized as follows:
\begin{itemize}[leftmargin=*]
    \itemsep=-0.9mm
    \item We propose to online correct the soft pseudo labels according to the relative feature distances to the prototypes, whereas the prototypes are also updated on-the-fly. The network thereby learns from denoised pseudo labels throughout the training. 
    
    \item We propose to rely on the soft prototypical assignment to teach the learning of an augmented view so that a compact target feature space can be obtained.
    
    \item We show that distilling the already-learned knowledge to a self-supervised pretrained model further improves the performance significantly.
    
    \item The proposed ProDA substantially outperforms state-of-the-art. With the Deeplabv2~\cite{chen2017deeplab} network, our method achieves the Cityscapes~\cite{cordts2016cityscapes} segmentation mIOU by 57.5 and 55.5 when adapting from the GTA5~\cite{richter2016playing} and SYNTHIA~\cite{ros2016synthia} datasets, improving the adaption gain\footnote{The mIoU gain relative to the model without domain adaption.} by 52.6\% and 58.5\% respectively over the prior leading approach. 
\end{itemize}

\section{Related Work}
\begin{figure*}[t!]
    \small
    \centering
    \begin{overpic}
    [scale=0.6]{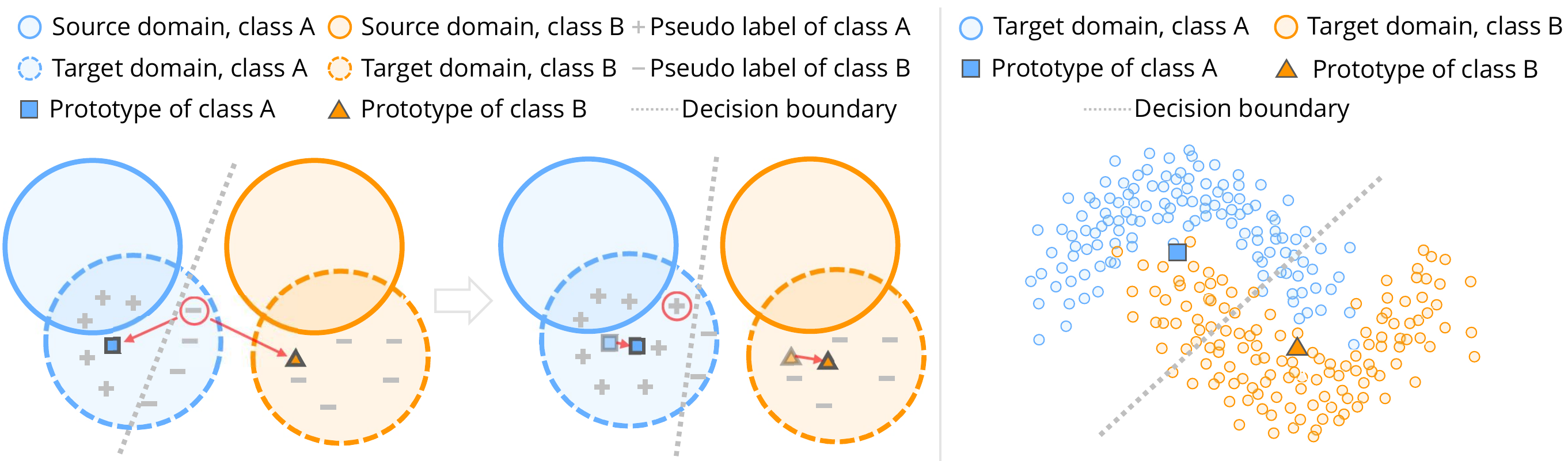}
    \put(28,-1.5){\footnotesize (a)}
    \put(80,-1.5){\footnotesize (b)}
    \end{overpic}
    \vspace{1.5em}
    \caption{\textbf{We illustrate the existing issues of self-training by visualizing the feature space.} (a) The decision boundary (dashed line)  crosses the distribution of the target data and induces incorrect pseudo label predictions. This is because the network is unaware of the target distribution when generating pseudo labels. To solve this, we calculate the prototypes of each class on-the-fly and rely on these prototypes to online correct the false pseudo labels. (b) The network may induce dispersed feature distribution in the target domain which is hardly differentiated by a linear classifier. } 
    \label{figure:diagram}
\end{figure*}

\noindent\textbf{Unsupervised domain adaptation.} As suggested by the theoretical analysis~\cite{ben2010theory}, domain alignment methods focus on reducing the distribution mismatch by optimizing some divergence~\cite{long2015learning,lee2019sliced} or adopting adversarial training~\cite{goodfellow2014generative,nowozin2016f} at either the image level~\cite{hoffman2018cycada,gong2019dlow,chen2019crdoco,sankaranarayanan2018learning,wu2019ace,abramov2020keep,choi2019self}, the intermediate feature level~\cite{hong2018conditional,saito2018maximum,chang2019all,wan2020bringing} or the output level~\cite{tsai2018learning}. However, aligning global distribution cannot guarantee a small expected error on the target domain~\cite{kumar2018co,chen2019progressive,zhao2019learning}.
Recent approaches~\cite{du2019ssf,luo2019taking,wang2020classes} attempt to align distribution in a class-wise manner,
%by incorporating the class information into discriminators, 
aiming to promote fine-grained feature alignment. 
In fact, it is unnecessary to rigorously align the distribution
%for a good domain adaptation 
as long as the features are well-separated. 
% \begin{equation}
%     \epsilon_t(h) \leq \epsilon_s(h) + \frac{1}{2}d_{\cH\Delta\cH}(\cD_s,\cD_t) + \lambda^*
% \end{equation}

On the other hand, techniques originated from semi-supervised learning (SSL) offer competitive performance.  Entropy minimization and its variants~\cite{saito2019semi,vu2019advent,chen2019domain} encourage the network to make sharp predictions on the unlabeled target data, and the resulting network is prone to be over-confident on false predictions. To address this, self-training~\cite{zou2018unsupervised} that leverages iteratively generated pseudo labels has been proposed. 
%which alternatively generates pseudo labels based on the most confident predictions on the target data and then uses these pseudo labels for retraining.
However, the pseudo labels are inevitably noisy.
Hence, \cite{zou2019confidence} adds confidence regularization terms to the network,
%Realizing that the pseudo labels are quite noisy, ~\cite{zou2019confidence} regularizes the network not to be over-confident 
while~\cite{zheng2020rectifying} explicitly estimates a prediction confidence map to reduce the side-effect of unreliable labels. In~\cite{li2019bidirectional}, self-training and image translation are found mutually beneficial. 
A recent work~\cite{zhang2019category} generates pseudo labels based on categorical centroids 
%\ie, the prototypes in this paper, 
and 
%the self-training on these pseudo labels 
enforces feature alignment in category level. 
However, these self-training approaches are optimized in an alternative manner --- labels are fixed over the course of representation learning, and only get updated after the entire training stage. In contrast, we propose an online pseudo label updating scheme where the false predictions are rectified according to the prototypical context estimated in the target domain.

\noindent\textbf{Unsupervised representation learning.} A surge of research interest has been recently attracted to unsupervised learning~\cite{qi2019small}. Early efforts dedicate to designing  pretext tasks~\cite{larsson2016learning,zhang2017split,gidaris2018unsupervised,donahue2019large}, which
%and the proposed pretext tasks 
are proven beneficial for UDA when utilized as auxiliary tasks on target data~\cite{sun2019unsupervised,xu2019self,saito2020universal}.
%however, their performance improvement is marginal. 
The gap with supervised learning is considerably closed by a few prominent works~\cite{he2020momentum,chen2020simple} that build on contrastive learning. 
A series of recent works~\cite{grill2020bootstrap,misra2020self,asano2019self,caron2020unsupervised} find that the network is able to learn rich semantic features as long as they are consistent under different augmented views.
Yet these methods assume image-wise discrimination~\cite{wu2018unsupervised}, making them unsuitable for learning pixel-level semantics for segmentation tasks. %Without the need to constitute negative pairs required by the contrastive learning, a series of recent works~\cite{grill2020bootstrap,misra2020self,asano2019self,caron2020unsupervised} find that the network is able to learn rich semantic features as long as they are consistent under different augmented views. 
%In practice, our approach is also related to 
%Deepcluster~\cite{caron2018deep}, which alternatively clusters the features and trains the network to predict the cluster assignment. 
In this work, we find the marriage of consistent learning and cluster-based representation learning fits remarkably well with the UDA problem and learn a compact target feature space inspired from Deepcluster~\cite{caron2018deep}.
Differently, we align relative feature distances rather than cluster assignments for different augmented views.
% The prototypes (cluster centroids) from the source serve as a good estimate for the target and can be used to predict the cluster assignment for another augmented view. 

\noindent\textbf{Learning from noisy labels.} Self-training even with careful thresholding still gives noisy pseudo labels. Therefore, this work is also motivated by emerging techniques~\cite{song2020learning} of learning from noisy labels. A straightforward way is to design robust losses~\cite{zhang2018generalized, wang2019symmetric}, but these methods fail to handle real-world noisy data. Self-label correction~\cite{zheng2019unsupervised,zheng2020rectifying,song2019selfie,wang2020proselflc} is a more appealing approach.
%which iteratively corrects the noisy labels. 
A typical manner~\cite{mendel2020semi,lee2018cleannet} under this category is to train two or multiple learners simultaneously and exploit their agreement of predictions to measure the label reliability.
%~\cite{mendel2020semi,lee2018cleannet} propose to train an auxiliary network to spot and amend the noisy labels. 
Our proposed pseudo label denoising is more close to~\cite{han2019deep} which online corrects the incorrect labels according to the prototypes
%In comparison to~\cite{han2019deep} that relies on 
determined by some complex heuristics.
%for determining the prototypes, 
In contrast, we are able to compute prototypes on-the-fly. 
On the other hand, knowledge distillation (KD)~\cite{hinton2015distilling,li2017learning,xie2020self,li2019learning} is proven effective to transfer clean knowledge from the teacher model to the student even when the network learns from itself~\cite{chen2020simple,li2019learning}. In this work, we demonstrate that the knowledge distillation to a self-supervised pretrained model further pushes the performance limit in our task.

\section{Preliminary}
% \begin{figure}[t!]
%     \small
%     \centering
%     \begin{overpic}
%     [scale=0.5]{figure/diagram.pdf}
%     \put(20,34){\footnotesize (a)}
%     \put(73,34){\footnotesize (b)}
%     \put(50,0){\footnotesize (c)}
%     \end{overpic}
%     \vspace{1em}
%     \caption{Illustration of the issues of pseudo labeling and our online pseudo label refinement scheme. (a) The decision boundary (dashed line) of the network crosses the target distributions as the network is unaware of the target distribution, thus making false pseudo predictions. (b) The network may induce dispersed feature distribution in the target domain which is hardly differentiated by linear classifier. (c) We propose to calculate the prototypes of each class on-the-fly, and these prototypes are used to online correct the false pseudo labels.} 
%     \label{figure:diagram}
% \end{figure}

Given the source dataset $\cX_s=\{x_s\}_{j=1}^{n_s}$ with segmentation labels $\cY_s=\{y_s\}_{j=1}^{n_s}$, we aim to train a segmentation network that learns the knowledge from the source and is expected to achieve low risk on the target dataset $\cX_t=\{x_t\}_{j=1}^{n_t}$ without accessing its ground truth labels $\cY_t$, where $\cY_s$ and $\cY_t$ share the same $K$ classes. Generally, the network $h=g \circ f$ can be regarded as a composite of a feature extractor $f$ and a classifier $g$. 

Typically, the networks trained on the source data, \ie, the source model, cannot generalize well to the target data due to the domain gap. To transfer the knowledge, traditional self-training techniques~\cite{zou2018unsupervised,zou2019confidence} optimize the categorical cross-entropy (CE) with pseudo labels $\hat{y}_t$:
% in the target domain:
\begin{equation}
    \ell_{ce}^t = -\sum_{i=1}^{H\times W} \sum_{k=1}^K \hat{y}^{(i,k)}_t \log(p^{(i,k)}_t),
\end{equation}
where $p_t = h(x_t)$ and $p^{(i,k)}_t$ represents the softmax probability of pixel $x_t^{(i)}$ belonging to the $k$th class. Typically, one can use the most probable class predicted by the source network as pseudo labels:
\begin{equation}
    \hat{y}^{(i,k)}_t =
    \begin{cases}
    1, & \text{if }\ k=\arg \max_{k'} p_t^{(i,k')} 
    \\
    0, & \text{otherwise}
    \end{cases}
\end{equation}
Here 
%the pseudo label for each pixel is a one-hot vector, and 
we denote this conversion from the soft predictions to hard labels by $\hat{y}_t = \xi(p_t)$. In practice, since the pseudo labels are noisy, only the pixels whose prediction confidence is higher than a given threshold are accounted for the pseudo label retraining. In this way, the network in the target domain is bootstrapped by learning from pseudo labels that only get update till convergence, and then the updated labels are employed for the next training stage.

% Nonetheless, there exist two issues in the self-training. First, pseudo labels even under strict confidence selection inevitably contain errors, and the noisy pseudo labels hurt the domain adaptive learning. Second, as shown in the toy example (Figure), pseudo labeling alone cannot produce decision boundaries that perform well on the target distribution even when the target features from the source model are well-separated. This is because the target distribution is only partially covered by the pseudo labels, and the training purely on them will ignore the target data lying on the remote side in the feature space. We hereby propose two techniques that rely on the prototypes to solve these issues respectively and we will elaborate on them in the next section.  

\section{Method}

\subsection{Prototypical pseudo label denoising}
We conjecture that updating the pseudo label
%To robustly transfer the source domain knowledge, we propose to correct the false predictions in the pseudo labels. In previous works, pseudo labels indeed evolve after the network has been trained on them till convergence, but we claim that the pseudo label update 
after one training stage is too late as the network has already overfitted the noisy labels. 
While simultaneously updating the pseudo labels and the network weights, on the other hand, is prone to give trivial solutions. 

In this work, we propose a simple yet effective approach to online update the pseudo labels while avoiding trivial solutions. The key is to fix the soft pseudo labels and progressively weight them by class-wise probabilities,
%correct them by weighting the class-wise probability,
%based on the ongoing learned features. In this way, the network can learn from the denoised labels that 
with the update in accordance with the freshly learned knowledge. Formally, we propose to use the  weighted pseudo labels for self-training:
\begin{equation}
    \hat{y}^{(i,k)}_t = \xi(\omega_t^{(i,k)} p^{(i,k)}_{t,0}),
    \label{eq:label_reweight}
\end{equation}
where $\omega_t^{(i,k)}$ is the weight we propose for modulating the probability and changes as the training proceeds. The $p^{(i,k)}_{t,0}$ is initialized by the source model and remains fixed throughout the learning process, thus serving as a boilerplate for the subsequent refinement.

We exploit the distances 
from the prototypes
to gradually rectify the pseudo labels.
%, in the target domain. 
Let $f(x_t)^{(i)}$ represent the feature of $x_t$ at index $i$. If it
is far from the prototype $\eta^{(k)}$, the feature centroids of class~$k$, it is more probable that
the learned feature
is an outlier, hence we downweight its probability of being classified into $k$th category.
%otherwise, we rely on the original pseudo label prediction. 
Concretely, we define the modulation weight in Equation~\ref{eq:label_reweight} as the softmax over feature distances to prototypes:
%in the feature space:
\begin{equation}
    \omega_t^{(i,k)} = \frac{\exp (-\|\tilde{f}(x_t)^{(i)}-\eta^{(k)}\| / \tau)}
    {\sum_{k'} \exp (-\|\tilde{f}(x_t)^{(i)}-\eta^{(k')}\| / \tau)},
    \label{eq:prototype_weight}
\end{equation}
where $\tilde{f}$ denotes the momentum encoder~\cite{he2020momentum} of the feature extractor $f$, as we desire a reliable feature estimation for $x_t$, and $\tau$ is the softmax temperature empirically set to $\tau=1$.
%Equation~\ref{eq:prototype_weight} 
In this equation, $\omega_t^{(i,k)}$ actually approximates the trust confidence of 
%labels with the probability of 
$x_t^{(i)}$ belonging to the $k$th class.
% defined by the prototypes. 
Note that this equation has a close formulation in~\cite{snell2017prototypical,qi2018low} that shows promise in few-shot learning.
Instead of relying on the prototypes for classifying new samples, we attempt to correct the mistakes in the pre-generated pseudo labels according to this prototypical context.

\noindent\textbf{Prototype computation.} The proposed label updating scheme requires to 
compute the prototypes on-the-fly. 
%At the beginning of the domain adaptation, 
At first, prototypes are initialized according to the predicted pseudo labels $\hat{y}_t$ for target domain images, which is
\begin{equation}
    \eta^{(k)} = \frac{\sum_{x_t\in\cX_t}\sum_i f(x_t)^{(i)} * \mathbbm{1}(\hat{y}^{(i,k)}_t == 1)}
    {\sum_{x_t\in \cX_t} \sum_i \mathbbm{1}(\hat{y}^{(i,k)}_t == 1)},
    \label{eq:prototype_init}
\end{equation}
where $\mathbbm{1}$ is the indicator function. However, such prototype calculation is computational-intensive during training. To address this, we estimate the prototypes as the moving average of the cluster centroids in mini-batches, so that we can track the prototypes that slowly move.
%during the training. 
Specifically, in each iteration, the prototype is estimated as,
%we estimate the prototypes for the whole target domain through
\begin{equation}
    \eta^{(k)} \gets \lambda \eta^{(k)} + (1-\lambda) \eta'^{(k)},
\end{equation}
where $\eta'^{(k)}$ is the mean feature of class $k$ calculated within the current training batch from the momentum encoder, and $\lambda$ is the momentum coefficient which we set to 0.9999.

\noindent\textbf{Pseudo label training loss.} With the online refined pseudo labels, we are able to train the network for target domain segmentation. Instead of using a standard cross-entropy, we adopt a more robust variant, symmetric cross-entropy (SCE) loss~\cite{wang2019symmetric}, to further enhance the noise-tolerance to stabilize the early training phase. Specifically, we enforce
\begin{equation}
    \ell^t_{sce} =  \alpha \ell_{ce}(p_t,\hat{y}_t) + \beta \ell_{ce}(\hat{y}_t, p_t),
\end{equation}
where $\alpha$ and $\beta$ are balancing coefficients and set to 0.1 and 1 respectively. Following~\cite{wang2019symmetric}, we clamp the one-hot label $\hat{y}_t$ to $[1\mathrm{e}{-4}, 1]$, so as to avoid the numerical issue of $\log 0$. 
%It is noteworthy that the symmetric cross-entropy alone does not take effect when using the raw pseudo labels for self-training, as shown in the ablation study. But we observe it helps the pseudo label denoising and we conjecture that a robust loss benefits the early training phase during which the label noises have not been thoroughly corrected using the initial prototypes. 

\noindent\textbf{Why are prototypes useful for label denoising?} 
First, 
the prototypes are less sensitive to the outliers that are assumed to be the minority.
Second, prototypes treat different classes equally regardless of their occurrence frequency, which is particularly useful to semantic segmentation as class imbalance poses a challenging issue.
Experiments show that the proposed label denoising considerably improves the performance for hard classes. 
More importantly, prototypes help to gradually rectify the incorrect pseudo labels, which we illustrate using a toy example. As shown in Figure~\ref{figure:diagram}(a), the classifier $g$, may still give a decision boundary crossing the target clusters and yields false pseudo labels. Training against them cannot further improve the classifier.
%The feature extractor $f_\theta$ by the source model produces clustered structures for the target images, whereas the classifier $g_\theta$, however, gives a decision boundary crossing the target features and yields incorrect pseudo labels. 
The prototypes, on the other hand, downweight 
%form another decision boundary defined by the nearest neighbor in the feature space, which helps to correct 
the false pseudo labels near the decision boundary of $g$ as they are far from the prototypes. 
In this way, the network improves, and in turn, makes the prototypes closer to the true cluster centroid. 
%In this way, the prototypical label denoising iteratively bootstraps the learning in the target domain.  

\subsection{Structure learning by enforcing consistency}
%In Figure~\ref{figure:diagram}(a), 
Pseudo labels can be denoised when the feature extractor $f$ generates compact target features.
However, due to the domain gap, the generated target distribution is more likely to be dispersed, as shown in Figure~\ref{figure:diagram}(b). In this case, the prototypes fail to rectify the labels of the data whose features lie in the far end of the cluster even when the target features from the source model are well-separated. A recent work~\cite{oliver2018realistic} has identified this issue in semi-supervised learning, but the issue becomes worse in domain adaptation since
a few pseudo labeled data cannot cover the entire distribution in the target domain.  

To this end, we aim to learn the underlying structure of target domain, and hope to obtain more compact features that are friendly to the pseudo label refinement. Motivated by the recent success of unsupervised learning, we perform simultaneously clustering and representation learning.
As opposed to learning against the prototypical assignment, we use the prototypical assignment under weak augmentation to guide the learning for the strong augmented view. Specifically, let $\cT(x_t)$ and $\cT'(x_t)$ respectively denote the weak and strong augmented views for $x_t$. We make use of the momentum encoder $\tilde{f}$ to generate a reliable prototypical assignment for $\cT(x_t)$ which is,
\begin{equation}
    z_{\cT}^{(i,k)} = \frac{\exp (-\|\tilde{f}(\cT(x_t))^{(i)}-\eta^{(k)}\| / \tau)}
    {\sum_{k'} \exp (-\|\tilde{f}(\cT(x_t))^{(i)}-\eta^{(k')}\| / \tau)},
    \label{eq:soft_label}
\end{equation}
where $\tau=1$ by default. Likewise, the soft assignment $z_{\cT'}$ for $\cT'({x_t})$ can be obtained in the same manner except that we use the original trainable feature extractor $f$. Since $z_{t}$ is more reliable as the feature is given by a momentum encoder $\tilde{f}$ and the input $x_t$ suffers from less distortion, we use it to teach $f$ to produce consistent assignments for $\cT(x_t)$. Hence, we minimize the Kullback–Leibler (KL) divergence between the prototypical assignments under two views:
\begin{equation}
    \ell_{kl}^t = \text{KL}\left(z_{\cT} \| z_{\cT'}\right). 
    % \\&= -\sum_{i=1}^{H\times W}\sum_{j=1}^{K} z_\cT^{(i,k)} \log z_\cT^{(i,k)} - z_\cT^{(i,k)} \log z_{\cT'}^{(i,k)}
    \label{eq:consistency}
\end{equation}
Intuitively, 
%augmentations introduce small perturbation in the feature space, and
this equation enforces the network to give consistent prototypical labeling for the adjacent feature points, which results in more compact feature space in the target domain.
%as the feature points with low prediction confidence can be gradually contracted towards the surrounding point with higher confidence.  

Similar to the previous works that simultaneously learn the clustering and representation, the proposed prototypical consistent learning may suffer from degeneration issue, \ie, one cluster becomes empty. To amend this, we use a regularization term from~\cite{zou2019confidence}, which encourages the  output to be evenly distributed to different classes, 
\begin{equation}
    \ell_{reg}^t = -\sum_{i=1}^{H\times W}\sum_{j=1}^{K} \log p_t^{(i,k)}.
\end{equation}
%This equation essentially enforces the KL divergence of the uniform distribution and the softmax output.

Now equipped with the online label correction and the prototypical consistent learning, we train the domain adaptation network with the following total loss:
\begin{equation}
    \ell_{total} = \ell_{ce}^s + \ell_{sce}^t + \gamma_1\ell_{kl}^t + \gamma_2\ell_{reg}^t.
    \label{eq:total_loss}
\end{equation}
By default, the weighting coefficients $\gamma_1$ and $\gamma_2$ are set to 10, 0.1 respectively. 

\subsection{Distillation to self-supervised model}
%Self-supervised pretraining has spurred remarkable progress in various vision tasks~\cite{he2020momentum,chen2020simple,chen2020big}, and this motivates us to leverage the powerful transferability of the self-supervised pretrained models in our domain adaptive semantic segmentation. To this end, 
After the training with Equation~\ref{eq:total_loss} converges, we further transfer knowledge from the learned target model to a student model with the same architecture but pretrained in a self-supervised manner. To be concrete, we initialize the feature extractor of the student model $h^{\dagger}$ with SimCLRv2~\cite{chen2020big} pretrained weights, and we apply a knowledge distillation (KD) loss, which lets the student mimic the teacher by minimizing the KL divergence of their predictions on the unlabeled target images. Besides, following the self-training paradigm, we rely on the teacher model $h$ to generate one-hot pseudo labels $\xi({p_t})$ so as to teach the student model. To prevent the model from forgetting the source domain,  the source images are also utilized. Altogether, we train the student model using the following loss,
\begin{equation}
    \ell_{\text{KD}} = \ell_{ce}^s(p_s,y_s) + \ell_{ce}^t(p_t^{\dagger},\xi(p_t)) + \beta\text{KL}(p_t \| p_t^{\dagger}),
\end{equation}
where $p_t^{\dagger} = h^{\dagger}(x_t)$ is the output of the student model, and we set $\beta=1$. In practice, such self-distillation can be applied multiple times once the model converges, which helps the domain adaptation to achieve even higher performance. 

%-------------------------------------------------------------------------
\section{Experiments}

\begin{table*}[htbp]
\centering
\footnotesize
%\resizebox{\columnwidth}{!}
\setlength\tabcolsep{2.4pt}{
\begin{tabular}{c|*{19}{c}|cc}
\toprule 
 & \multicolumn{1}{c}{\begin{sideways}road\end{sideways}} & \multicolumn{1}{c}{\begin{sideways}sideway\end{sideways}} & \multicolumn{1}{c}{\begin{sideways}building\end{sideways}} & \multicolumn{1}{c}{\begin{sideways}wall\end{sideways}} & \multicolumn{1}{c}{\begin{sideways}fence\end{sideways}} & \multicolumn{1}{c}{\begin{sideways}pole\end{sideways}} & \multicolumn{1}{c}{\begin{sideways}light\end{sideways}} & \multicolumn{1}{c}{\begin{sideways}sign\end{sideways}} & \multicolumn{1}{c}{\begin{sideways}vege.\end{sideways}} & \multicolumn{1}{c}{\begin{sideways}terrace\end{sideways}} & \multicolumn{1}{c}{\begin{sideways}sky\end{sideways}} & \multicolumn{1}{c}{\begin{sideways}person\end{sideways}} & \multicolumn{1}{c}{\begin{sideways}rider\end{sideways}} & \multicolumn{1}{c}{\begin{sideways}car\end{sideways}} & \multicolumn{1}{c}{\begin{sideways}truck\end{sideways}} & \multicolumn{1}{c}{\begin{sideways}bus\end{sideways}} & \multicolumn{1}{c}{\begin{sideways}train\end{sideways}} & \multicolumn{1}{c}{\begin{sideways}motor\end{sideways}} & \multicolumn{1}{c}{\begin{sideways}bike\end{sideways}} & \multicolumn{1}{|l}{mIoU} & \multicolumn{1}{l}{gain}\\
\midrule
 Source & 75.8 & 16.8 & 77.2 & 12.5 & 21.0 & 25.5 & 30.1 & 20.1 & 81.3 & 24.6 & 70.3 & 53.8 & 26.4 & 49.9 & 17.2 & 25.9 & 6.5 & 25.3 & 36.0 & 36.6 & +0.0\\
%\cline{1-1} \cline{3-22} 
\midrule
AdaptSeg~\cite{tsai2018learning} &86.5 & 25.9 & 79.8 & 22.1& 20.0& 23.6& 33.1& 21.8& 81.8& 25.9& 75.9& 57.3& 26.2& 76.3& 29.8& 32.1& 7.2& 29.5& 32.5& 41.4& +4.8\\
%\cline{1-1} \cline{3-22}
CyCADA~\cite{hoffman2018cycada} & 86.7& 35.6& 80.1& 19.8& 17.5& 38.0& 39.9& 41.5& 82.7& 27.9& 73.6& 64.9& 19.0& 65.0& 12.0& 28.6& 4.5& 31.1& 42.0& 42.7& +6.1\\
%\cline{1-1} \cline{3-22}    
CLAN~\cite{luo2019taking} & 87.0 & 27.1 & 79.6 & 27.3 & 23.3 & 28.3 & 35.5 & 24.2 & 83.6 & 27.4 & 74.2 & 58.6 & 28.0 & 76.2 & 33.1 & 36.7 & 6.7 & 31.9 & 31.4 & 43.2& +6.6\\
%\cline{1-1} \cline{3-22}    
APODA~\cite{yang2020adversarial}&85.6 & 32.8 & 79.0 & 29.5 & 25.5 & 26.8 & 34.6 & 19.9 & 83.7 & 40.6 & 77.9 & 59.2 & 28.3 & 84.6 & 34.6 & 49.2 & 8.0 & 32.6 & 39.6 & 45.9& +9.3\\
%\cline{1-1} \cline{3-22}   
PatchAlign~\cite{tsai2019domain}&\textbf{92.3} & 51.9 & 82.1 & 29.2 & 25.1 & 24.5 & 33.8 & 33.0 & 82.4 & 32.8 & 82.2 & 58.6 & 27.2 & 84.3 & 33.4 & 46.3 & 2.2 & 29.5 & 32.3 & 46.5& +9.9\\
%\cline{1-1} \cline{3-22}   
ADVENT~\cite{vu2019advent} &89.4&  33.1&  81.0&  26.6&  26.8&  27.2&  33.5&  24.7&  83.9&  36.7&  78.8&  58.7&  30.5&  84.8&  38.5&  44.5&  1.7&  31.6&  32.4&  45.5& +8.9\\
%\cline{1-1} \cline{3-22}  
BDL~\cite{li2019bidirectional} &91.0& 44.7& 84.2& 34.6& 27.6& 30.2& 36.0& 36.0& 85.0& 43.6& 83.0& 58.6& 31.6& 83.3& 35.3& 49.7& 3.3& 28.8& 35.6& 48.5& +11.9\\
FADA~\cite{Haoran_2020_ECCV} &91.0& 50.6& \textbf{86.0}& 43.4& 29.8& 36.8& 43.4& 25.0& 86.8& 38.3& \textbf{87.4}& 64.0& 38.0& 85.2& 31.6& 46.1& 6.5& 25.4& 37.1& 50.1 & +13.5\\
\midrule 
%\cline{1-1} \cline{3-22}   
CBST~\cite{zou2018unsupervised} & 91.8 & 53.5& 80.5& 32.7& 21.0& 34.0& 28.9& 20.4& 83.9& 34.2& 80.9& 53.1& 24.0& 82.7& 30.3& 35.9& 16.0& 25.9& 42.8& 45.9 & +9.3\\
%\cline{1-1} \cline{3-22}    
MRKLD~\cite{zou2019confidence} &91.0& 55.4& 80.0& 33.7& 21.4& 37.3& 32.9& 24.5& 85.0& 34.1& 80.8& 57.7& 24.6& 84.1& 27.8& 30.1& 26.9 & 26.0& 42.3& 47.1& +10.5\\
%\cline{1-1} \cline{3-22}  
CAG\_UDA~\cite{zhang2019category}  & 90.4& 51.6& 83.8& 34.2& 27.8& 38.4& 25.3& 48.4& 85.4& 38.2& 78.1& 58.6& 34.6& 84.7& 21.9& 42.7& \textbf{41.1}& 29.3& 37.2& 50.2 & +13.6\\
Seg-Uncertainty~\cite{zheng2020rectifying} &90.4& 31.2& 85.1& 36.9& 25.6& 37.5& 48.8& 48.5& 85.3& 34.8& 81.1& 64.4& 36.8& 86.3& 34.9& 52.2& 1.7& 29.0& 44.6& 50.3& +13.7\\
%\cline{1-1} \cline{3-22} 
\emph{ProDA}  & \cellcolor{gray}87.8 & \cellcolor{gray}\textbf{56.0}  & \cellcolor{gray}79.7  & {\cellcolor{gray}}\textbf{46.3} & {\cellcolor{gray}}\textbf{44.8}  & {\cellcolor{gray}}\textbf{45.6}  & {\cellcolor{gray}}\textbf{53.5}  & {\cellcolor{gray}}\textbf{53.5} & {\cellcolor{gray}}\textbf{88.6}  & {\cellcolor{gray}}\textbf{45.2}  & {\cellcolor{gray}}82.1 & {\cellcolor{gray}}\textbf{70.7} & {\cellcolor{gray}}\textbf{39.2}  & {\cellcolor{gray}}\textbf{88.8}  & {\cellcolor{gray}}\textbf{45.5}  & {\cellcolor{gray}}\textbf{59.4}  & {\cellcolor{gray}}1.0  & {\cellcolor{gray}}\textbf{48.9} & {\cellcolor{gray}}\textbf{56.4}  &  {\cellcolor{gray}}\textbf{57.5}& {\cellcolor{gray}}\textbf{+20.9}\\
\bottomrule  
\end{tabular}%
}
\caption {Comparison results of GTA5$\to$Cityscapes adaptation in terms of mIoU. The best score for each column is highlighted.}
\label{tab:Comparison results from GTA5 to Cityscapes}%
\end{table*}%

\begin{table*}[htbp]
\centering
\footnotesize
\setlength\tabcolsep{2.4pt}{
\begin{tabular}{@{}c|*{16}{c}|cc|cc@{}}
\toprule
 & \multicolumn{1}{c}{\begin{sideways}road\end{sideways}} & \multicolumn{1}{c}{\begin{sideways}sideway\end{sideways}} & \multicolumn{1}{c}{\begin{sideways}building\end{sideways}} & \multicolumn{1}{c}{\begin{sideways}wall*\end{sideways}} & \multicolumn{1}{c}{\begin{sideways}fence*\end{sideways}} & \multicolumn{1}{c}{\begin{sideways}pole*\end{sideways}} & \multicolumn{1}{c}{\begin{sideways}light\end{sideways}} & \multicolumn{1}{c}{\begin{sideways}sign\end{sideways}} & \multicolumn{1}{c}{\begin{sideways}vege.\end{sideways}} & \multicolumn{1}{c}{\begin{sideways}sky\end{sideways}} & \multicolumn{1}{c}{\begin{sideways}person\end{sideways}} & \multicolumn{1}{c}{\begin{sideways}rider\end{sideways}} & \multicolumn{1}{c}{\begin{sideways}car\end{sideways}}& \multicolumn{1}{c}{\begin{sideways}bus\end{sideways}} & \multicolumn{1}{c}{\begin{sideways}motor\end{sideways}} & \multicolumn{1}{c}{\begin{sideways}bike\end{sideways}}& \multicolumn{1}{|l}{mIoU}& \multicolumn{1}{l}{gain} & \multicolumn{1}{|l}{mIoU*}& \multicolumn{1}{l}{gain*} \\
\midrule
 Source & 64.3 & 21.3& 73.1& 2.4& 1.1& 31.4& 7.0& 27.7& 63.1& 67.6& 42.2& 19.9& 73.1& 15.3& 10.5& 38.9& 34.9 & + 0.0&40.3 & +0.0\\
\midrule
%\cline{1-1} \cline{3-22} 
AdaptSeg~\cite{tsai2018learning} & 79.2& 37.2& 78.8& -& -& -& 9.9& 10.5& 78.2& 80.5& 53.5& 19.6& 67.0& 29.5& 21.6& 31.3&- &- &45.9 & +5.6\\
%\cline{1-1} \cline{3-22}   
PatchAlign~\cite{tsai2019domain} & 82.4 & 38.0 & 78.6 & 8.7 & 0.6 & 26.0 & 3.9 & 11.1 & 75.5 & \textbf{84.6} & 53.5 & 21.6 & 71.4 & 32.6 & 19.3 & 31.7 & 40.0 & +5.1& 46.5& +6.2\\
%\cline{1-1} \cline{3-22}    
CLAN~\cite{luo2019taking} &81.3 & 37.0 & 80.1 & - & - & - & 16.1 & 13.7 & 78.2 & 81.5 & 53.4 & 21.2 & 73.0 & 32.9 & 22.6 & 30.7 & - & - & 47.8 & +7.5\\
%\cline{1-1} \cline{3-22}   
APODA~\cite{yang2020adversarial} & 86.4 & 41.3 & 79.3 & - & - & - & 22.6 & 17.3 & 80.3 & 81.6 & 56.9 & 21.0 & 84.1 & 49.1 & 24.6 & 45.7  & - & -& 53.1 & +12.8\\
%\cline{1-1} \cline{3-22}  
ADVENT~\cite{vu2019advent} & 85.6& 42.2& 79.7& 8.7& 0.4& 25.9& 5.4& 8.1& 80.4& 84.1& 57.9& 23.8& 73.3& 36.4& 14.2& 33.0& 41.2 & +6.3& 48.0 & +7.7\\ 
BDL~\cite{li2019bidirectional}   & 86.0& \textbf{46.7}& 80.3& -& -& -& 14.1& 11.6& 79.2& 81.3& 54.1& 27.9& 73.7& 42.2& 25.7& 45.3 &- & -& 51.4 & +11.1\\
FADA~\cite{Haoran_2020_ECCV}  & 84.5& 40.1& 83.1& 4.8& 0.0& 34.3& 20.1& 27.2& 84.8& 84.0& 53.5& 22.6& 85.4& 43.7& 26.8& 27.8  &45.2 & +10.3& 52.5 & +12.2\\
\midrule  
CBST~\cite{zou2018unsupervised}  & 68.0& 29.9& 76.3& 10.8& 1.4& 33.9& 22.8& 29.5& 77.6& 78.3& 60.6& 28.3& 81.6& 23.5& 18.8& 39.8& 42.6 & +7.7& 48.9 & +8.6\\
%\cline{1-1} \cline{3-22}    
MRKLD~\cite{zou2019confidence}  &67.7& 32.2& 73.9& 10.7& 1.6& 37.4& 22.2& 31.2& 80.8& 80.5& 60.8& \textbf{29.1}& 82.8& 25.0& 19.4& 45.3& 43.8 & +8.9& 50.1 & +9.8\\
%\cline{1-1} \cline{3-22}  
%\cline{1-1} \cline{3-22}  
CAG\_UDA~\cite{zhang2019category}  &84.7& 40.8& 81.7& 7.8& 0.0& 35.1& 13.3& 22.7& 84.5& 77.6& 64.2& 27.8& 80.9& 19.7& 22.7& 48.3 & 44.5 & +9.6&51.5 & +11.2\\
Seg-Uncertainty~\cite{zheng2020rectifying} & 87.6& 41.9& 83.1& 14.7& \textbf{1.7}& 36.2& 31.3& 19.9& 81.6& 80.6& 63.0& 21.8& 86.2& 40.7& 23.6& \textbf{53.1} & 47.9 & +13.0& 54.9 & +14.6\\
%\cline{1-1} \cline{3-22} 
\emph{ProDA}  & {\cellcolor{gray}}\textbf{87.8} &   {\cellcolor{gray}}45.7 & {\cellcolor{gray}}\textbf{84.6} &  {\cellcolor{gray}}\textbf{37.1} & {\cellcolor{gray}}0.6 &  {\cellcolor{gray}}\textbf{44.0} &  {\cellcolor{gray}}\textbf{54.6} &  {\cellcolor{gray}}\textbf{37.0} &  {\cellcolor{gray}}\textbf{88.1} &  {\cellcolor{gray}}84.4 & {\cellcolor{gray}}\textbf{74.2}  &  {\cellcolor{gray}}24.3 &  {\cellcolor{gray}}\textbf{88.2} &  {\cellcolor{gray}}\textbf{51.1} &  {\cellcolor{gray}}\textbf{40.5} &  {\cellcolor{gray}}45.6 & {\cellcolor{gray}}\textbf{55.5} & {\cellcolor{gray}}\textbf{+20.6}& {\cellcolor{gray}}\textbf{62.0} & {\cellcolor{gray}}\textbf{+21.7}\\
\bottomrule  
\end{tabular}%
}
\caption {Comparison results of SYNTHIA$\to$Cityscapes adaptation in terms of mIoU. The best score for each column is highlighted. mIoU and mIoU* denote the averaged scores across 16 and 13 categories respectively.}
\label{tab:Comparison results from SYNTHIA to Cityscapes}%
\end{table*}%

\subsection{Implementation}
\textbf{Training}. 
We use the DeepLabv2~\cite{chen2017deeplab} for segmentation with the backbone ResNet-101~\cite{he2016deep}. Following~\cite{zhang2019category,zheng2020rectifying}, we utilize ~\cite{Tsai_adaptseg_2018} that applies adversarial training at the segmentation output as a warm-up. We apply the SGD solver with the initial learning rate as 1e-4 which is decayed by 0.9 every training epoch, and the training lasts 80 epochs. During the structure learning, the augmentation is composed of random crop, RandAugment~\cite{cubuk2020randaugment} and  Cutout~\cite{devries2017improved}. For knowledge distillation, we utilize the pretrained SimCLRv2 model with the ResNet-101 backbone as well. An extra batch normalization (BN) layer is introduced after its feature extraction layer so as to accommodate the activation magnitude for our task, with the learning rate set to 6e-4 and 6e-3 respectively before and after this BN layer. During the distillation stage, we use hard pseudo labels with the selection threshold 0.95. Readers can refer to the supplementary material for more training details and algorithm flow. We conduct all the experiments on 4 Tesla V100 GPUs with PyTorch implementation. 

\textbf{Dataset}. 
For evaluation, we adapt the segmentation from game scenes, GTA5~\cite{richter2016playing} and SYNTHIA~\cite{ros2016synthia} datasets, to real scene, the Cityscapes~\cite{cordts2016cityscapes} dataset. GTA5 contains 24,966 training images with the resolution of 1914$\times$1052 and we use its 19 categories shared with Cityscapes. SYNTHIA dataset contains 9,400 1280$\times$760 images and we use its 16 common categories with Cityscapes.
We also report the results on 13 common categories on this dataset following the protocol of some methods. The Cityscapes dataset contains 2,975 training images and 500 images for validation with the resolution of 2048$\times$1024. Since its testing set does not provide ground truth labeling, we conduct evaluations on its validation set. 

\subsection{Comparisons with state-of-the-art methods}
We comprehensively compare our proposed method with the recent leading approaches. These methods could be divided into two categories: 1) domain alignment methods that align the distribution through adversarial training, which include AdaptSeg~\cite{tsai2018learning}, CyCADA~\cite{hoffman2018cycada}, CLAN~\cite{luo2019taking}, APODA~\cite{yang2020adversarial}, PatchAlign~\cite{tsai2019domain}, ADVENT~\cite{vu2019advent}, BDL~\cite{li2019bidirectional} and FADA~\cite{Haoran_2020_ECCV}; 2) self-training approaches, including CBST~\cite{zou2018unsupervised}, MRKLD~\cite{zou2019confidence}, Seg-Uncertainty~\cite{zheng2020rectifying}, CAG\_UDA~\cite{zhang2019category}.

Table~\ref{tab:Comparison results from GTA5 to Cityscapes} shows the comparisons of GTA5 $\to$ Cicyscapes adaptation. our ProDA arrives at the state-of-the-art mIoU score 57.5, outperforming existing methods by a large margin. Among all the 19 categories, we achieve the best scores on 15 categories. ProDA shows evident advantage in hard classes, \eg, fence, terrace, motor, that cannot be well handled in previous works. Indeed, the performance improvement of ProDA mostly comes from these challenging cases, the small or rare objects, as we regard different categories equally thanks to the prototypes. Note that this is achieved without any heuristic class-balance strategies as~\cite{zou2018unsupervised}. Compared with the non-adapted baseline (\ie, the model purely trained on the source), ProDA offers the mIoU gain by 20.9, outperforming the second-best method by 52.6\%. 

In Table~\ref{tab:Comparison results from SYNTHIA to Cityscapes}, we show the adaptation results on SYNTHIA $\to$ Cityscapes, where ProDA also shows tremendous improvement. The proposed ProDA achieves the mIoU score by 55.5 and 62.0 over the 16 and 13 categories respectively. To be specific, we arrive at the best on 11 out of 16 categories, mostly the hard classes. Relative to the non-adaptive model, the segmentation model after our adaption sees the gain by 20.6, surpassing that of the prior leading approach by 58.4\%. While the adaptation from SYNTHIA is more challenging than that from GTA5, our ProDA performs equally well on both datasets.

\begin{figure*}[t]
    \center
    \small
    \setlength\tabcolsep{1pt}
    {
    
    \newcolumntype{P}[1]{>{\centering\arraybackslash}p{#1}}
    \begin{tabular}{@{}*{10}{P{0.201\columnwidth}}@{}}
         {\cellcolor[rgb]{0.5,0.25,0.5}}\textcolor{white}{road} &{\cellcolor[rgb]{0.957,0.137,0.91}}sidewalk &{\cellcolor[rgb]{0.275,0.275,0.275}}\textcolor{white}{building} &{\cellcolor[rgb]{0.4,0.4,0.612}}\textcolor{white}{wall} &{\cellcolor[rgb]{0.745,0.6,0.6}}fence &{\cellcolor[rgb]{0.6,0.6,0.6}}pole &{\cellcolor[rgb]{0.98,0.667,0.118}}traffic light&{\cellcolor[rgb]{0.863,0.863,0}}traffic sign &{\cellcolor[rgb]{0.42,0.557,0.137}}vegetation & {\cellcolor[rgb]{0,0,0}}\textcolor{white}{n/a.}\\
         
         {\cellcolor[rgb]{0.596,0.984,0.596}}terrain &{\cellcolor[rgb]{0,0.51,0.706}}sky &{\cellcolor[rgb]{0.863,0.078,0.235}}\textcolor{white}{person} &{\cellcolor[rgb]{1,0,0}}\textcolor{white}{rider} &{\cellcolor[rgb]{0,0,0.557}}\textcolor{white}{car} &{\cellcolor[rgb]{0,0,0.275}}\textcolor{white}{truck} &{\cellcolor[rgb]{0,0.235,0.392}}\textcolor{white}{bus}&{\cellcolor[rgb]{0,0.314,0.392}}\textcolor{white}{train} &{\cellcolor[rgb]{0,0,0.902}}\textcolor{white}{motorcycle} & {\cellcolor[rgb]{0.467,0.043,0.125}}\textcolor{white}{bike}\\
    
    \end{tabular}
    
    \renewcommand{\arraystretch}{0.6}
    \begin{tabular}{@{}cccc@{}}
    %    \multicolumn{4}{c}{\includegraphics[width=2.088\columnwidth]{figure/class_color.png}}\\
    
        \includegraphics[width=0.518\columnwidth]{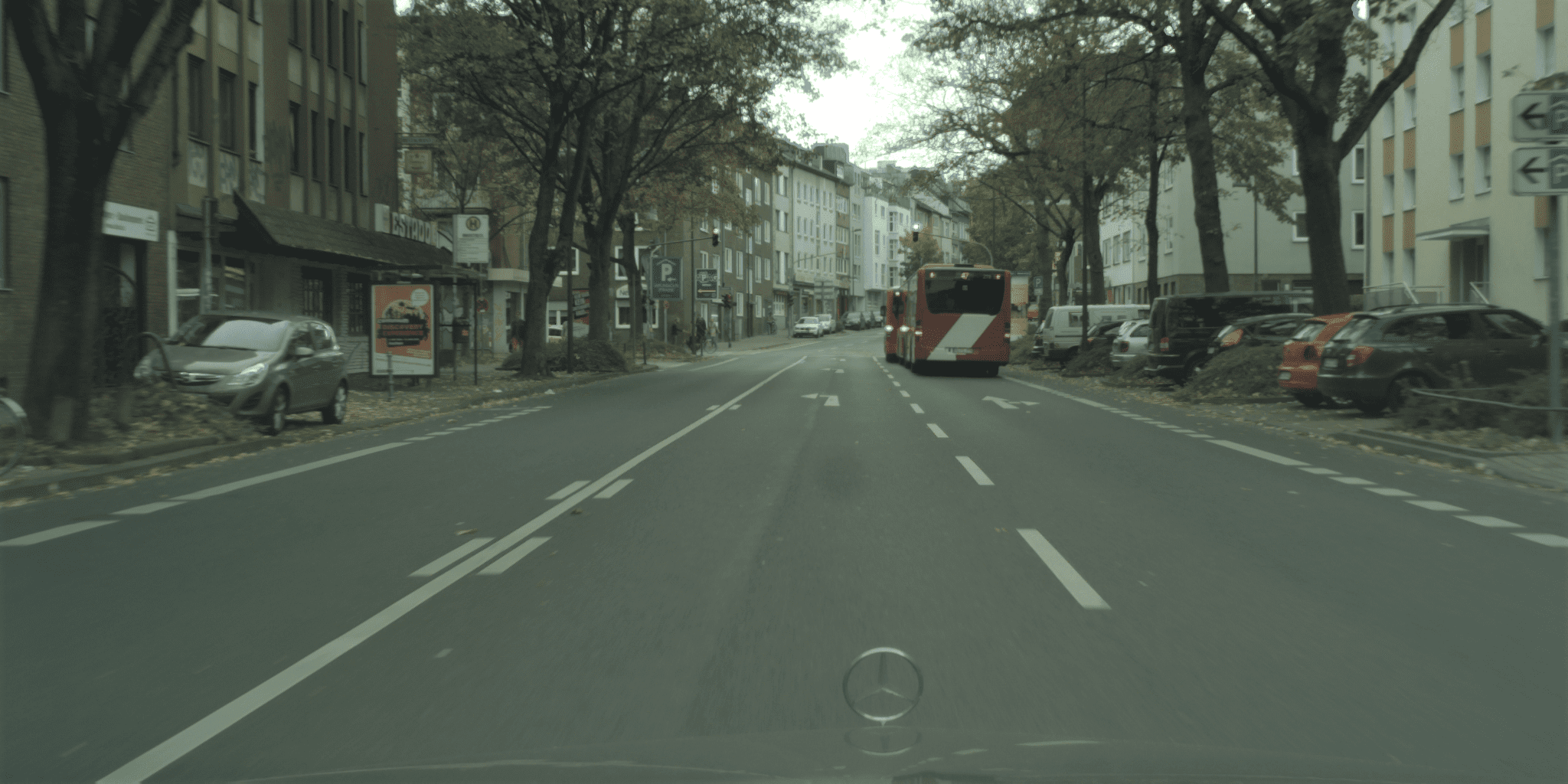}&
        \includegraphics[width=0.518\columnwidth]{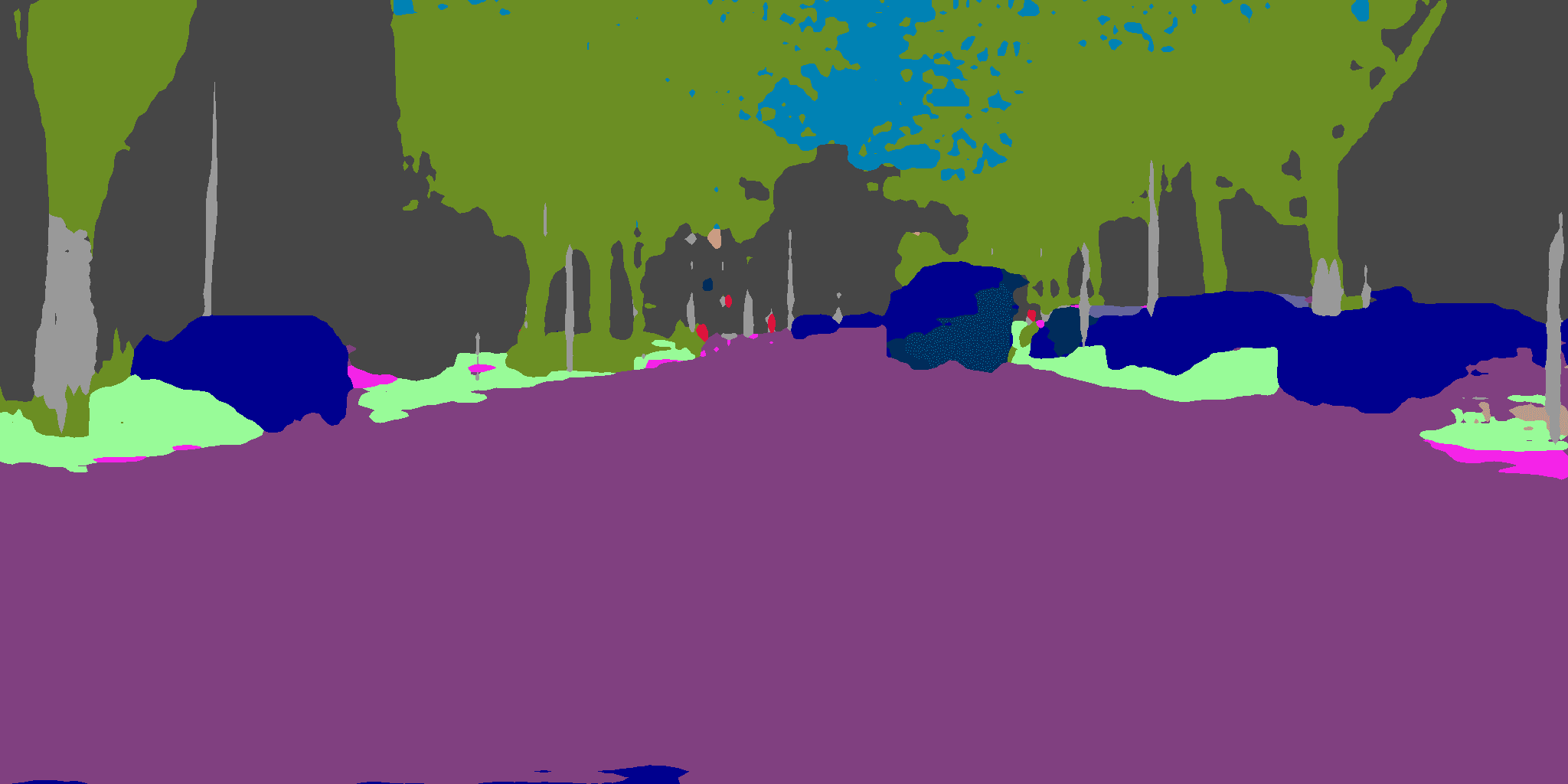}&
        \includegraphics[width=0.518\columnwidth]{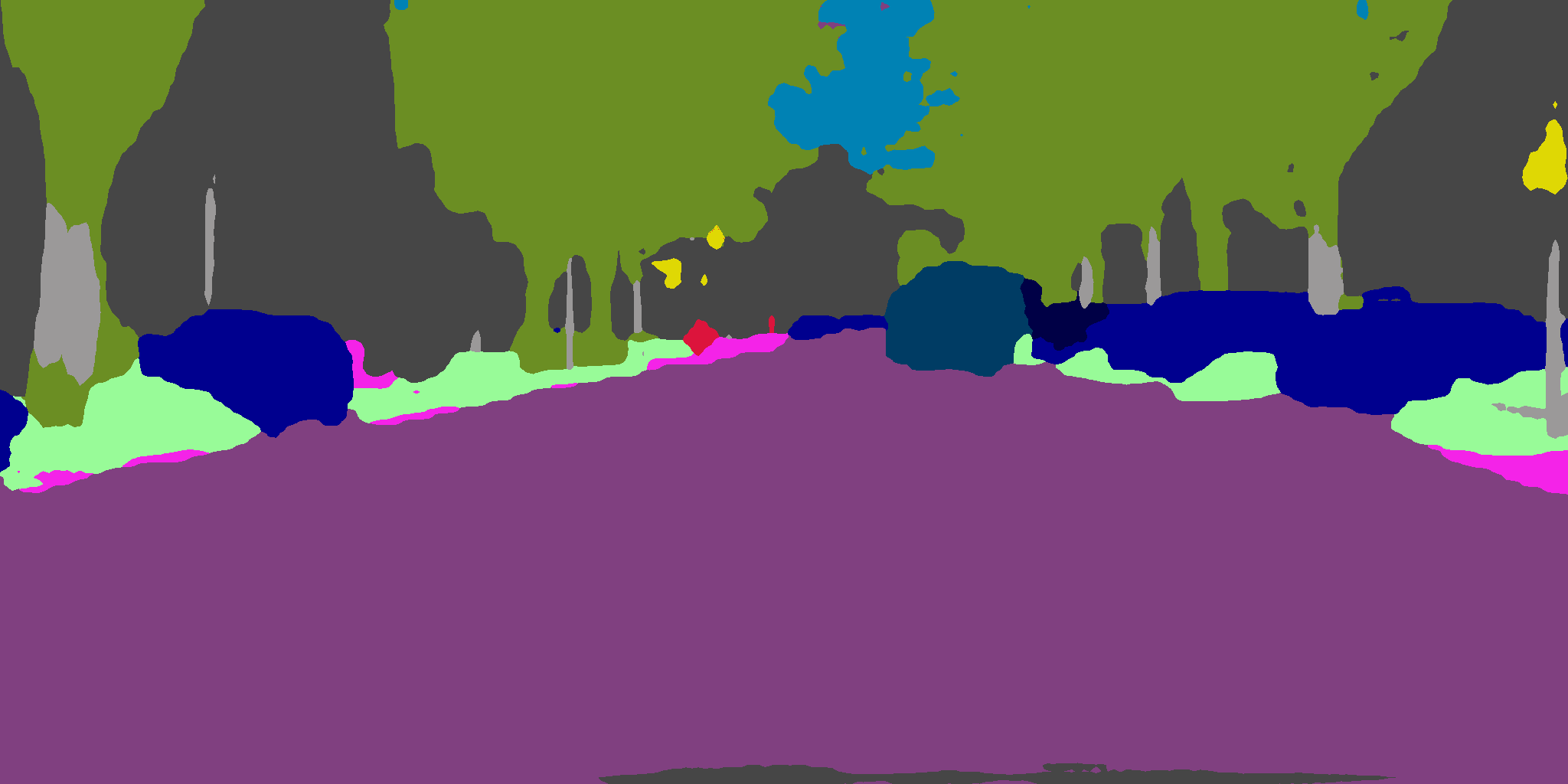}&
        \includegraphics[width=0.518\columnwidth]{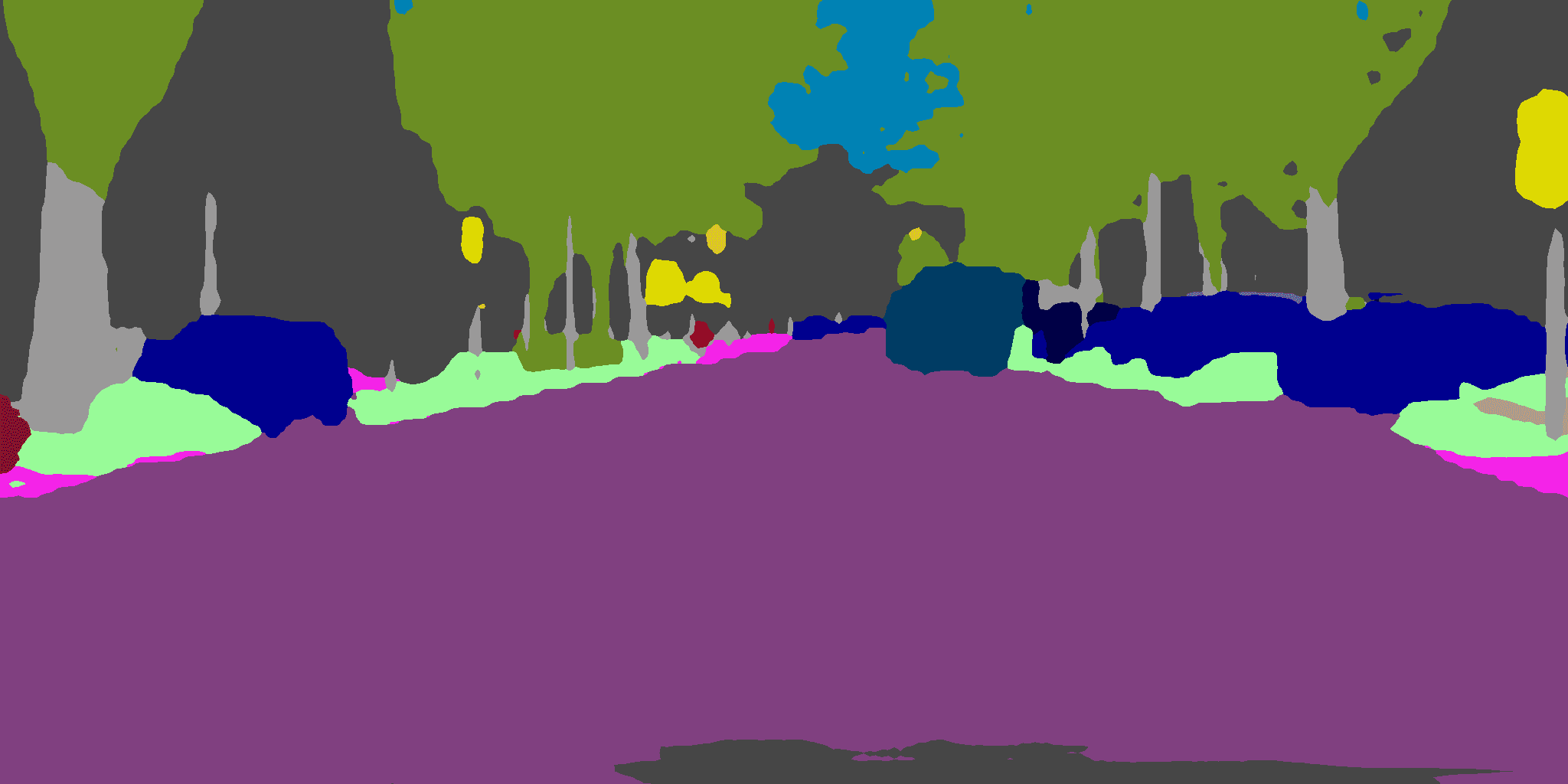}\\

        \includegraphics[width=0.518\columnwidth]{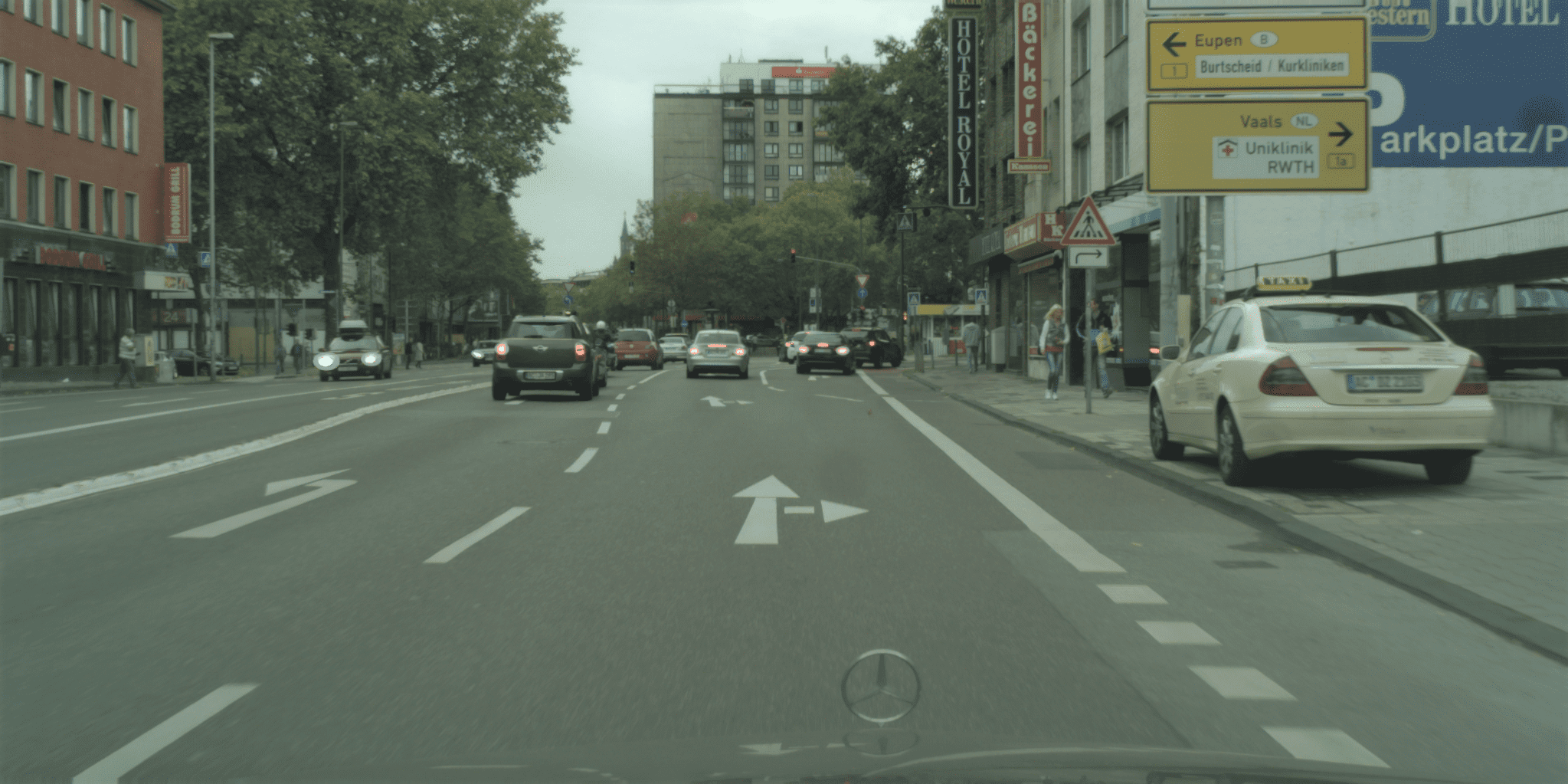}&
        \includegraphics[width=0.518\columnwidth]{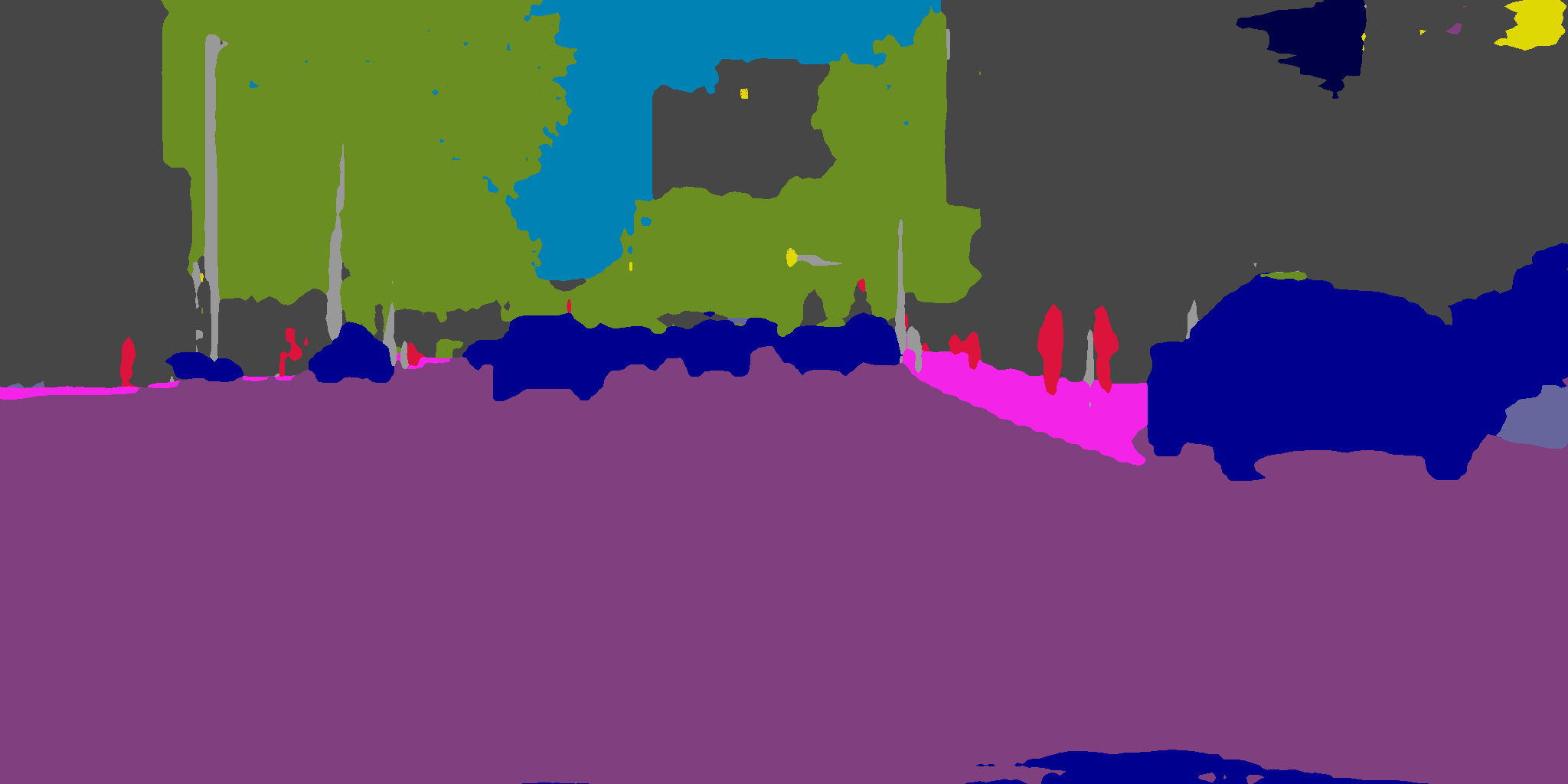}&
        \includegraphics[width=0.518\columnwidth]{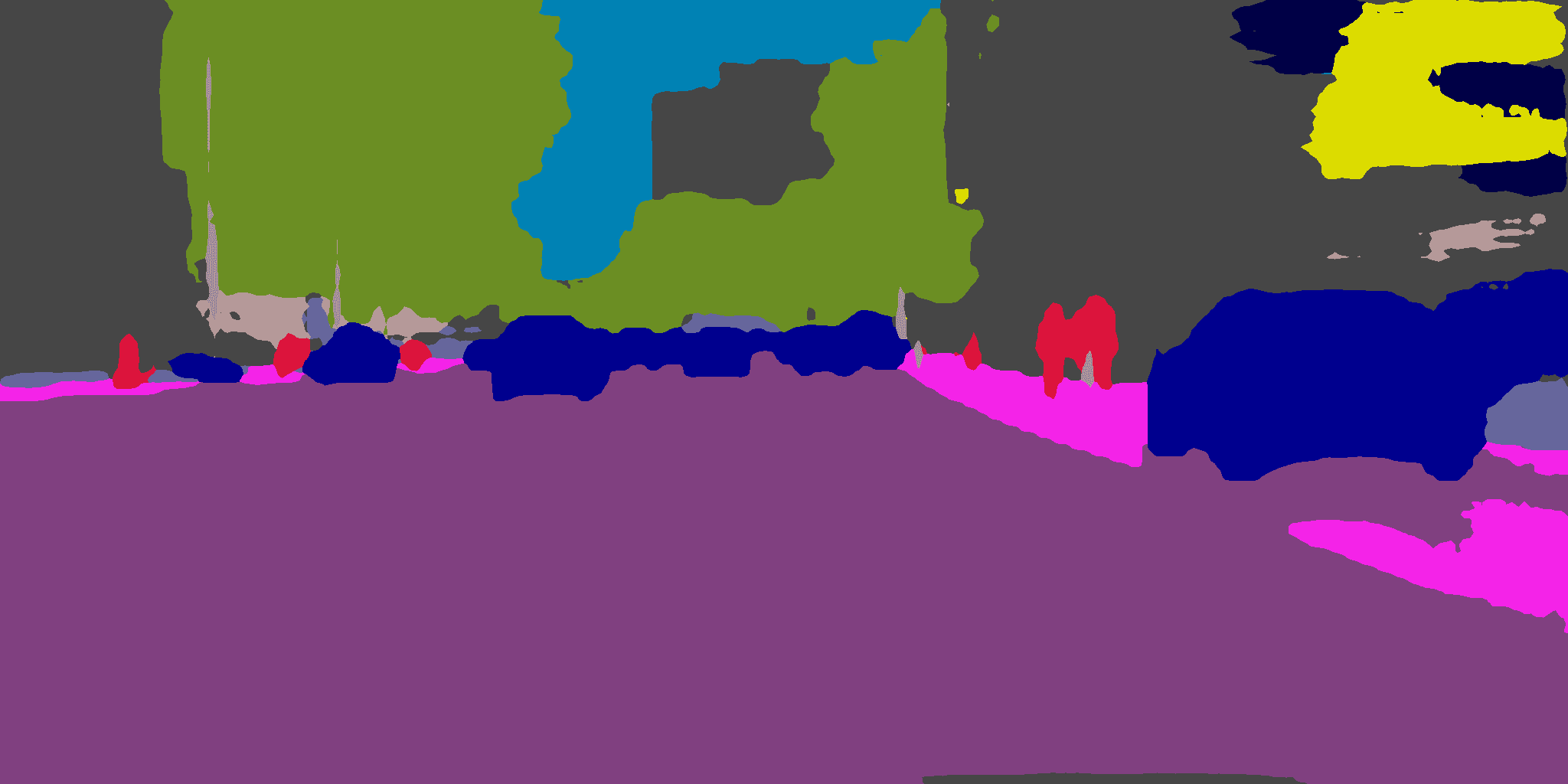}&
        \includegraphics[width=0.518\columnwidth]{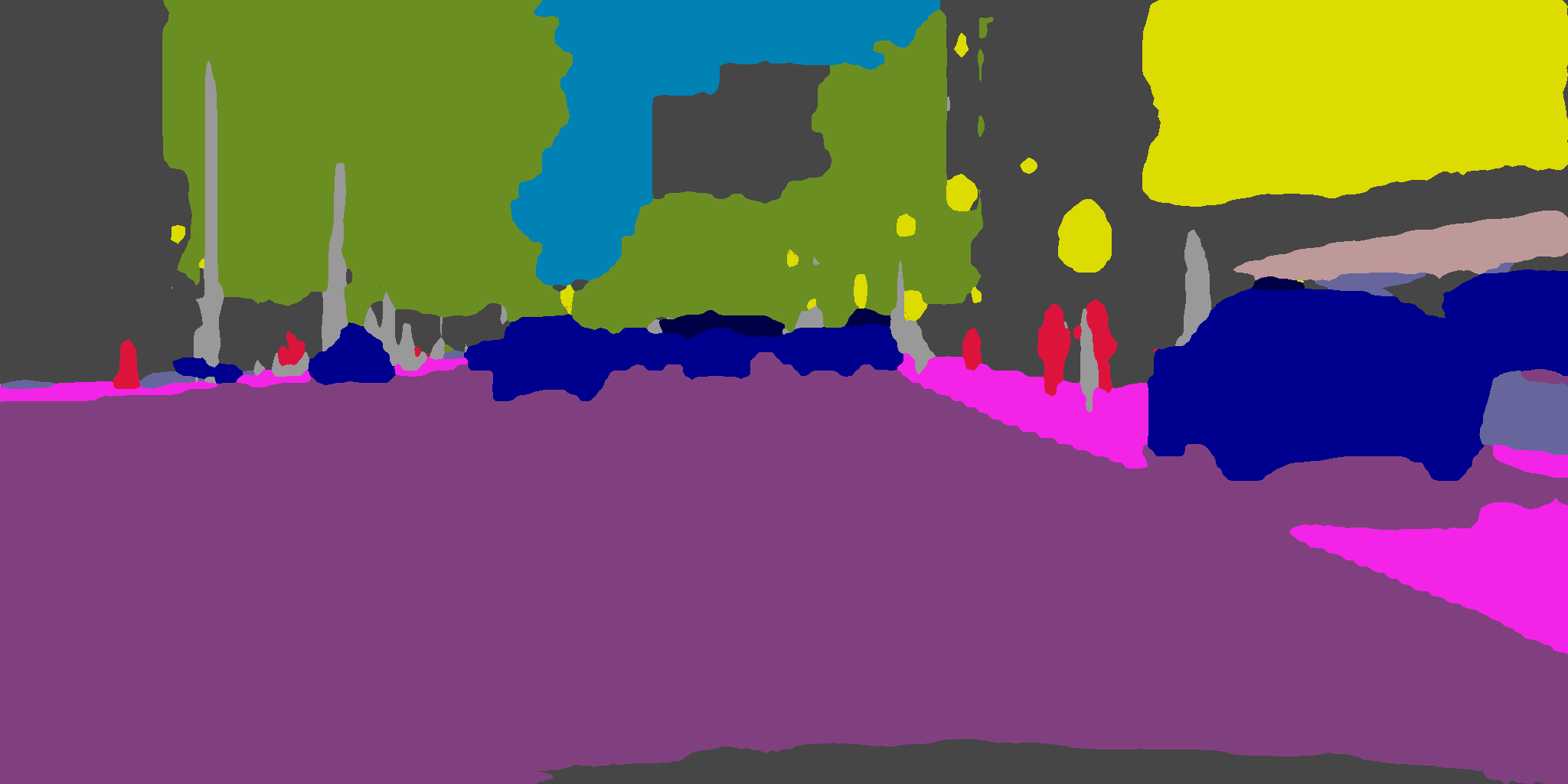}\\
        
    \end{tabular}
    }
    \caption{ProDA online refines the pseudo labels throughout the training. The first column is the segmentation input. The 2nd to 4th columns illustrate the pseudo labels after 1k, 10k, and 40k iterations.}
    \label{figure:online_updating_example}
    \end{figure*}

\subsection{Discussion}

\newcommand{\tabincell}[2]{\begin{tabular}{@{}#1@{}}#2\end{tabular}}  
\begin{table}[t]
\centering
\footnotesize
\setlength\tabcolsep{1.8pt}{
\begin{tabular}{@{}c|cccc|cc@{}}
\toprule 
&\multicolumn{4}{c|}{components} &\multicolumn{1}{l}{mIoU} & \multicolumn{1}{l}{gain}\\
\midrule
\multirow{2}{*}{initialization}&\multicolumn{4}{c|}{source} & 36.6 &\\
%\cline{1-1} \cline{3-22} 
&\multicolumn{4}{c|}{warm up} &  41.6 & +5.0\\
\midrule

\multirow{6}{*}{stage 1}&\tabincell{c}{self\\training} & sce &  \tabincell{c}{prototypical\\denoising} &  \tabincell{c}{structure\\learning}&\multicolumn{1}{l}{mIoU} & \multicolumn{1}{l}{gain}\\
%\midrule
\cline{2-7} 
&\checkmark & &  & & 45.2 & +8.6\\
&\checkmark & \checkmark &  & &  45.6 & +9.0\\
&\checkmark&\checkmark & \checkmark  & &  52.3 & +15.7\\
%\cline{1-1} \cline{3-22}    
&\checkmark&\checkmark &  & \checkmark &47.6 & +11.0\\
%\cline{1-1} \cline{3-22}    
&\checkmark&\checkmark & \checkmark & \checkmark & 53.7 & +17.1\\
\midrule  
\multirow{5}{*}{stage 2}& \tabincell{c}{self\\distill.} & \tabincell{c}{stage 1\\init.} & \tabincell{c}{supervised\\init.} & \tabincell{c}{self-supervised\\init.} &\multicolumn{1}{l}{mIoU} & \multicolumn{1}{l}{gain}\\
%\midrule  
\cline{2-7} 
&& & & \checkmark & 55.8 & +19.2\\
&\checkmark&\checkmark & & & 56.3 & +19.7\\
&\checkmark& & \checkmark &  & 55.7 & +19.1\\
&\checkmark& & & \checkmark & 56.9 & +20.3\\
\midrule 
stage 3 & \checkmark & & & \checkmark & 57.5 & +20.9\\
\bottomrule  
\end{tabular}%
}
\caption {Ablation study of each proposed component. The whole training involves three stages, where knowledge distillation can be applied in the last two stages. }
\label{tab:Component Ablation}%
\end{table}%

\noindent\textbf{The effectiveness of pseudo label denoising.}   
In Table~\ref{tab:Component Ablation}, the non-adapted source model only gives 36.6 mIoU on the target domain. After the model warm-up, we get a 5.0 mIoU increase. Initialized with the warm-up, the baseline model, \ie, the vanilla self-training, which is trained using the offline pseudo labels, improves the mIoU to 45.2. Adding symmetric cross-entropy (SCE) brings +0.4 mIoU gain. By contrast, we are able to update the pseudo labels on-the-fly and the training with the denoised pseudo labels significantly boosts the performance to 52.3. The model with this component alone sets the state-of-the-art, outperforming the prior best score by 2.0.

\begin{figure}[!t]
\center
\small
\setlength\tabcolsep{0pt}
{
    \begin{tabular}{cc}
        \includegraphics[width=0.5\columnwidth, trim=4cm 2cm 4cm 2cm, clip]{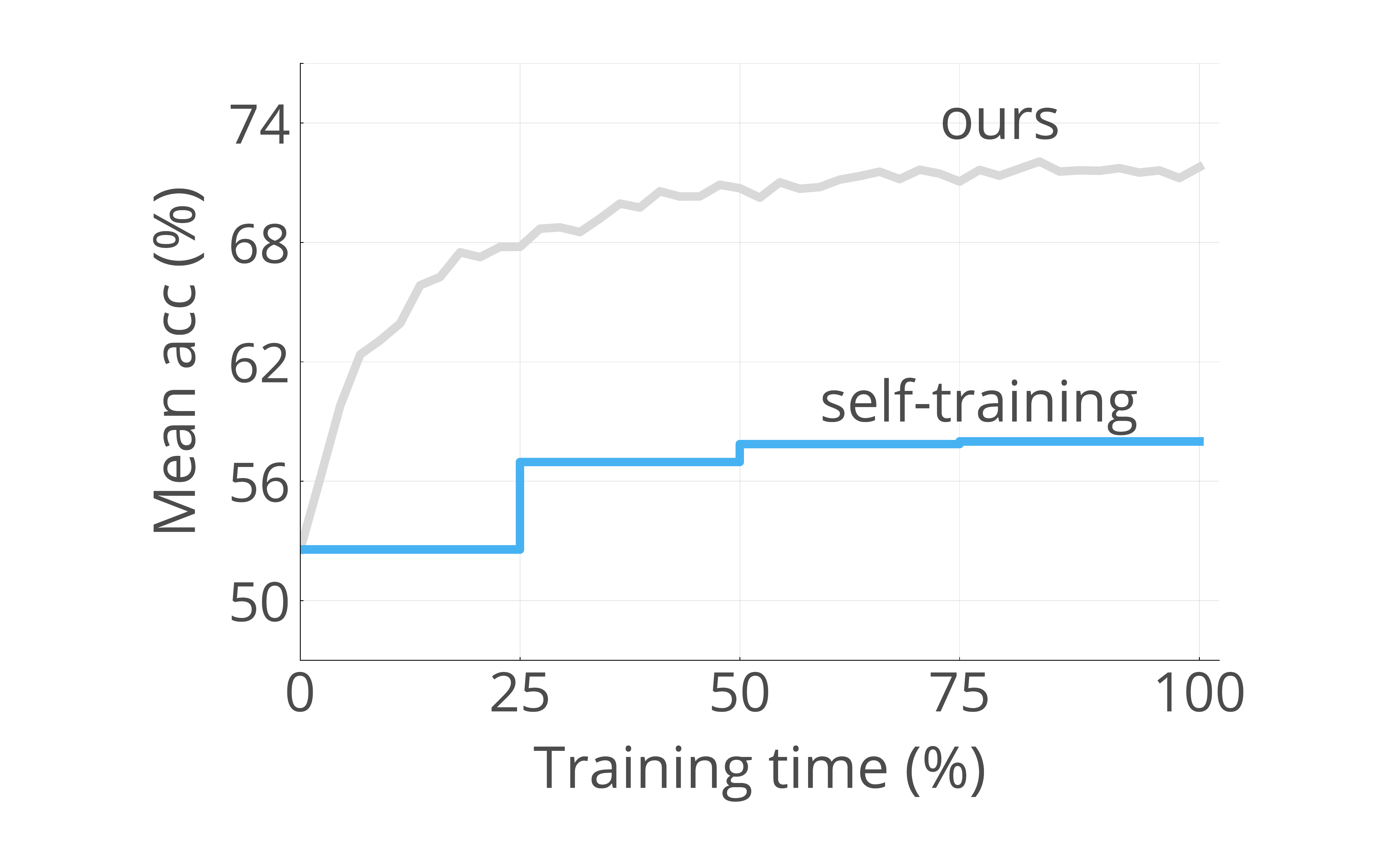} &
        \includegraphics[width=0.5\columnwidth, trim=4cm 2cm 4cm 2cm, clip]{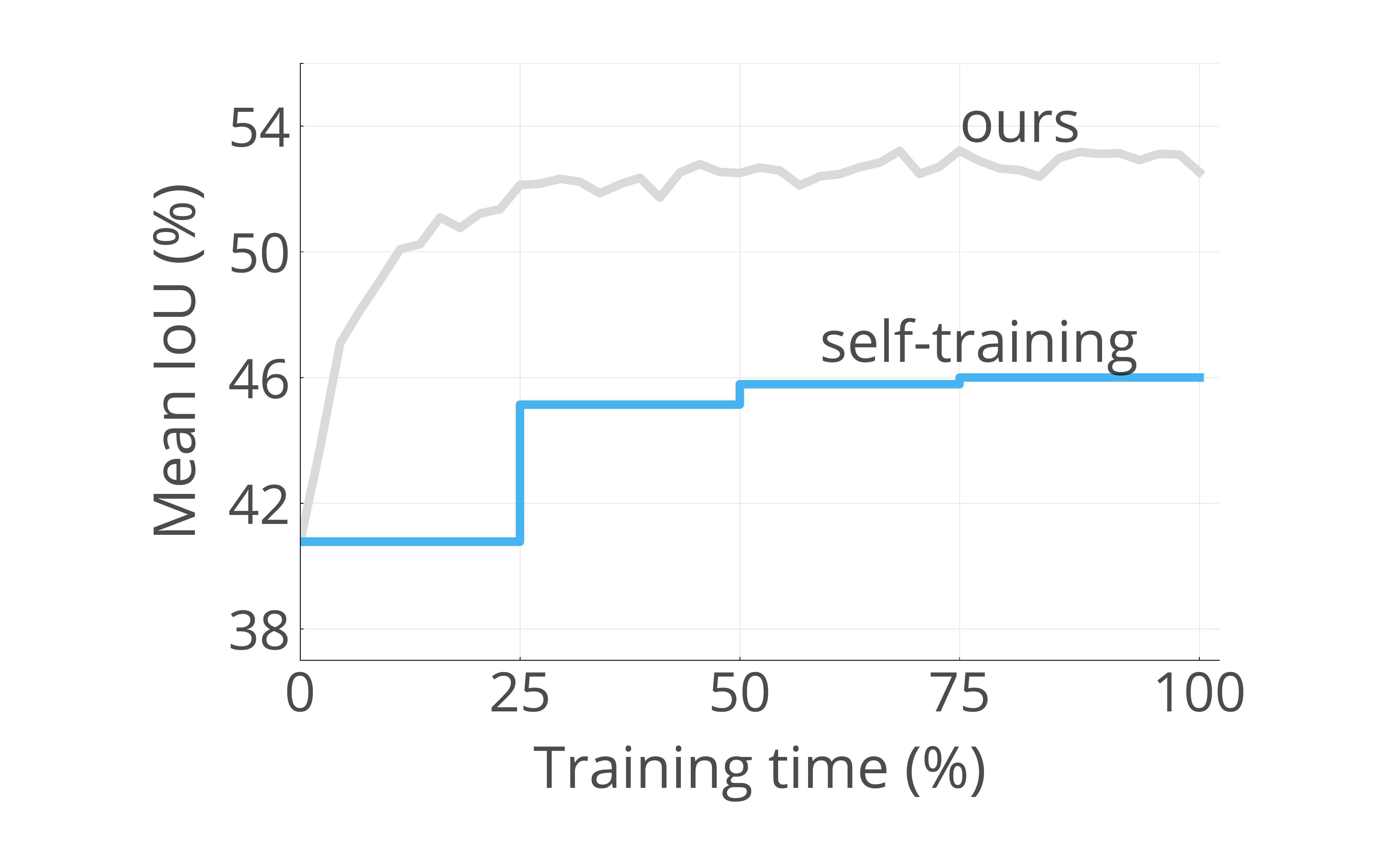}\\
    \end{tabular}
}
\caption{The mean accuracy and mean IoU score of the pseudo labels throughout the training. Comparing to the conventional self-training that updates pseudo labels only after the training stage, the pseudo labels in our method steadily improves as the training proceeds.}
\label{fig:supervison}
\end{figure}

Figure~\ref{figure:online_updating_example} illustrates how the pseudo labels progress as the training proceeds. In the early phase, false pseudo labels are likely to occur and fail to recognize the tiny objects, \eg, the poles and traffic signs. As the training goes on, pseudo labels get denoised by virtue of the prototypes, and tiny objects can be gradually identified in the refined pseudo labels. The training after 40k iterations can already correct most of the incorrect labels. In Figure~\ref{fig:supervison}, we further quantify the mIoU score of the pseudo labels during the whole training process. While convention self-training only updates the pseudo labels after each stage and leads to step-wise mIoU increase, the pseudo labels given by ProDA can quickly attain a high quality, showing a distinct advantage over the vanilla self-training of multiple training stages.

\begin{figure}[!t]
\center
\small
\setlength\tabcolsep{0pt}
{
    \includegraphics[width=1.0\columnwidth]{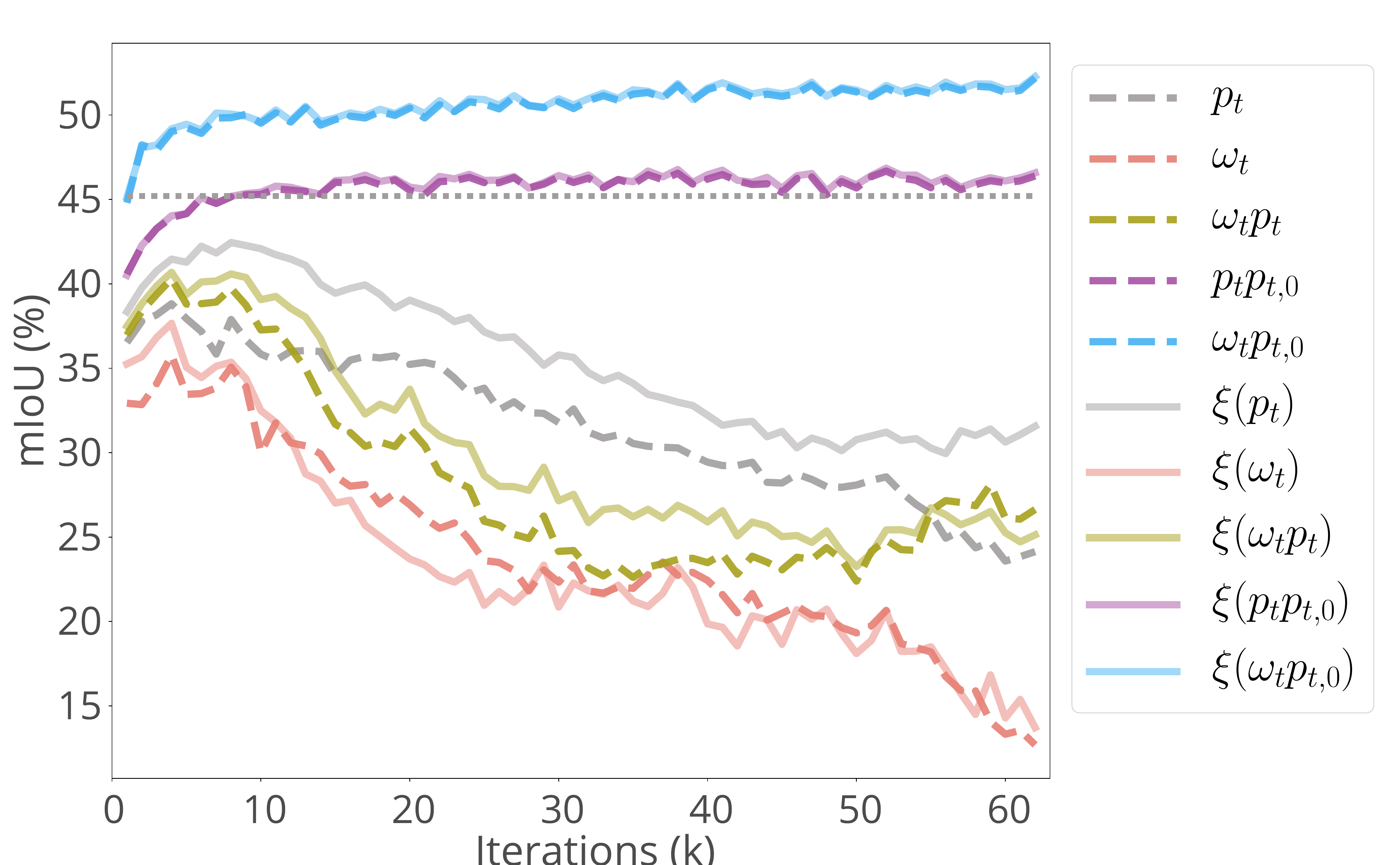}\\
}
\caption{The training curves of different label reweighting schemes when online updating the labels. Adopting fixed $p_{t,0}$ labels avoid the trivial solutions that plague the training with non-fixed pseudo labels $p_t$. The dotted line denotes the performance of conventional self-training.}
\vspace{-1.5em}
\label{fig:online_anaylse}
\end{figure}

\begin{figure*}[!t]
    \center
    \small
    \setlength\tabcolsep{0pt}
    {
    \renewcommand{\arraystretch}{0.0}
    \begin{tabular}{cccc}
        \includegraphics[width=0.5\columnwidth, trim=2cm 2cm 1cm 2cm, clip]{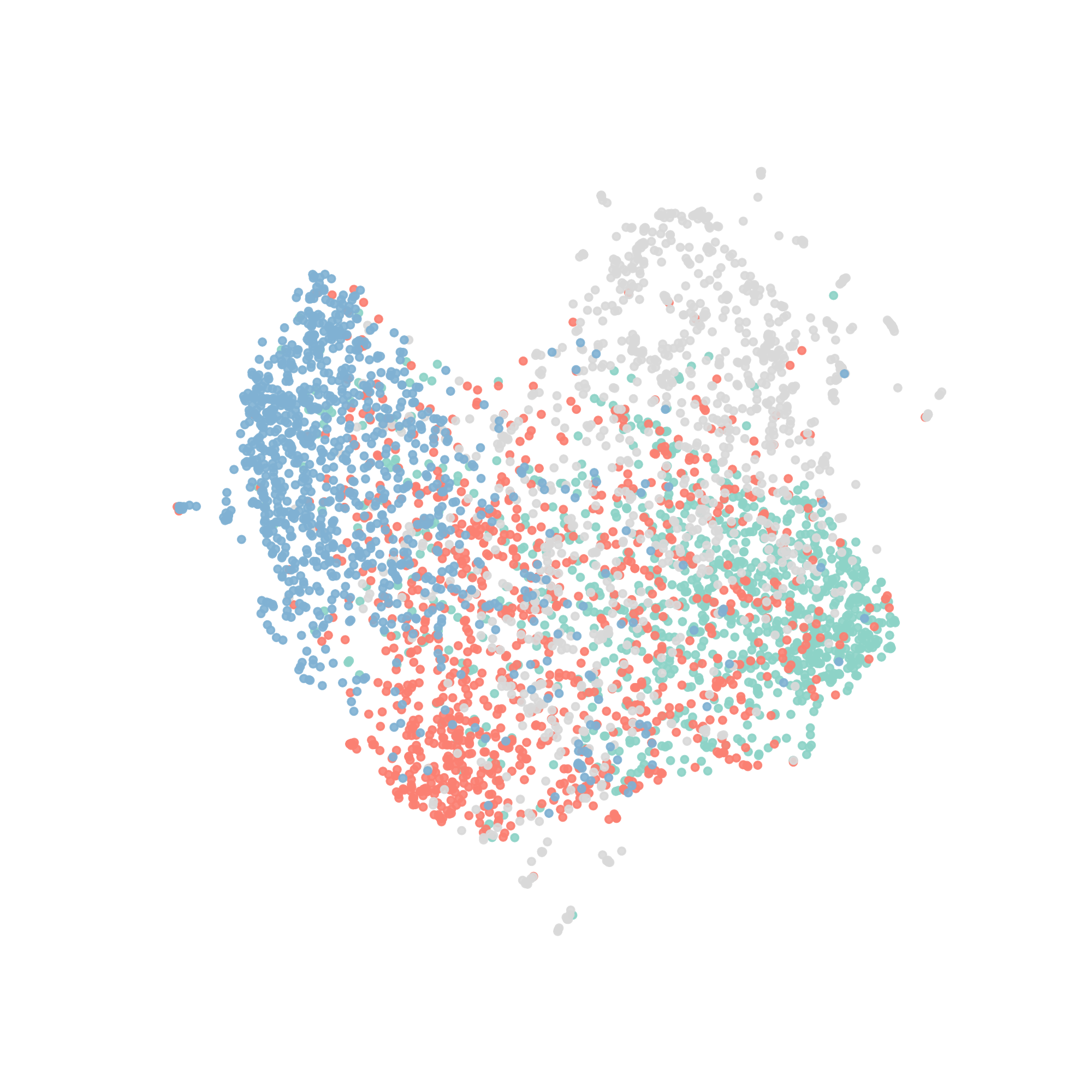} & 
        \includegraphics[width=0.5\columnwidth, trim=2cm 2cm 1cm 2cm, clip]{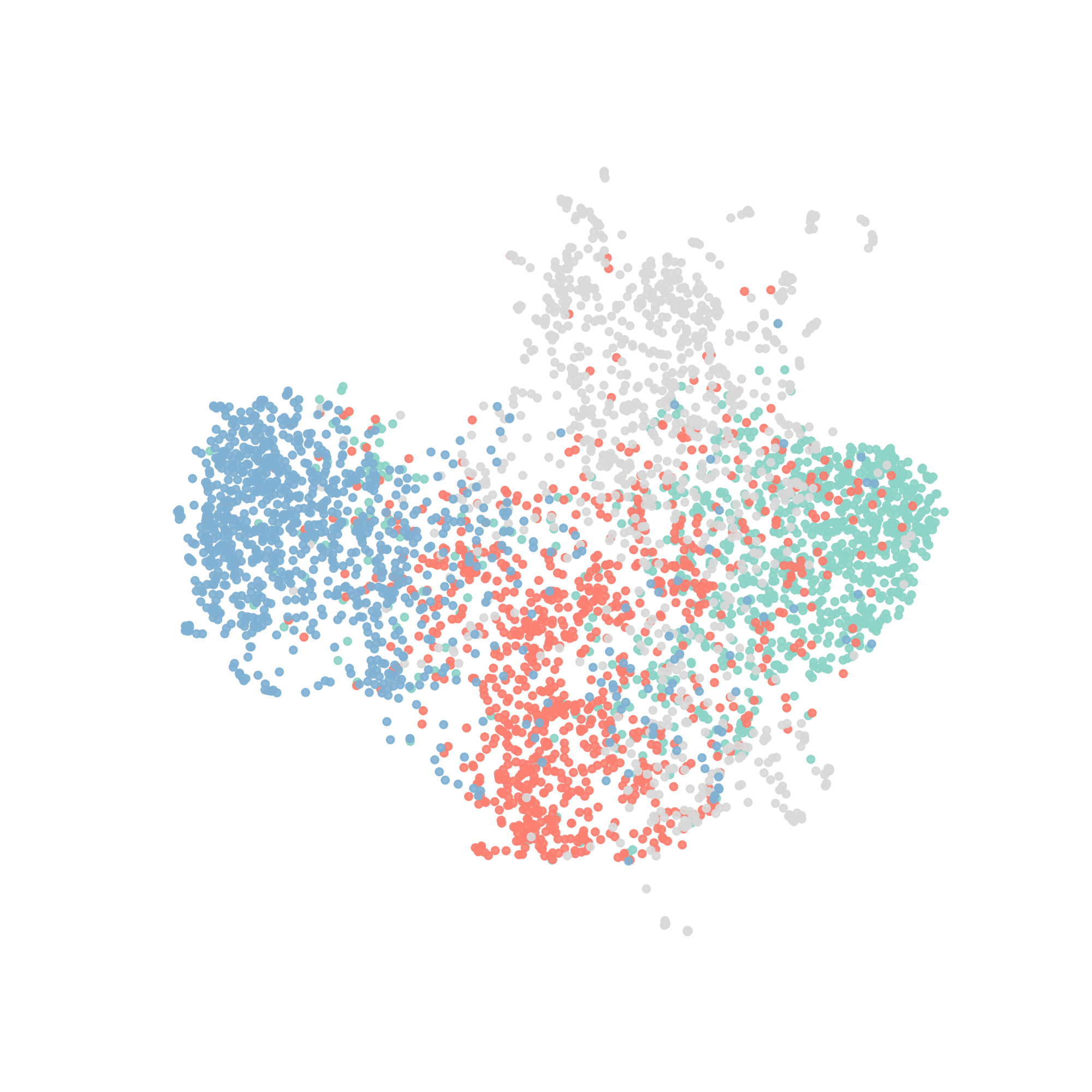} & 
        \includegraphics[width=0.5\columnwidth, trim=2cm 2cm 1cm 2cm, clip]{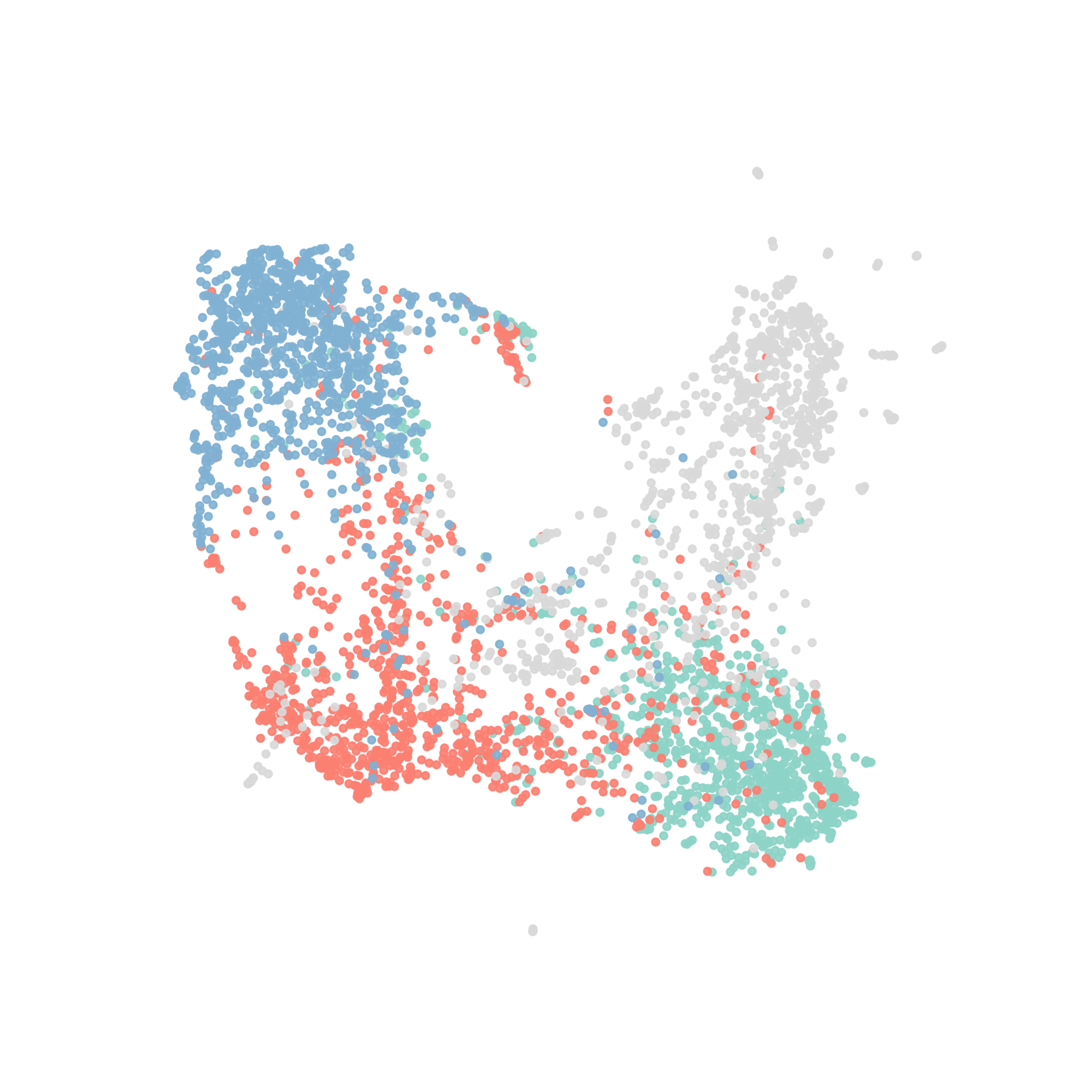} & 
        \includegraphics[width=0.5\columnwidth, trim=2cm 2cm 1cm 2cm, clip]{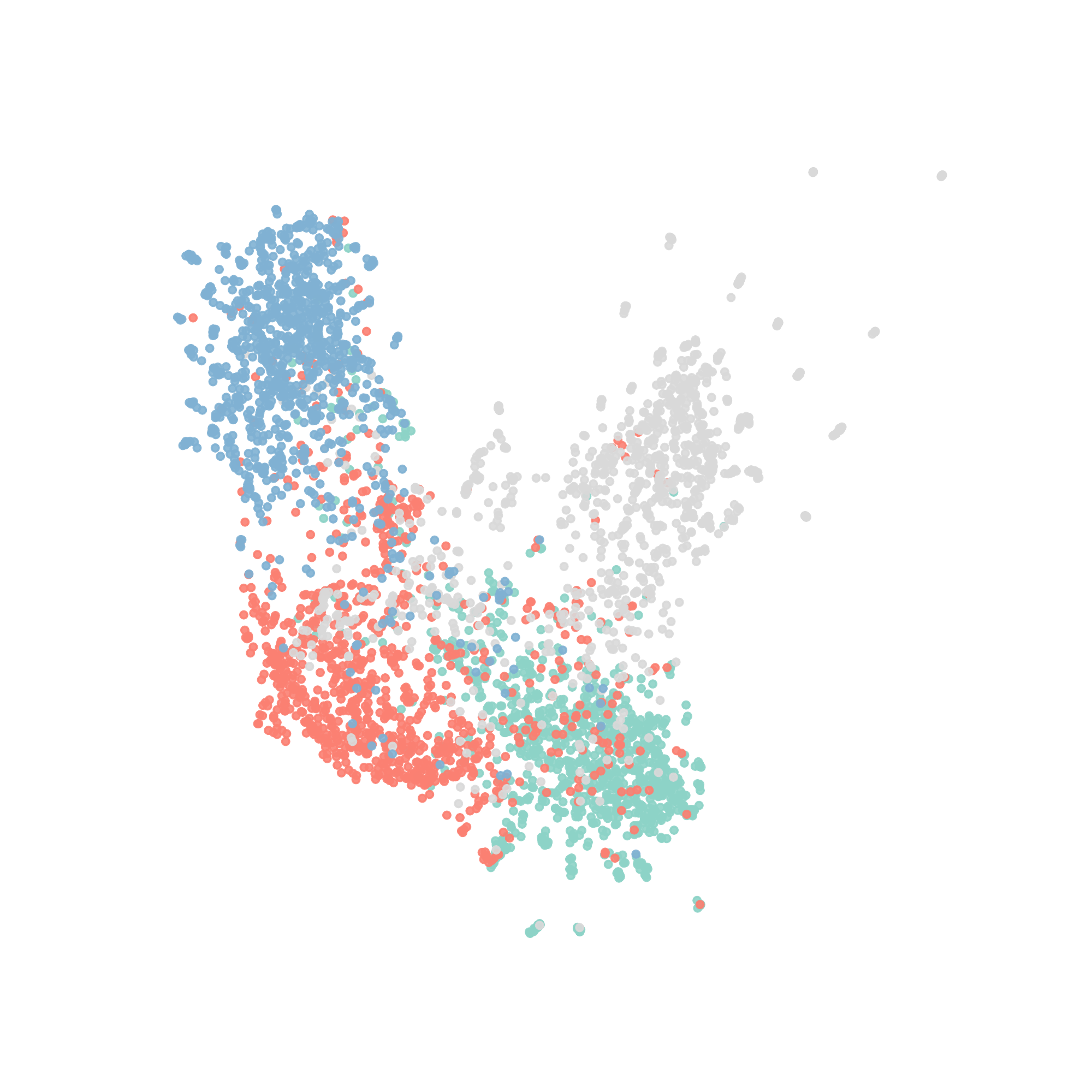}\\
        (a) w/o domain adaptation & (b) conventional self-training & (c) w/ prototypical denoising (1st stage) & (d) our full model\\
    \end{tabular}
    }
    \caption{The visualization of feature space, where we map features to 2D space with UMAP~\cite{mcinnes2018umap}. For a clear illustration, we only show four categories, \ie, blue for building, gray for traffic sign, orange for pole, and green for vegetation. }
    \label{fig:umap}
\end{figure*}

\noindent\textbf{How to prevent degenerate solution?}
Simultaneously learning the feature and updating the labels during self-training derives degenerate solutions. The key to avoiding such degeneration is to adopt a fixed soft label $p_{t,0}$ as the boilerplate upon which we apply the weight $\omega_t$ for rectification (Equation~\ref{eq:label_reweight}). To explain this, we investigate different variants for the online label updating in Figure~\ref{fig:online_anaylse}, where the labels for self-training can be dynamic $p_t$ or fixed $p_{t,0}$, and the modulation weight can be the network prediction $p_t$ or the confidence $w_t$ estimated according to prototypes. In addition, we investigate the benefit of using hard labels ($\xi(\cdot)$). Figure~\ref{fig:online_anaylse} shows that the performance climbs up for a while and then starts to drop significantly when using non-fixed soft predictions ($p_t$ and its variants), whereas the learning with the fixed ones ($p_{t,0}$ and its variants) steadily improves throughout the training and surpasses the conventional self-training. We conjecture that fixing $p_{t,0}$ makes the refined pseudo labels hardly deviate from this initial prediction, thus avoiding the trivial solution. Moreover, in contrast to using $p_t$ for reweighting, prototypical reweighting improves the mIoU by more than 5.0, corroborating the importance of using prototypes for label denoising. Besides, we observe slight improvement ($\sim$0.2) using hard labels over the soft ones.

\noindent\textbf{The effectiveness of target structure learning.} We propose to cultivate the intrinsic knowledge for the target domain and learns its underlying structure. As shown in Table~\ref{tab:Component Ablation}, without the label denoising, the structure learning improves the performance from 45.6 to 47.6. Note that this is also competitive among self-training approaches, but we do not need to carefully choose the threshold for selecting the confident pseudo labels. The target structure learning assists the pseudo label denoising by forming compact feature clusters and brings the performance gain by 1.4.

\noindent\textbf{The effectiveness of distilling to self-supervised model.}
At the second training stage, we apply knowledge distillation and transfer the dark knowledge of the 1st stage model to the current phase. Table~\ref{tab:Component Ablation} also compares different initialization strategies for this stage. Compared to resuming the last stage training, the initialization from a self-supervised pretrained model (\ie, SimCLRv2) improves the mIoU by~0.6, whereas the initialization with the supervised pretraining degrades the performance. This is because the self-supervised pretraining possesses stronger transferability and can benefit a broad of downstream tasks. The initialization in this way helps the model escape from the local optima in the last stage.
Table~\ref{tab:Component Ablation} also proves the effectiveness of the knowledge distillation: ablating this component drops the mIoU  by 1.1. It is surprising to see that the third stage with the knowledge distillation further improves the performance by 0.6, attaining the 20.9 mIoU improvement relative to the model without domain adaptation.

\noindent\textbf{The UMAP visualization of target features.}
To better develop intuition, we visualize the learned features for ProDA in Figure~\ref{fig:umap}. The model before the domain adaptation mixes the features of the same class. The conventional self-training could produce more separated feature space, yet it is still hard for linear classification. When we apply prototypical pseudo label denoising, features among different classes are better separated, though the distribution is still dispersed. In comparison, the full model gives the most compact feature clusters that are amenable to classification.

\subsection{Parameter sensitivity analysis.}
To showcase that ProDA is robust to the hyper-parameter selection, we analyze the impact of parameters.  In Table~\ref{tab:threshold in prototypical online label refinement}, we use different threshold values to select the pseudo labels and the performance is not sensitive to this threshold as opposed to the conventional self-training. Hence, ProDA does not apply thresholding for convenient usage. We also study the effect of momentum value when online computing the prototypes, and ProDA is robust to a wide numerical range as shown in Table~\ref{tab:prototype momentum}.

\begin{table}
\centering
\footnotesize
\begin{tabular}{c|ccccccc}
\midrule
threshold & 0.0 & 0.2 & 0.4 & 0.6 & 0.8 & 0.9 & 0.95\\
\midrule
 mIoU & 53.7 & 53.8 & 53.7 & 53.7 & 53.8 & 53.7 & 53.3\\
\midrule
\end{tabular}
\caption {The effect of threshold in pseudo label selection during the prototypical label denoising.}
\label{tab:threshold in prototypical online label refinement}
\end{table}

\begin{table}
\centering
\footnotesize
\begin{tabular}{c|cccc}
\midrule
momentum & 0.99 & 0.999 & 0.9999 & 0.99999 \\
\midrule
 mIoU & 53.5 & 53.6 & 53.7 & 52.3\\
\midrule
\end{tabular}
\caption {The effect of prototype momentum during the prototype online update.}
\label{tab:prototype momentum}
\end{table}

\section{Conclusions}
In this paper, we propose ProDA which resorts to prototypes to online denoise the pseudo labels and learn the compact feature space for the target domain. Knowledge distillation to a self-supervised pretrained model further boosts the performance. The proposed method outperforms state-of-the-art methods by a large margin, greatly reducing the gap with supervised learning. We will make the code publicly available. We believe the proposed method is a general amelioration to the self-training and we hope to explore its capability in other tasks in the future.  

% \section{TODO}
% \begin{itemize}
%     \itemsep=0em
%     \item name the method
% \end{itemize}

\clearpage

{\small
\bibliographystyle{ieee_fullname}
\bibliography{egbib}
}

\cleardoublepage
\onecolumn

\renewcommand\thesubsection{Appendix \Alph{subsection}}

\subsection{Influences of Design Choices}
\noindent\textbf{Prototype initialization strategy.} 
In our implementation, the proposed ProDA initializes the prototypes of the target domain according to the pseudo predictions for the target images. Alternatively, the target prototypes can also be initialized according to the ground truth labeling in the source domain. However, both choices have their pros and cons: the former suffers from the noises in the pseudo labels whereas the latter suffers from domain gap as the prototypes of the two domains may not accurately align.  Table~\ref{tab:prototype initialization} shows that the two initialization strategies induce comparable results, as the prototypes are online updated and can rapidly converge to the true cluster centroids. The quantitative performance is measured on the dataset GTA5 $\rightarrow$ Cityscapes, whereas the other dataset shows similar results.
\begin{table}[h!]
\centering
\small
\begin{tabular}{@{}ccc@{}}
\toprule
 & source ground truth &  target pseudo label\\
\midrule
 mIoU & 53.6 & 53.7\\
\bottomrule
\end{tabular}
\caption {The performance of different target prototype initialization strategies. Here we only report the performance for the 1st training stage in the gta5 $\to$ Cityscapes task.}
\label{tab:prototype initialization}
\end{table}

\noindent\textbf{Strong augmentation.} In the target structure learning, we take weak and strong augmentation views for the target image. We employ random crop for weak augmentation and explore the effects of different augmentation types for the strong augmented view. As shown in Table~\ref{tab:augmentation}, random crop only gives the mIoU score 52.7, whereas adding RandAugment~\cite{cubuk2020randaugment} and CutOut~\cite{devries2017improved} respectively improve the mIoU by 0.78 and 0.5. The strongest augmentation gives the best performance, indicating the importance of data augmentation when learning the compact feature space for the target domain. 
\begin{table}[h!]
\centering
\small
\begin{tabular}{@{}ccccc@{}}
\toprule
 & crop & {crop \& RandAug} &  {crop \& Cutout} & {crop \& RandAug \& Cutout}\\
\midrule
mIoU & 52.7 & 53.5 & 53.2 & 53.7\\
\bottomrule
\end{tabular}
\caption {The influence of various strong augmentations. Here we only report the performance for the 1st training stage in the gta5 $\to$ Cityscapes task.}
\label{tab:augmentation} 
\end{table}

\noindent\textbf{Effect of temperature during prototypical denoising.} 
We rely on the prototypical context to rectify the pseudo labels. We compute the softmax over feature distance to all the prototypes, and the softmax temperature $\tau$ influences the denoising effect and requires balancing: when $\tau \to 0$, only the nearest prototype dominates whereas $\tau \to \infty$ causes that all the prototypes are accounted equally. The influence of the temperature is shown in Table~\ref{tab:prototype temperature}. We empirically set $\tau=1$ in our experiments.
\begin{table}[h!]
\centering
\small
\begin{tabular}{@{}cccccccc@{}}
\toprule
 & 0.1 & 0.5 & 1 & 2 & 3 & 5 & 10\\
\midrule
 mIoU & 48.8 & 52.1 & 53.7 & 51.9 & 47.5 & 44.9 & 40.9\\
\bottomrule
\end{tabular}
\caption {The effects of temperature during the prototypical denoising. Here we only report the performance for the 1st training stage in the gta5 $\to$ Cityscapes task.}
\label{tab:prototype temperature}
\end{table}

\noindent\textbf{Symmetric cross-entropy loss.} We employ the symmetric cross-entropy loss (SCE) for robust learning to stabilize the early training phase. The SCE has coefficients $\alpha$ and $\beta$ that balance the cross-entropy and the reverse cross-entropy. Table~\ref{tab:sce} shows that the final result is not sensitive to these hyper-parameters if $\beta$ is not too small. Here, we follow the suggested setting  as~\cite{wang2019symmetric}, 
\ie, $\alpha=0.1, \beta=1$.

\begin{table}[h!]
\centering
\small
\setlength\tabcolsep{8.0pt}
\begin{tabular}{c|cccc}
\toprule
\diagbox{$\alpha$}{$\beta$} & 0.1 & 0.5 & 1 & 5\\
\midrule
0.01 & 46.4 & 52.7 & 53.8  & 53.6\\
\midrule
0.1 & 47.6 & 52.9 & 53.7 & 53.5\\
\midrule
0.5 & 50.4 & 53.1 & 53.3 & 53.5\\
\midrule
1 & 51.1 & 52.7 & 53.1 & 53.6\\
\bottomrule
\end{tabular}
\caption {The influence of $\alpha$ and $\beta$ in the symmetric cross-entropy (SCE) loss. Here we only report the performance for the 1st training stage in the gta5 $\to$ Cityscapes task.}
\label{tab:sce} 
\end{table}

\noindent\textbf{The effect of loss weight.} Table~\ref{tab:loss weight} shows that the final result is not sensitive to the KL loss weight ($\gamma_1$) and the regularization loss weight ($\gamma_2$). In GTA5 $\rightarrow$ Cityscapes, we set $\gamma_1 = 10$ and $\gamma_1 = 0.1$, while in SYNTHIA $\rightarrow$ Cityscapes, we set $\gamma_1 = 10$ and $\gamma_1 = 0$.
\begin{table}[h!]
\centering
\small
\setlength\tabcolsep{8.0pt}
\begin{tabular}{c|ccc}
\toprule
\diagbox{$\gamma_1$}{$\gamma_2$} & 0.02 & 0.1 & 0.2\\
\midrule
2 & 52.9 & 53.7 & 53.5\\
\midrule
10 & 53.2 & 53.7 & 53.4\\
\midrule
20 & 53.4 & 53.6 & 52.1\\
\midrule
50 & 53.6 & 52.0 & 52.1\\
\bottomrule
\end{tabular}
\caption {The influence of the KL loss weight ($\gamma_1$) and the regularization loss weight ($\gamma_2$). Here we only report the performance for the 1st training stage in the gta5 $\to$ Cityscapes task.}
\label{tab:loss weight} 
\end{table}

\clearpage
\subsection{Algorithm}
The training procedure of our ProDA is summarized in Algorithm~\ref{ProDA}, which is composed of three stages. The first stage consists of prototypical pseudo label denoising and target structure learning. In the second and third stages, we apply knowledge distillation to a self-supervised model. For detailed equations and loss functions, please refer to our main paper.

\SetCommentSty{mycommfont}
\SetKwInput{KwInput}{Input}
\SetKwInput{KwOutput}{Output}
\begin{algorithm}[tbh]
\small
\DontPrintSemicolon
\KwInput{training dataset: ($\cX_s,\cY_s,\cX_t$); prototype momentum: $\lambda$; weak, strong augmentations: $\cT$, $\cT'$; the pretrained SimCLRv2 model: $h^{'}_\theta$; pseudo label selection threshold: $T$;}
\KwOutput{the output model $h_\theta$.}
Warmup: $h_\theta= g_\theta \circ f_\theta \leftarrow (\cX_s,\cY_s,\cX_t)$ according to \cite{Tsai_adaptseg_2018};\;
Generate soft pseudo label: $p_{t,0} \leftarrow h_\theta(\cX_t)$;\;
Prototype initialization: $\eta_c \leftarrow (f_\theta,\cX_t)$;\;
EMA model initialization: $\tilde{h}_\theta \leftarrow h_\theta$;\;
\For{$m\gets0$ \KwTo \text{epochs}}{
    \For{$i\gets0$ \KwTo len($X_t$)}{
          Get source images $x_s^{(i)}$;\;
          Train the model $h_\theta$ using loss $\ell_{ce}^s$;\;
          \;
          Get target images $x_t^{(i)}$;\;
          Calculate the denoising weight $\omega_t^{(i,k)}$;\;
          Update the pseudo label $\hat{y}^{(i,k)}_t$;\;
          Train model $h_\theta$ using loss $\ell^t_{sce}$;\;
          \;
          Calculate the soft label $z_{\cT}, z_{\cT'}$;\;
          Train the model $h_\theta$ using loss $\ell_{kl}^t$ and $\ell_{reg}^t$;\;
          \;
          Calculate the batch prototype $\eta_c'$ ;\;%according to Eq.~\ref{eqn:proto_iter};\;
          $\eta_c \leftarrow \lambda \eta_c + (1-\lambda) \eta_c'$;\;
          Update the EMA model $\tilde{h}_\theta$;\;
    }
}
\;
\For{$\text{stage}\gets1$ \KwTo $2$}{
Generate the pseudo label: $\hat{y}_t \leftarrow \xi( h_\theta(\cX_t), T)$;\;
Student model initialization: $h^{\dagger}_\theta \leftarrow h^{'}_\theta$;\;
\For{$m\gets0$ \KwTo \text{epochs}}{
    \For{$i\gets0$ \KwTo len($X_t$)}{
          Get source images $x_s^{(i)}$;\;
          Tune the model $h^{\dagger}_\theta$ using loss $\ell_{ce}^s$;\;
          \;
          Get target images $x_t^{(i)}$;\;
          Calculate the teacher probability $h_\theta(x_t^{(i)})$;\;
          Calculate the student probability $h^{\dagger}_\theta(x_t^{(i)})$;\;
          Tune the model $h^{\dagger}_\theta$ using loss $\ell_{ce}^t$ and KL loss;\;
    }
}
$h_\theta \leftarrow h^{\dagger}_\theta$;\;
}
\caption{ProDA}
\label{ProDA}
\end{algorithm}

\clearpage
\subsection{Detailed Ablation study}
Here we show a detailed ablation study for all the 19 classes on GTA5 $\to$ Cityscapes. 

\begin{table*}[h]
\centering
\footnotesize
\setlength\tabcolsep{1.8pt}{
\begin{tabular}{@{}c|cccc|*{19}{c}|cc@{}}
\toprule 
&\multicolumn{4}{c|}{components}&\multicolumn{1}{c}{\begin{sideways}road\end{sideways}} & \multicolumn{1}{c}{\begin{sideways}sideway\end{sideways}} & \multicolumn{1}{c}{\begin{sideways}building\end{sideways}} & \multicolumn{1}{c}{\begin{sideways}wall\end{sideways}} & \multicolumn{1}{c}{\begin{sideways}fence\end{sideways}} & \multicolumn{1}{c}{\begin{sideways}pole\end{sideways}} & \multicolumn{1}{c}{\begin{sideways}light\end{sideways}} & \multicolumn{1}{c}{\begin{sideways}sign\end{sideways}} & \multicolumn{1}{c}{\begin{sideways}vege.\end{sideways}} & \multicolumn{1}{c}{\begin{sideways}terrace\end{sideways}} & \multicolumn{1}{c}{\begin{sideways}sky\end{sideways}} & \multicolumn{1}{c}{\begin{sideways}person\end{sideways}} & \multicolumn{1}{c}{\begin{sideways}rider\end{sideways}} & \multicolumn{1}{c}{\begin{sideways}car\end{sideways}} & \multicolumn{1}{c}{\begin{sideways}truck\end{sideways}} & \multicolumn{1}{c}{\begin{sideways}bus\end{sideways}} & \multicolumn{1}{c}{\begin{sideways}train\end{sideways}} & \multicolumn{1}{c}{\begin{sideways}motor\end{sideways}} & \multicolumn{1}{c}{\begin{sideways}bike\end{sideways}} &\multicolumn{1}{|l}{mIoU} & \multicolumn{1}{l}{gain}\\
\midrule
\multirow{2}{*}{init.}&\multicolumn{4}{c|}{source}&75.8 & 16.8 & 77.2 & 12.5 & 21.0 & 25.5 & 30.1 & 20.1 & 81.3 & 24.6 & 70.3 & 53.8 & 26.4 & 49.9 & 17.2 & 25.9 & 6.5 & 25.3 & 36.0 & 36.6 &\\
%\cline{1-1} \cline{3-22} 
&\multicolumn{4}{c|}{warm up} & 86.7&34.2&79.3&26.6&21.6&38.4&33.7&15.8&82.1&31.0&73.2&60.4&21.0&82.3&23.2&32.0&2.9&24.1&20.9& 41.6 & +5.0\\
\midrule

\multirow{6}{*}{stage 1}&\tabincell{c}{ST} & SCE &  \tabincell{c}{PD} &  \tabincell{c}{SL}&&&&&&&&&&&&&&&&&&&&\multicolumn{1}{l}{mIoU} & \multicolumn{1}{l}{gain}\\
%\midrule
\cline{2-26} 
&\checkmark & &  & & 87.0&39.7&77.5&31.5&25.7&41.5&38.7&20.3&84.6&38.2&74.1&63.7&21.7&86.0&29.0&37.5&0.3&34.9&26.2& 45.2 & +8.6\\
&\checkmark & \checkmark &  & & 87.7&36.8&78.2&30.9&24.8&41.5&40.0&23.2&83.0&35.3&72.9&64.1&24.6&85.9&32.9&36.5&2.0&31.0&35.0&45.6 & +9.0\\
&\checkmark&\checkmark & \checkmark  & & 93.2&56.7&84.1&40.4&37.5&39.5&44.1&35.1&87.1&43.2&80.3&65.8&29.8&87.7&29.6&41.9&0.0&44.4&52.6&52.3 & +15.7\\
%\cline{1-1} \cline{3-22}    
&\checkmark&\checkmark &  & \checkmark &89.0&38.6&80.7&37.1&27.2&42.8&41.5&20.7&85.8&42.4&74.8&64.8&17.8&87.6&30.8&39.4&0.0&41.0&34.6&47.6 & +11.0\\
%\cline{1-1} \cline{3-22}    
&\checkmark&\checkmark & \checkmark & \checkmark & 91.5&52.4&82.9&42.0&35.7&40.0&44.4&43.3&87.0&43.8&79.5&66.5&31.4&86.7&41.1&52.5&0.0&45.4&53.8&53.7 & +17.1\\
\midrule  
\multirow{5}{*}{stage 2}& \tabincell{c}{self\\distill.} & \tabincell{c}{stage 1\\init.} & \tabincell{c}{sup\\init.} & \tabincell{c}{self-sup\\init.} &&&&&&&&&&&&&&&&&&&&\multicolumn{1}{l}{mIoU} & \multicolumn{1}{l}{gain}\\
%\midrule  
\cline{2-26} 
&& & & \checkmark &90.0&57.4&81.8&42.0&40.2&43.8&50.3&50.9&87.6&42.6&80.0&69.2&32.9&87.8&45.5&56.9&0.0&46.0&55.4& 55.8 & +19.2\\
&\checkmark&\checkmark & & &91.4&53.3&83.4&41.3&37.8&43.9&53.0&47.9&88.3&46.1&79.9&70.5&33.2&89.0&48.4&54.6&0.0&50.5&56.7& 56.3 & +19.7\\
&\checkmark& & \checkmark &  &91.0&50.2&83.1&40.1&39.8&43.5&51.9&48.1&87.9&45.9&78.5&69.6&34.3&87.9&41.3&56.6&0.0&51.7&57.1& 55.7 & +19.1\\
&\checkmark& & & \checkmark &89.4&56.5&81.3&46.3&42.7&45.1&52.2&51.3&88.5&47.0&82.9&69.3&36.5&87.4&46.0&57.5&0.6&45.9&54.5& 56.9 & +20.3\\
\midrule 
stage 3 & \checkmark & & & \checkmark &87.8&56.0&79.7&46.3&44.8&45.6&53.5&53.5&88.6&45.2&82.1&70.7&39.2&88.8&45.5&59.4&1.0&48.9&56.4& 57.5 & +20.9\\
\bottomrule  
\end{tabular}%
}
\caption {Ablation study of each proposed component. The whole training involves three stages, where knowledge distillation can be applied in the last two stages. Here, {ST} stands for self-training, {PD} for prototypical denoising, and {SL} for structure learning.}
\label{tab:Component Ablation}%
\end{table*}%

\clearpage
\subsection{Qualitative  comparison}
% Figure~\ref{figure:compare_baseline} shows the qualitative comparison between the proposed ProDA and the baseline methods. Figure~\ref{figure:compare1} and Figure~\ref{figure:compare2} give two comparison examples between ProDA and other state-of-the-art methods.
\begin{figure}[h!]
    \center
    \small
    \setlength\tabcolsep{1pt}
    {
    \newcolumntype{P}[1]{>{\centering\arraybackslash}p{#1}}
    \begin{tabular}{@{}*{10}{P{0.0974\columnwidth}}@{}}
         {\cellcolor[rgb]{0.5,0.25,0.5}}\textcolor{white}{road} &{\cellcolor[rgb]{0.957,0.137,0.91}}sidewalk &{\cellcolor[rgb]{0.275,0.275,0.275}}\textcolor{white}{building} &{\cellcolor[rgb]{0.4,0.4,0.612}}\textcolor{white}{wall} &{\cellcolor[rgb]{0.745,0.6,0.6}}fence &{\cellcolor[rgb]{0.6,0.6,0.6}}pole &{\cellcolor[rgb]{0.98,0.667,0.118}}traffic light&{\cellcolor[rgb]{0.863,0.863,0}}traffic sign &{\cellcolor[rgb]{0.42,0.557,0.137}}vegetation & {\cellcolor[rgb]{0,0,0}}\textcolor{white}{n/a.}\\
         
         {\cellcolor[rgb]{0.596,0.984,0.596}}terrain &{\cellcolor[rgb]{0,0.51,0.706}}sky &{\cellcolor[rgb]{0.863,0.078,0.235}}\textcolor{white}{person} &{\cellcolor[rgb]{1,0,0}}\textcolor{white}{rider} &{\cellcolor[rgb]{0,0,0.557}}\textcolor{white}{car} &{\cellcolor[rgb]{0,0,0.275}}\textcolor{white}{truck} &{\cellcolor[rgb]{0,0.235,0.392}}\textcolor{white}{bus}&{\cellcolor[rgb]{0,0.314,0.392}}\textcolor{white}{train} &{\cellcolor[rgb]{0,0,0.902}}\textcolor{white}{motorcycle} & {\cellcolor[rgb]{0.467,0.043,0.125}}\textcolor{white}{bike}\\
    
    \end{tabular}
    
    \renewcommand{\arraystretch}{0.3}
    \begin{tabular}{@{}cccc@{}}
    
        \includegraphics[width=0.25\columnwidth]{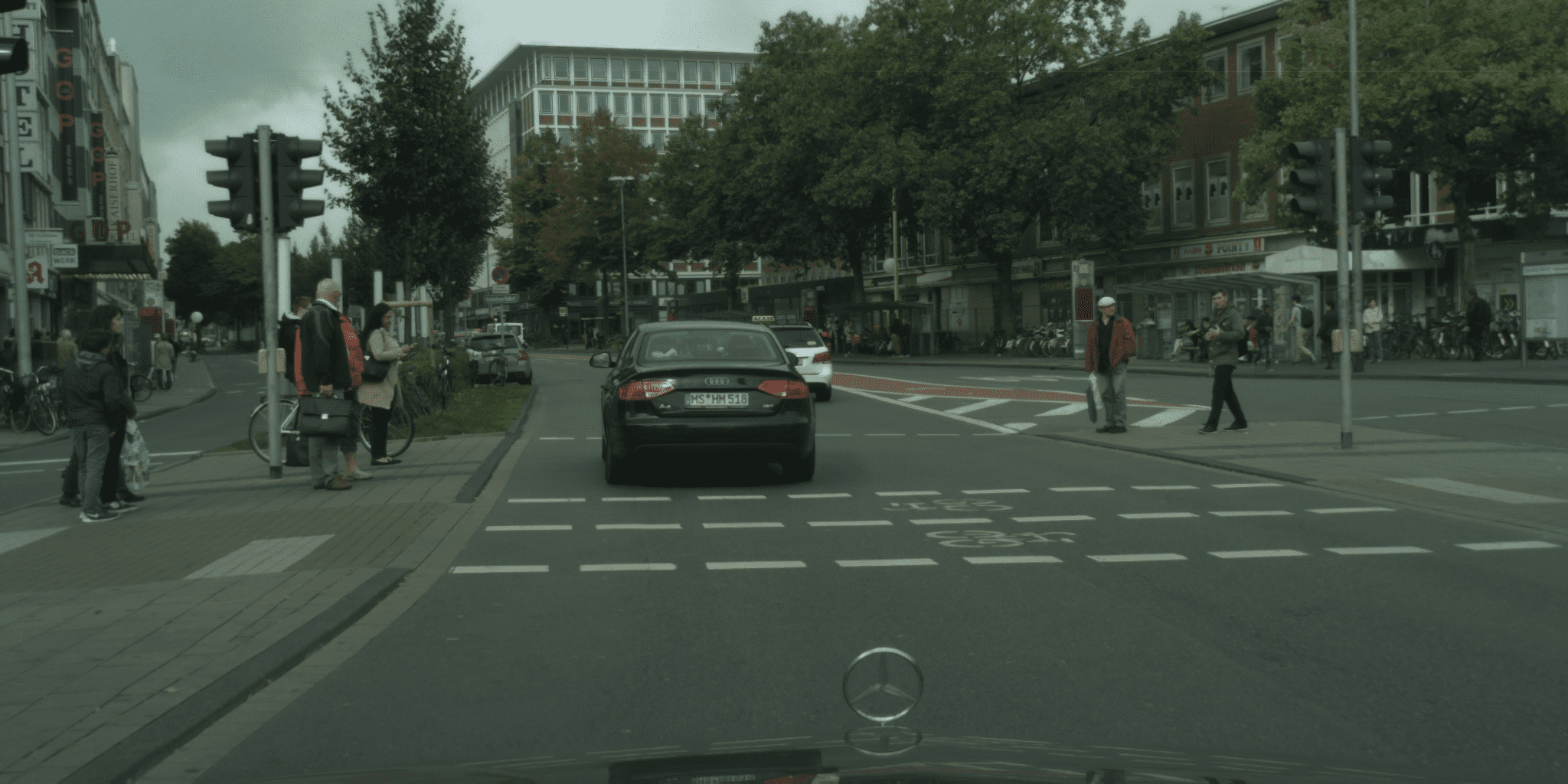}&
        \includegraphics[width=0.25\columnwidth]{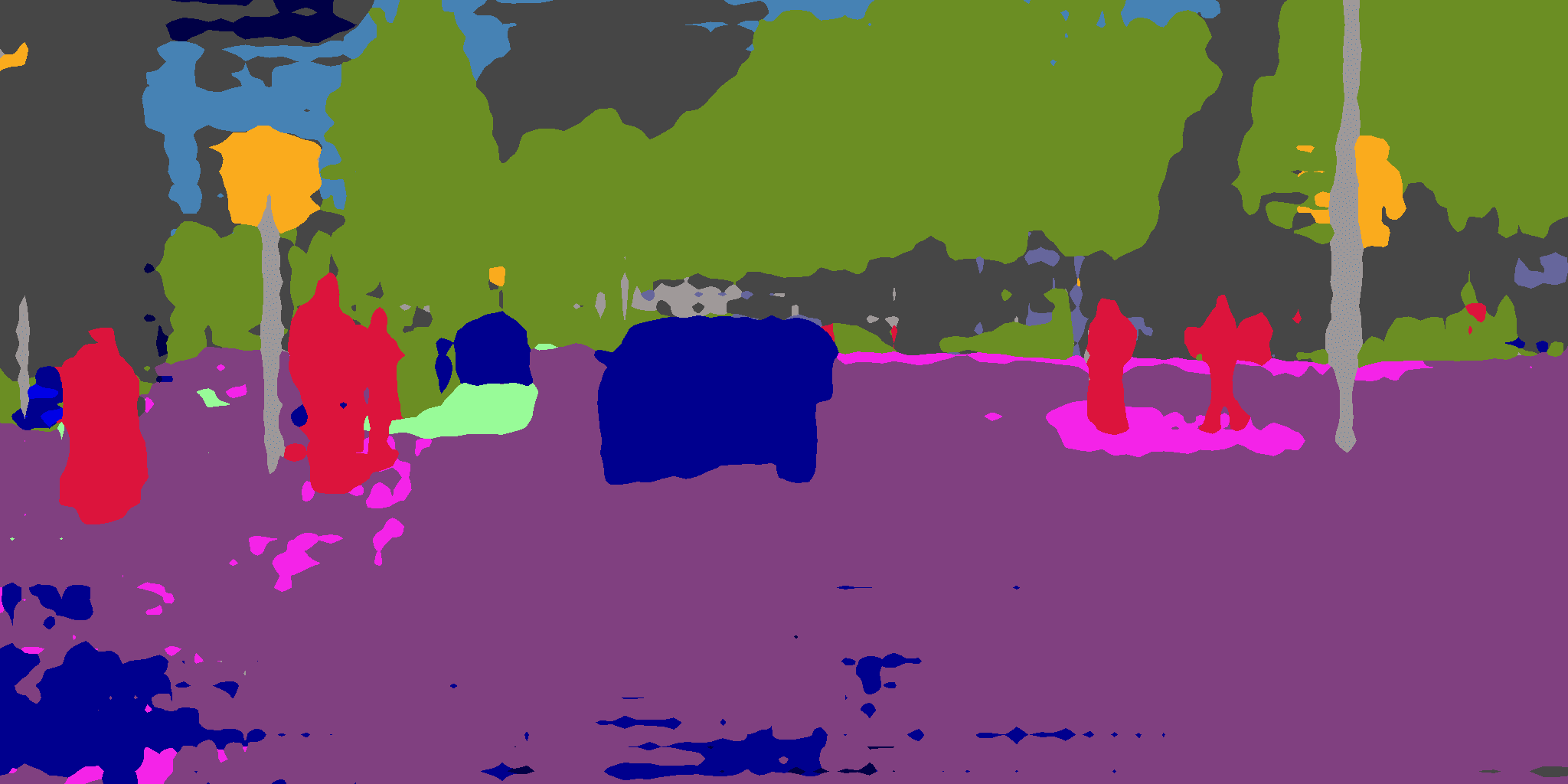}&
        \includegraphics[width=0.25\columnwidth]{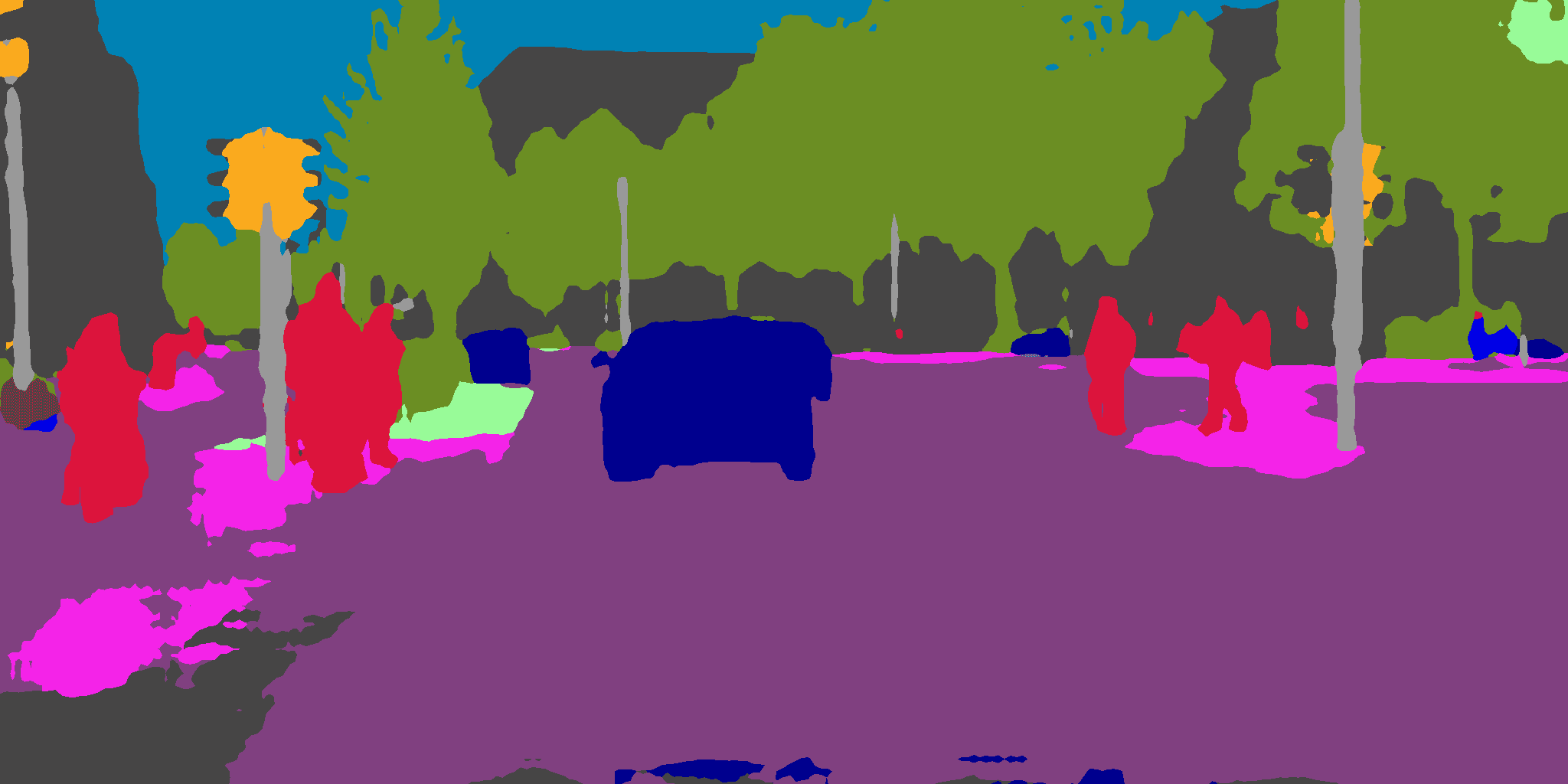}&
        \includegraphics[width=0.25\columnwidth]{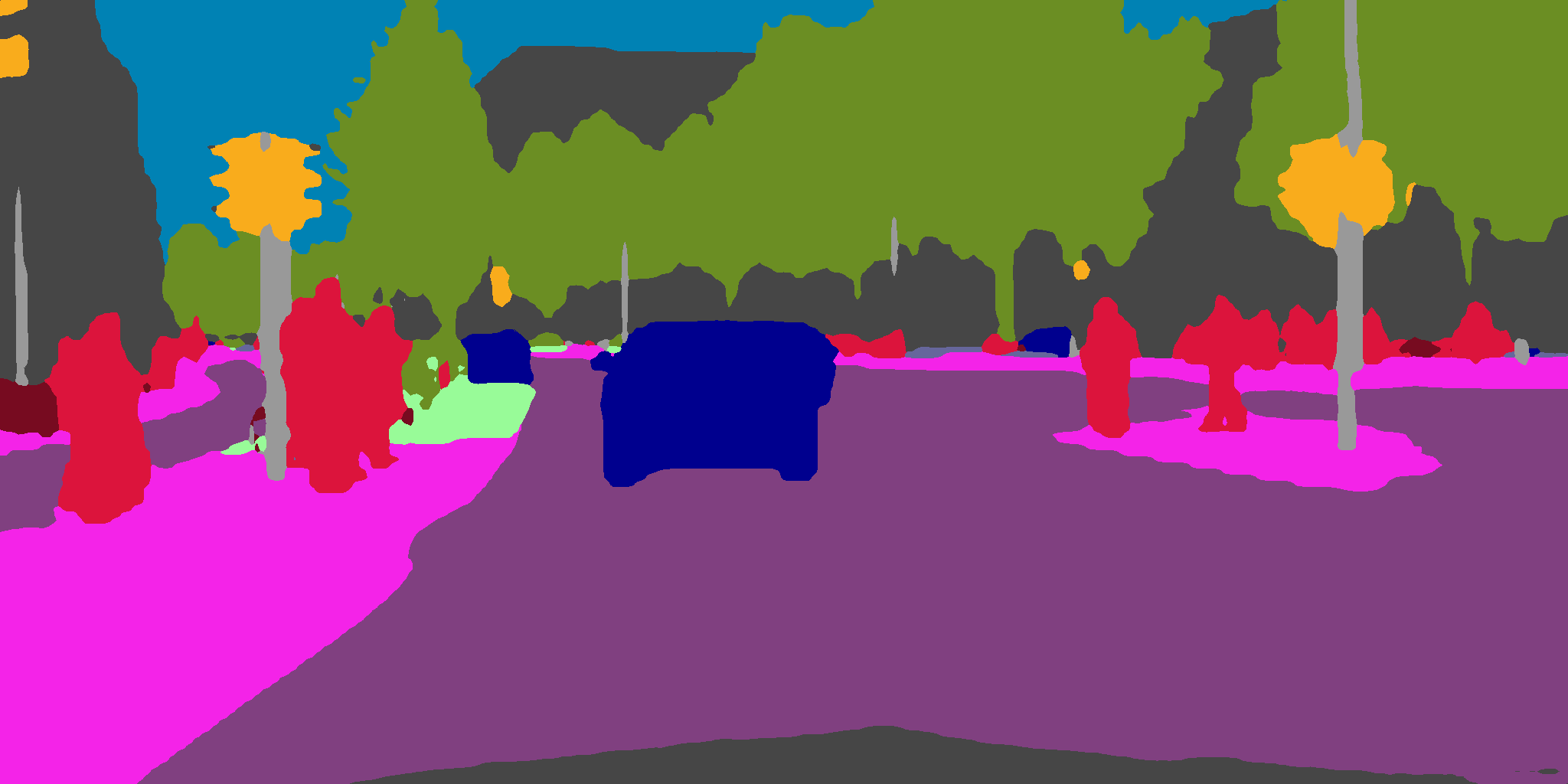}\\
    
        \includegraphics[width=0.25\columnwidth]{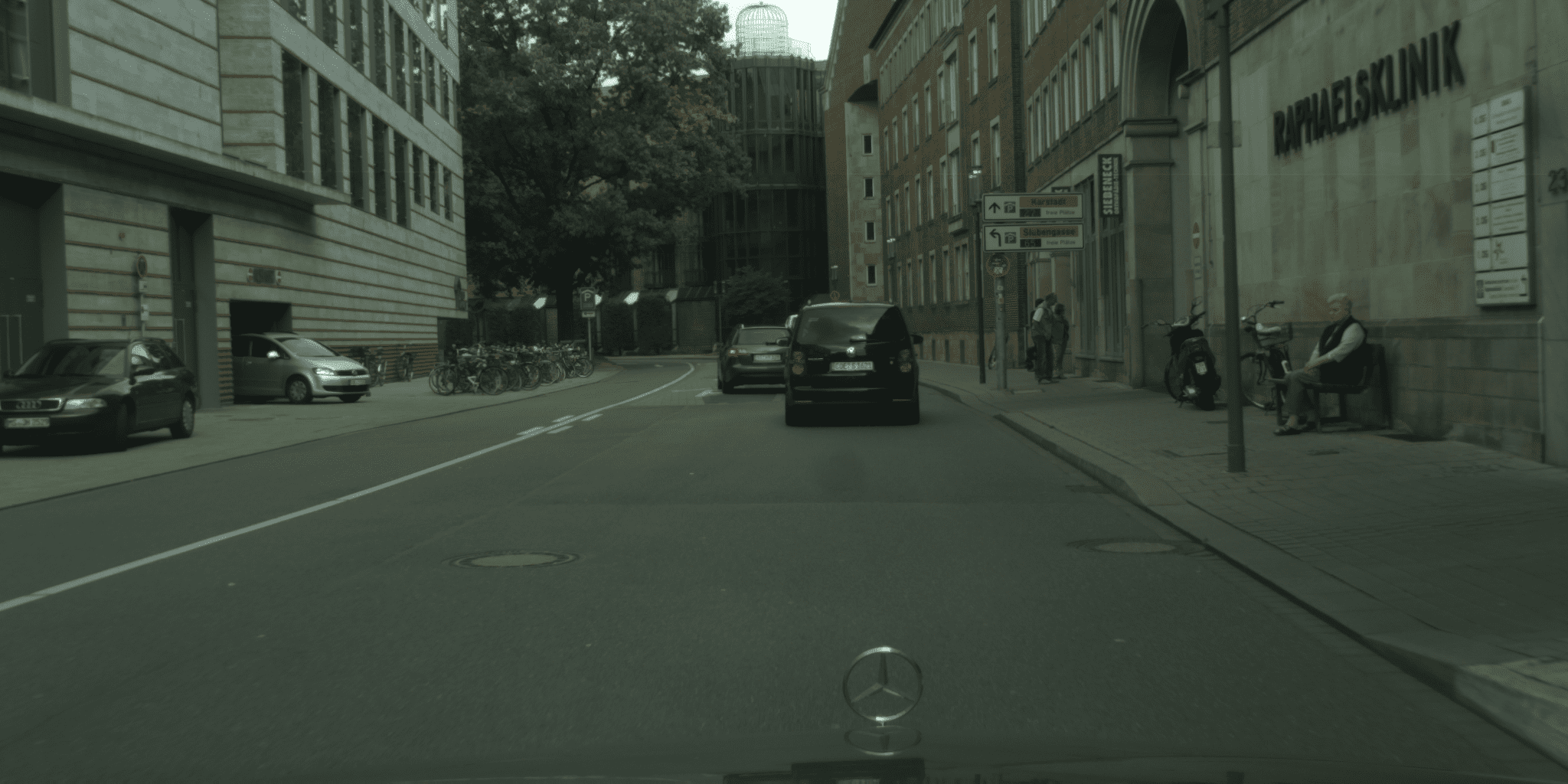}&
        \includegraphics[width=0.25\columnwidth]{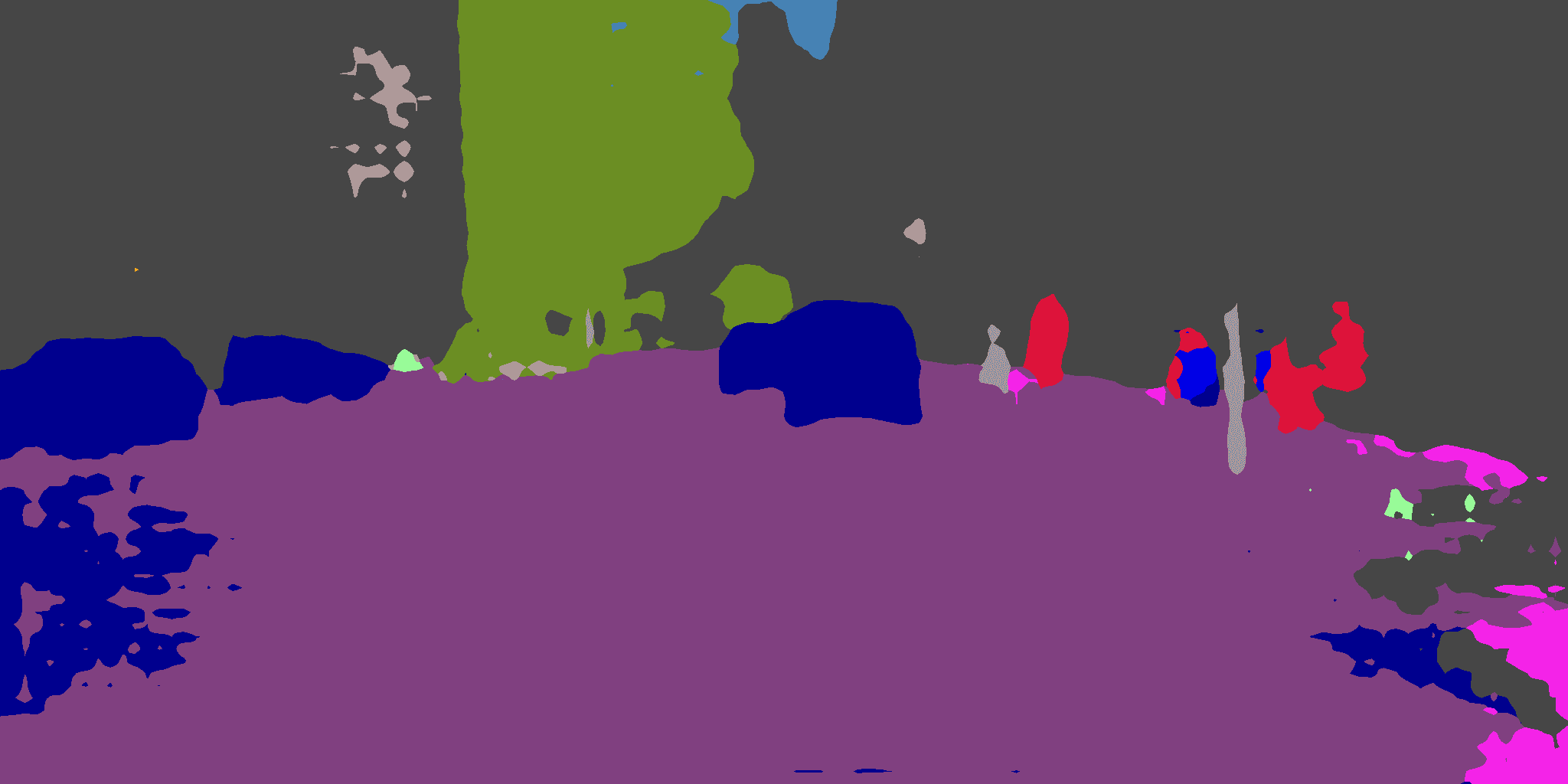}&
        \includegraphics[width=0.25\columnwidth]{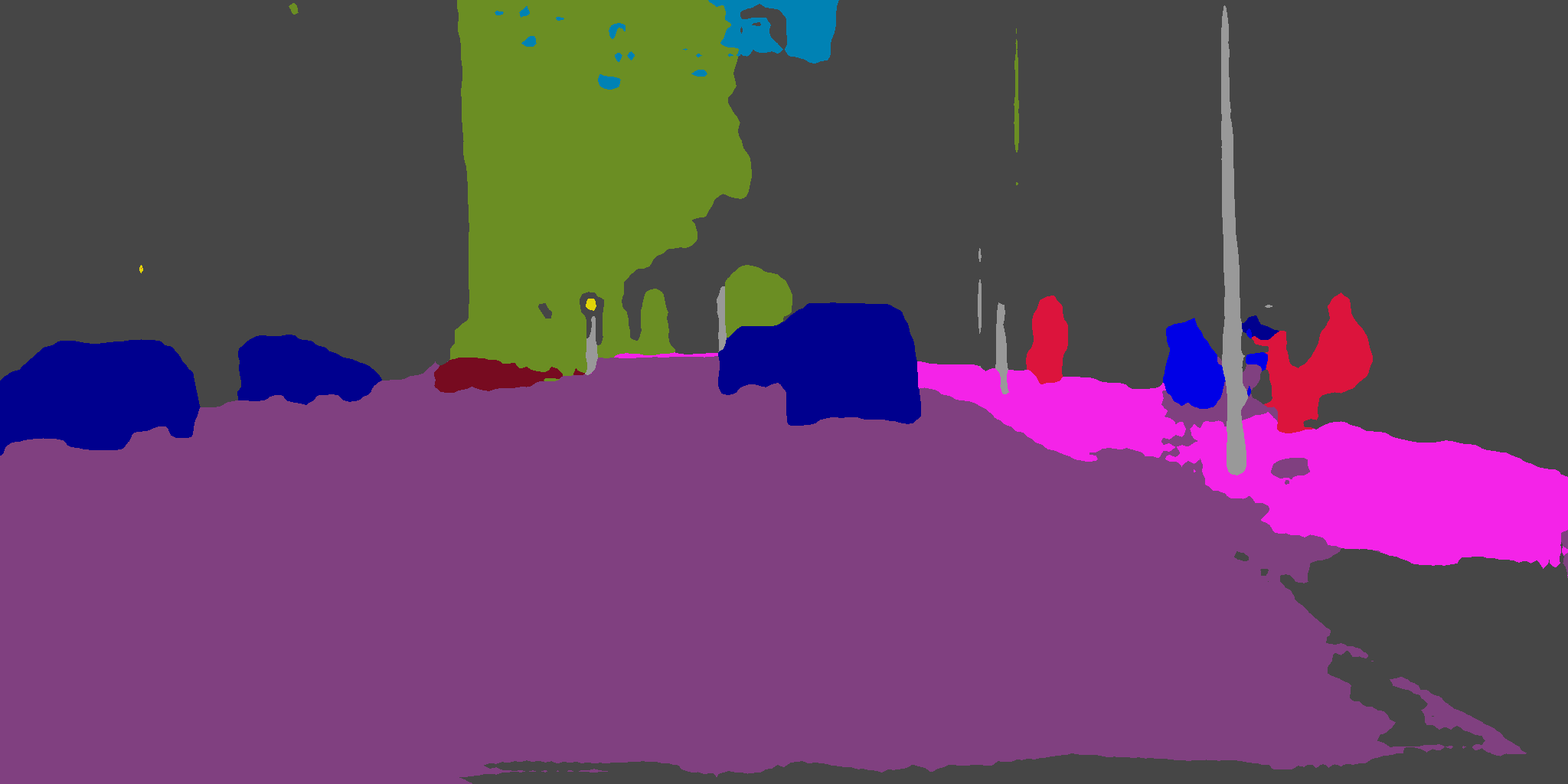}&
        \includegraphics[width=0.25\columnwidth]{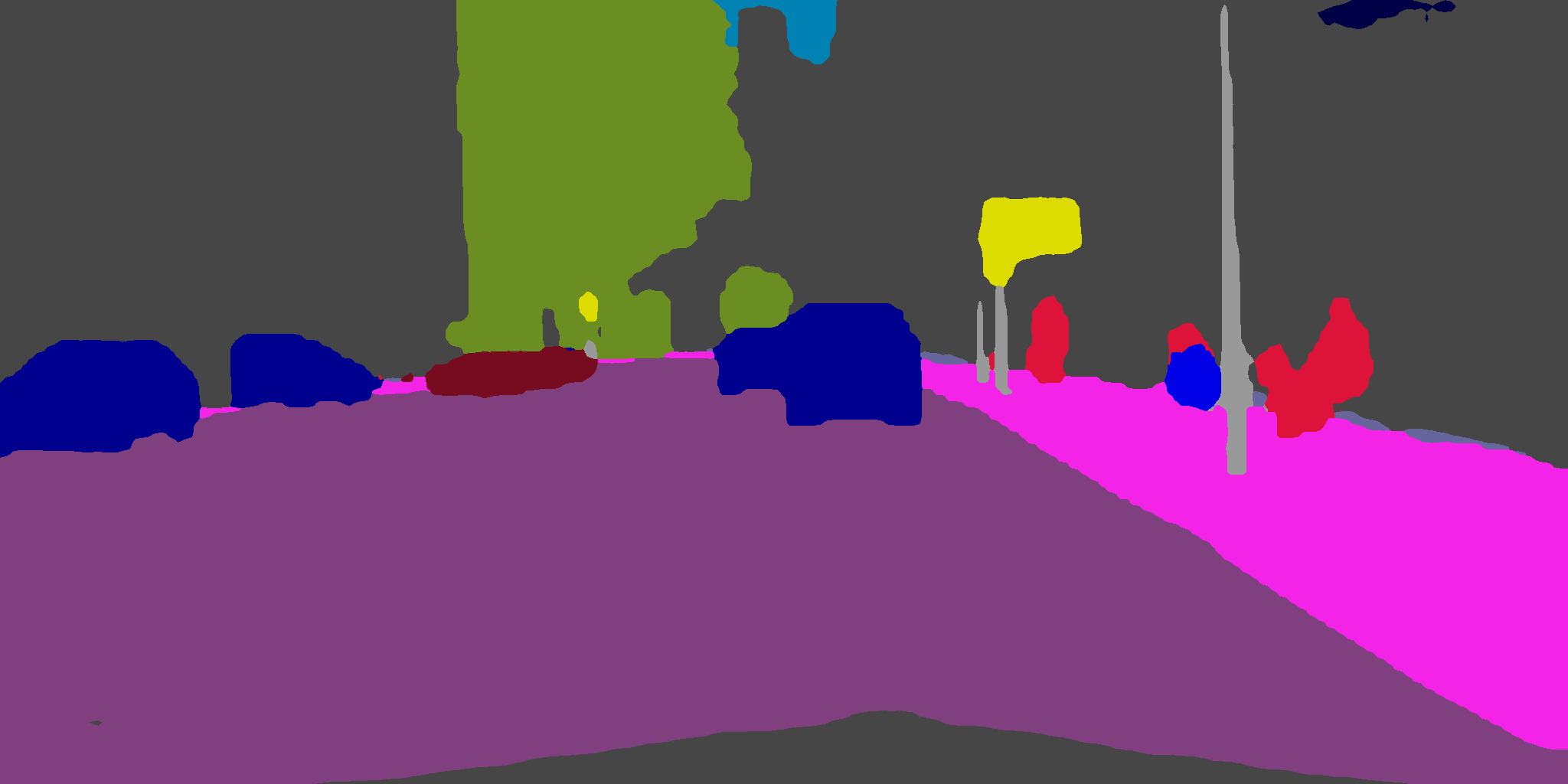}\\

        \includegraphics[width=0.25\columnwidth]{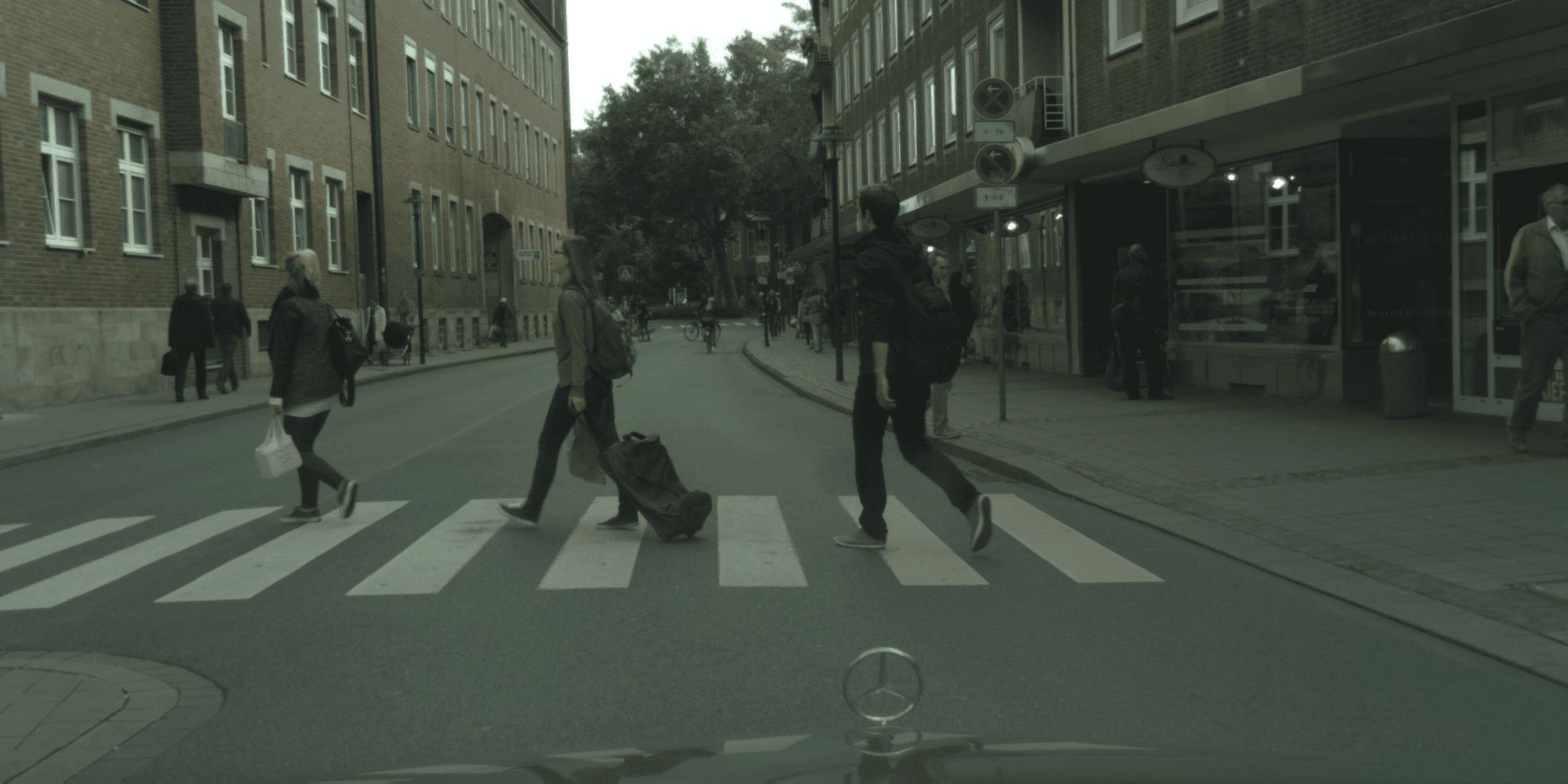}&
        \includegraphics[width=0.25\columnwidth]{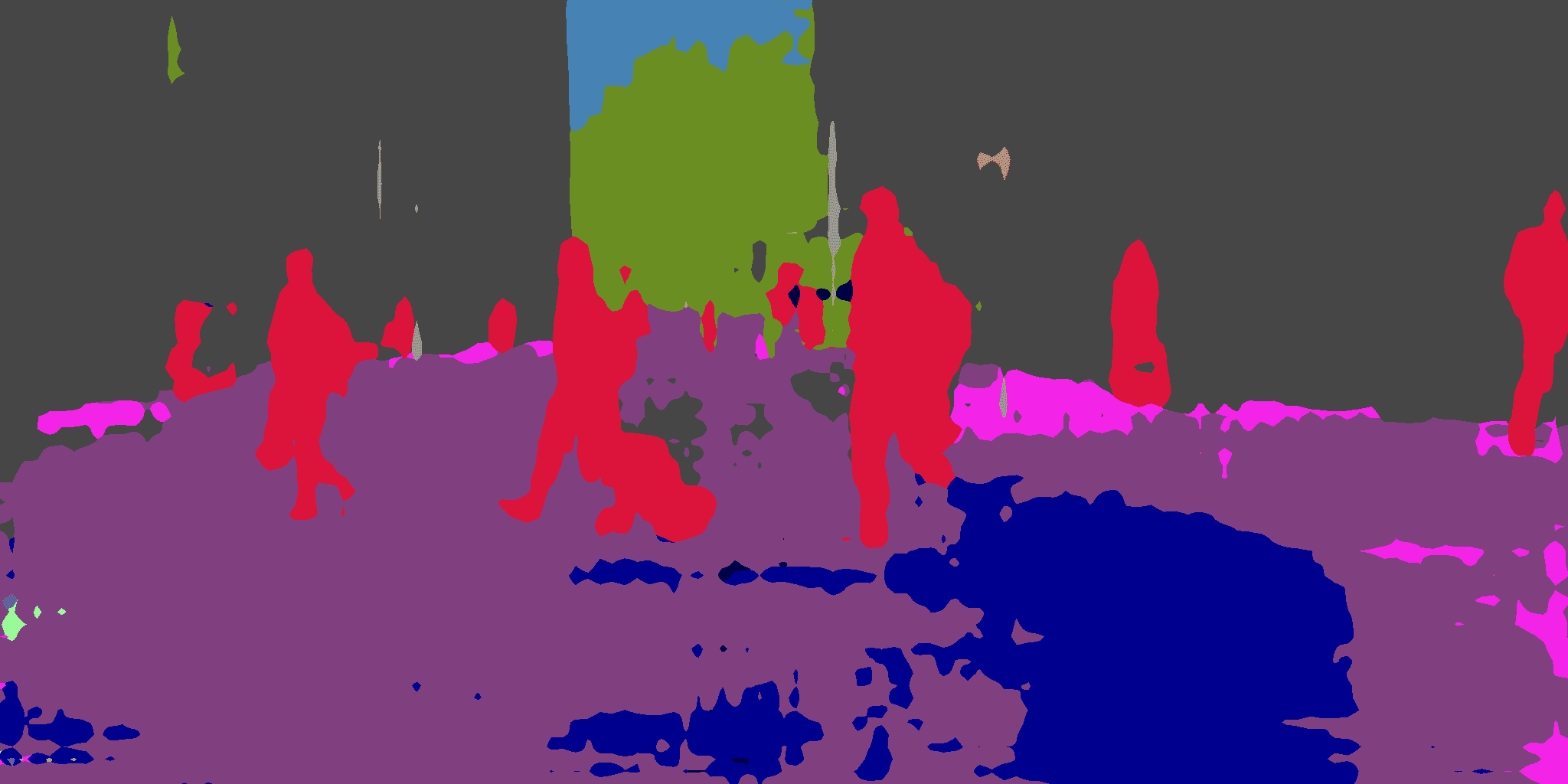}&
        \includegraphics[width=0.25\columnwidth]{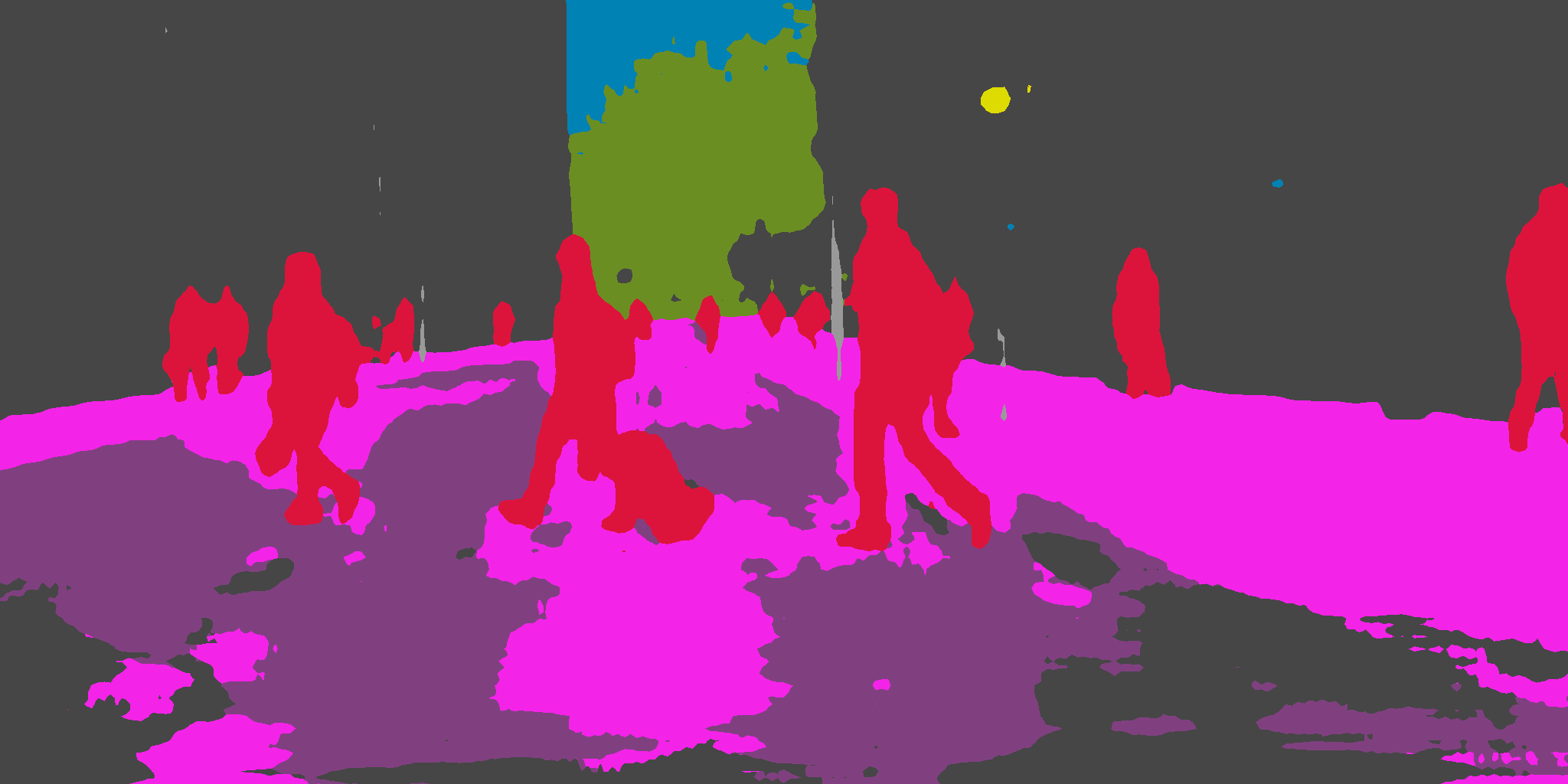}&
        \includegraphics[width=0.25\columnwidth]{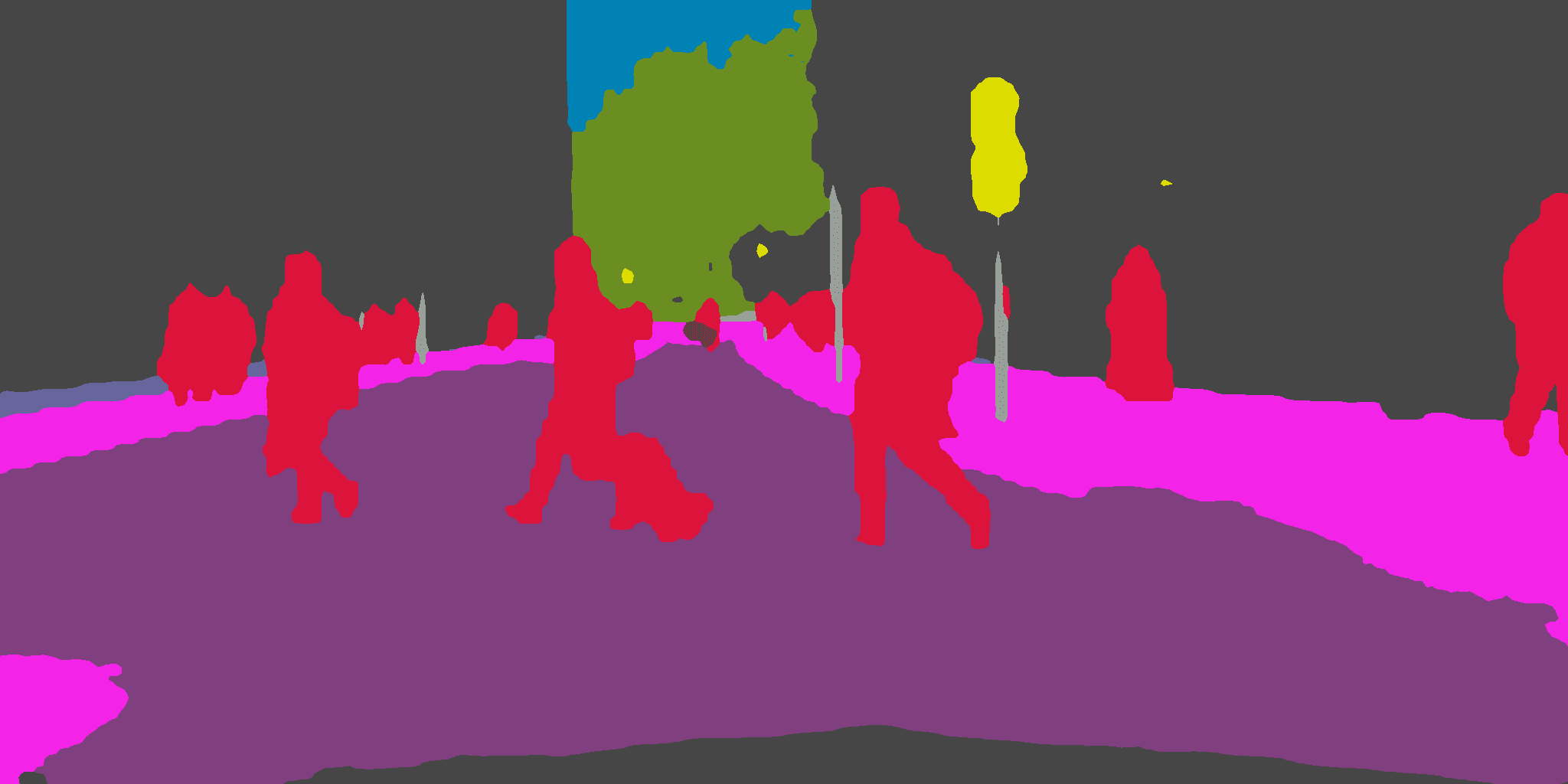}\\
        
        \includegraphics[width=0.25\columnwidth]{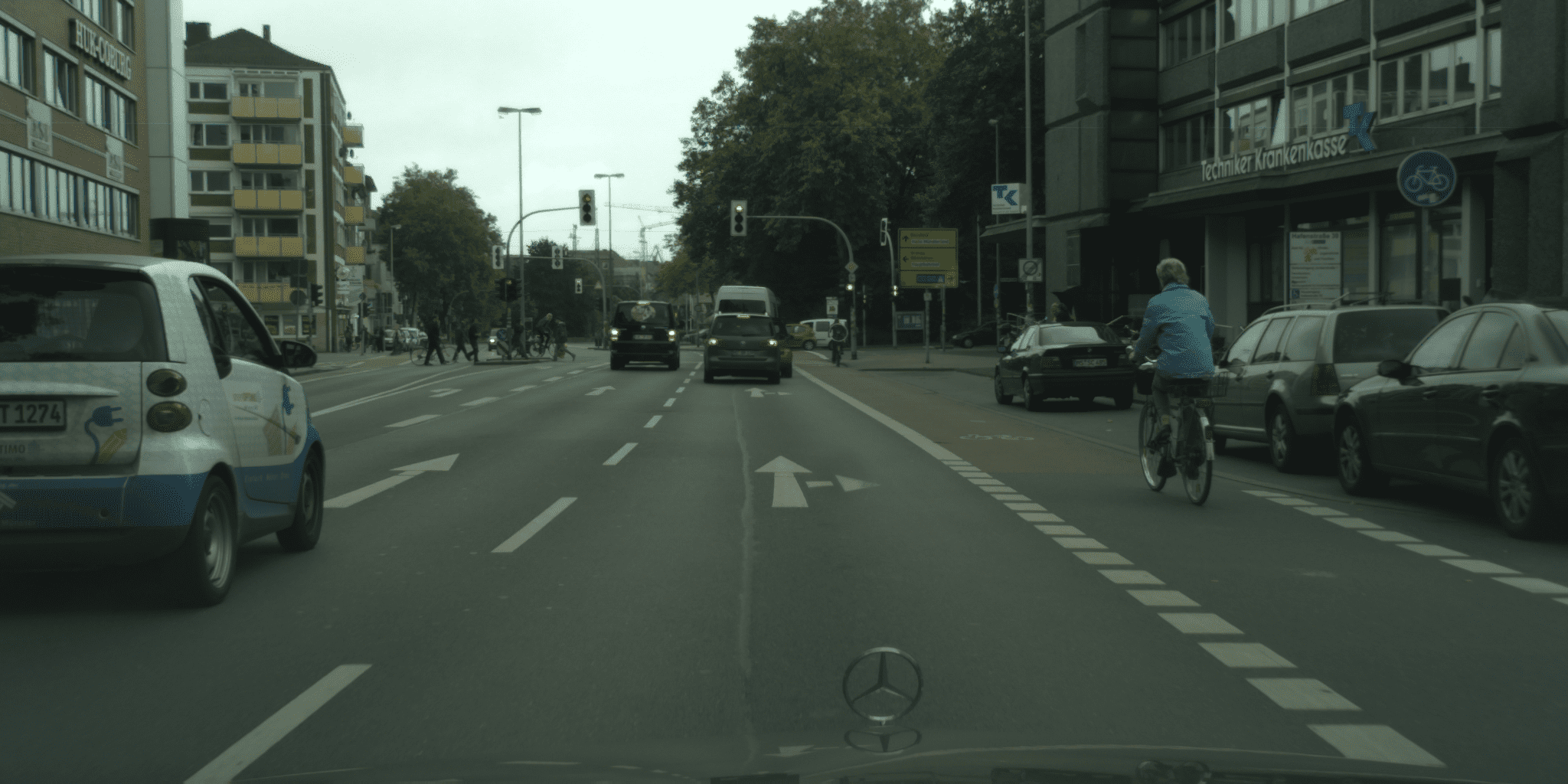}&
        \includegraphics[width=0.25\columnwidth]{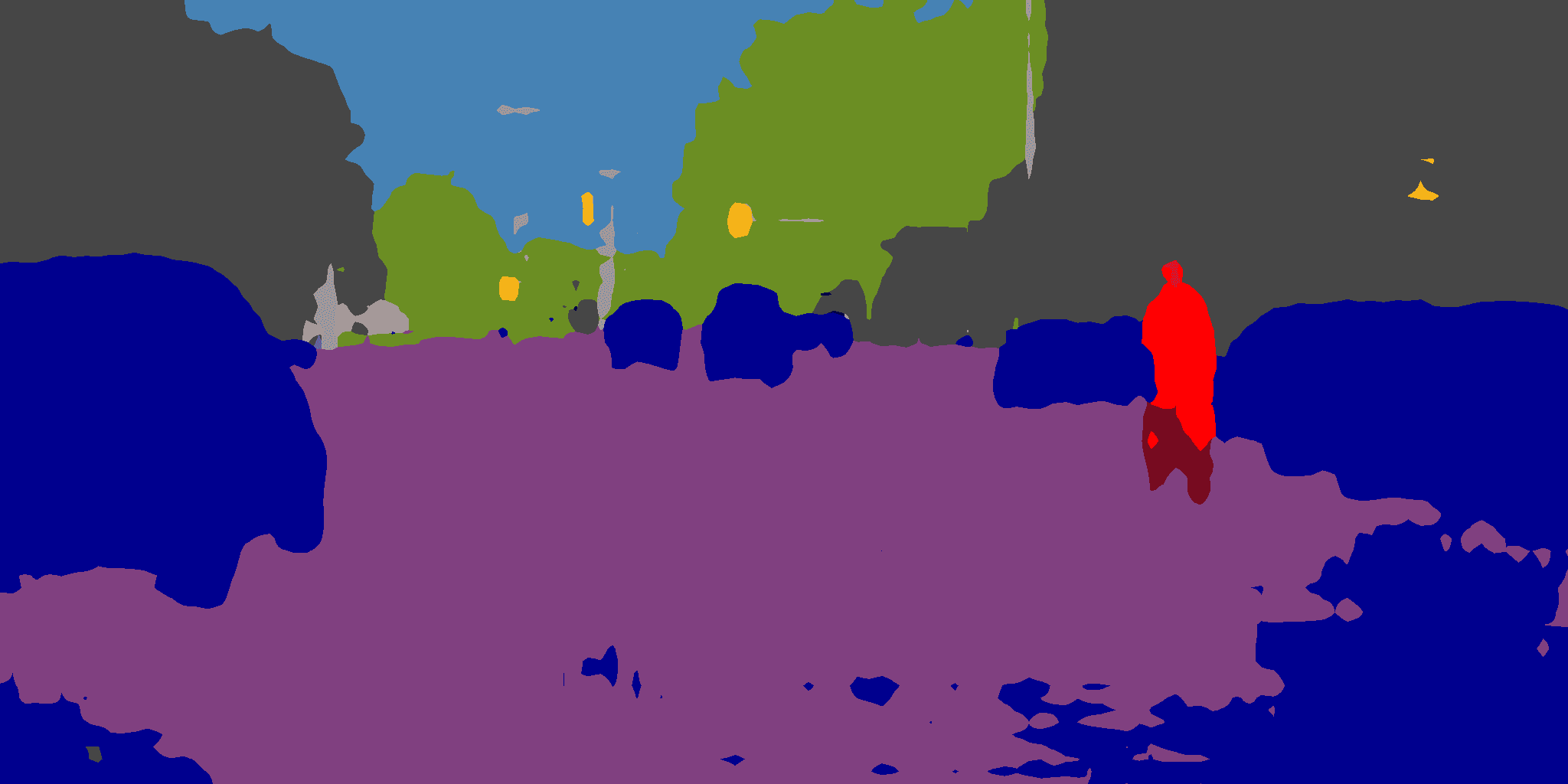}&
        \includegraphics[width=0.25\columnwidth]{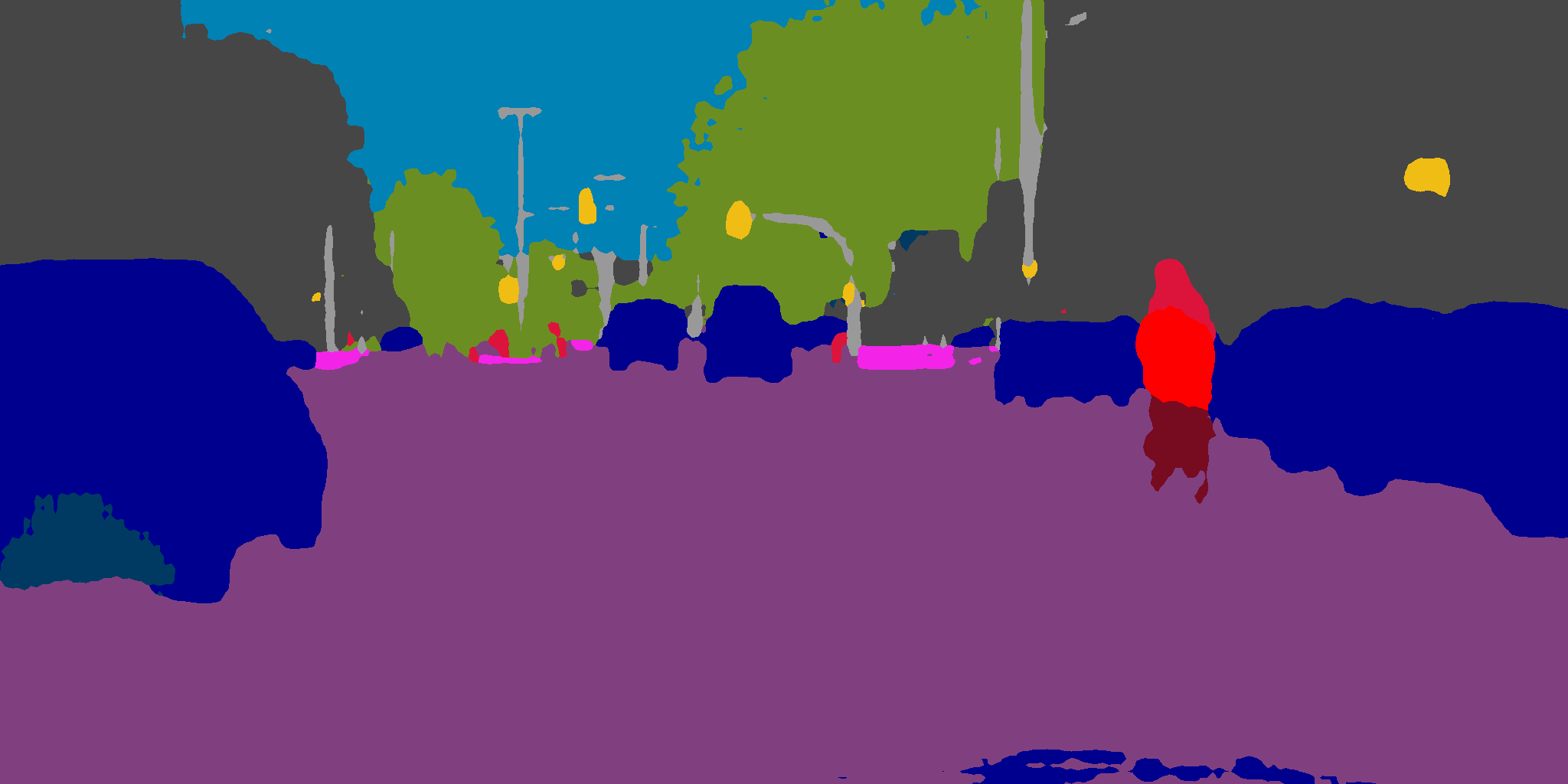}&
        \includegraphics[width=0.25\columnwidth]{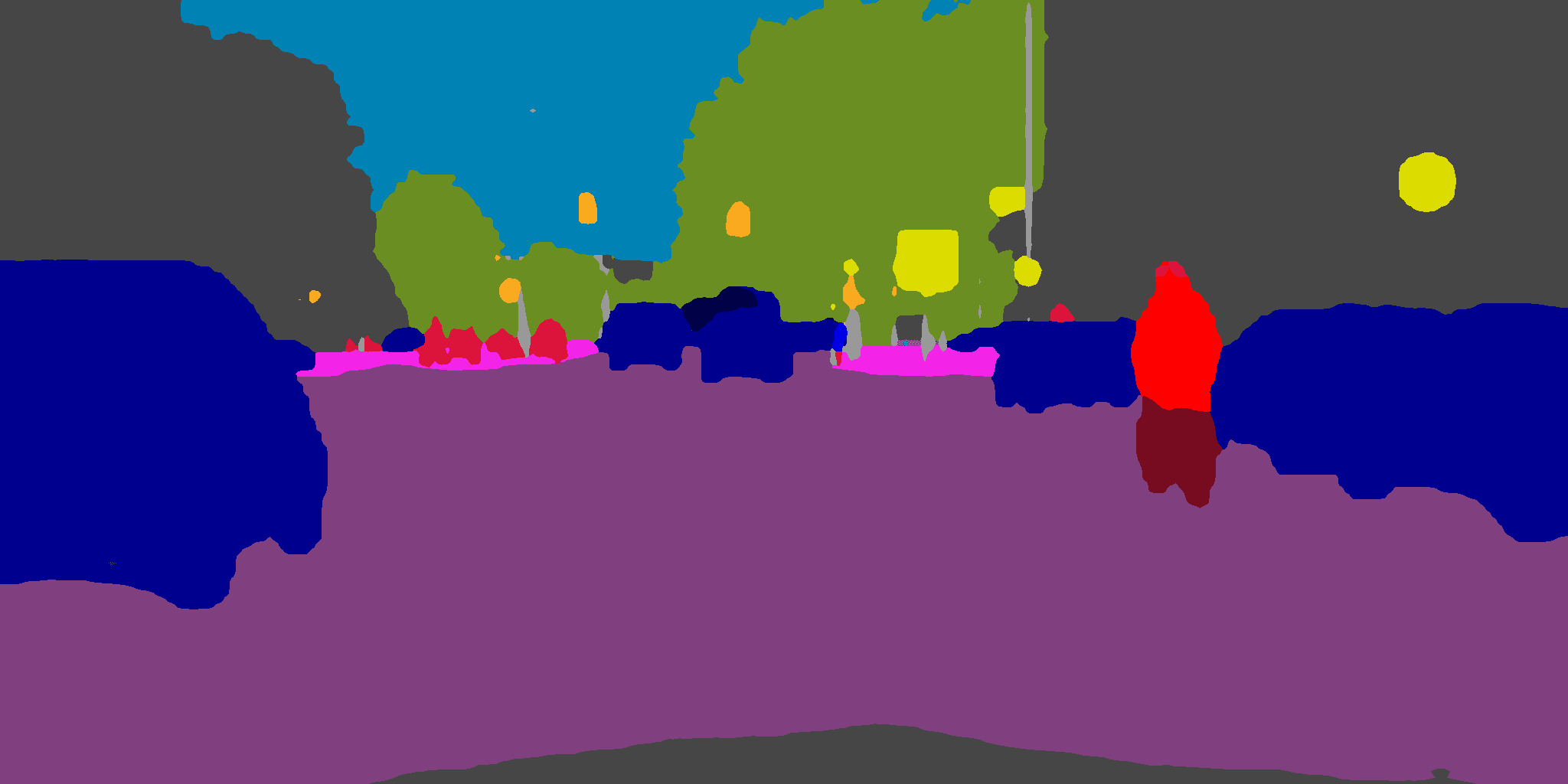}\\

        \includegraphics[width=0.25\columnwidth]{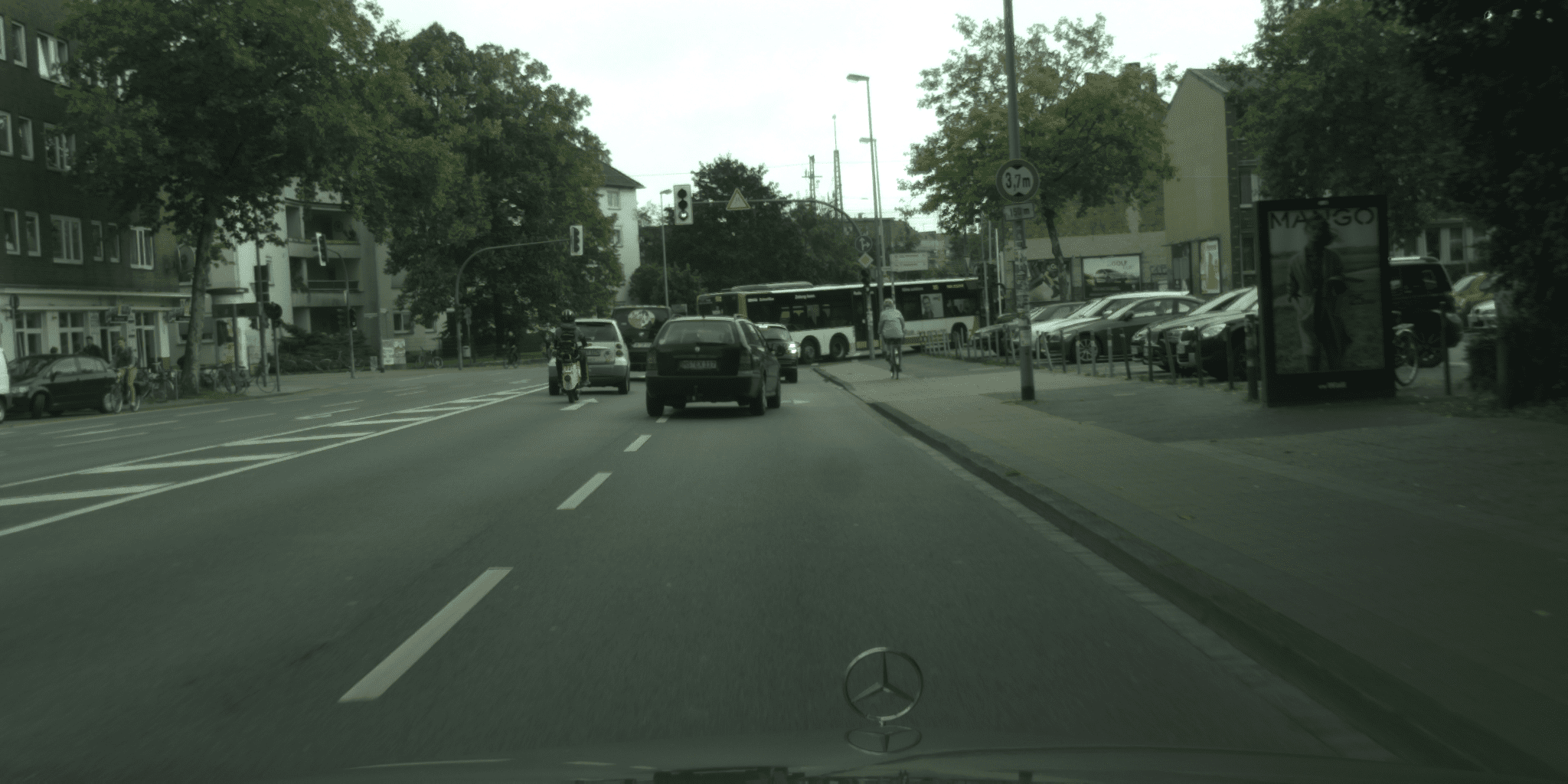}&
        \includegraphics[width=0.25\columnwidth]{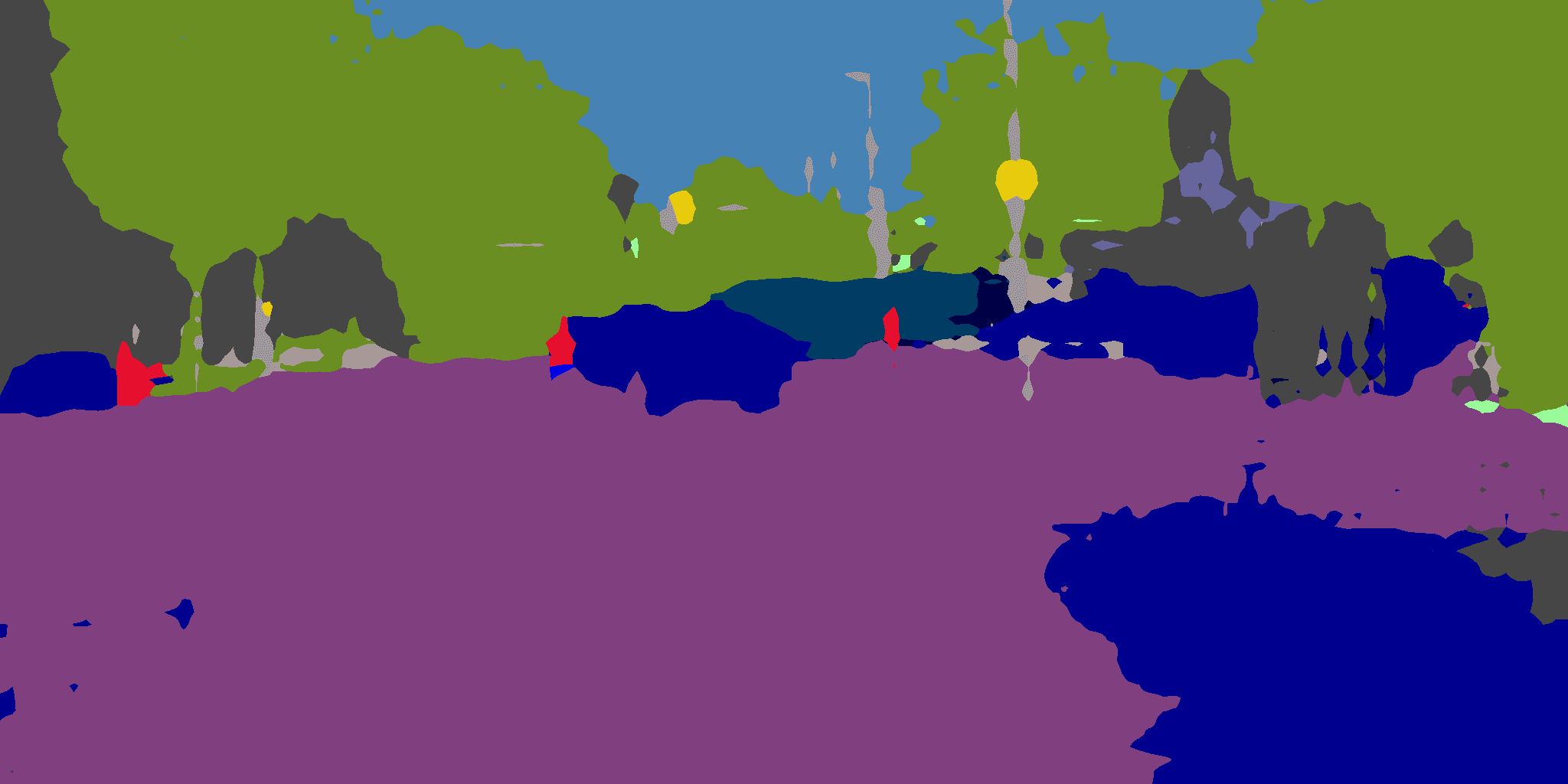}&
        \includegraphics[width=0.25\columnwidth]{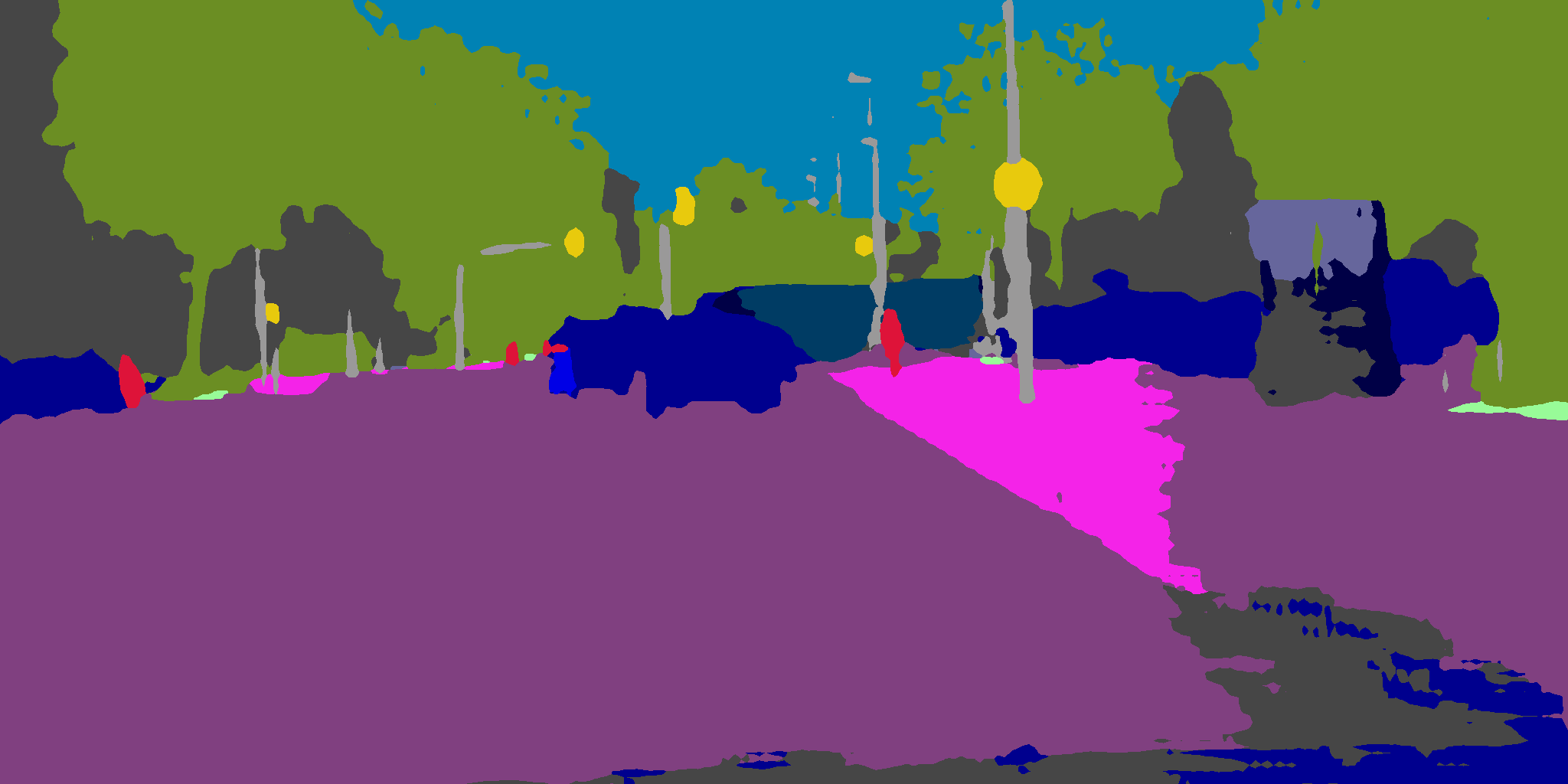}&
        \includegraphics[width=0.25\columnwidth]{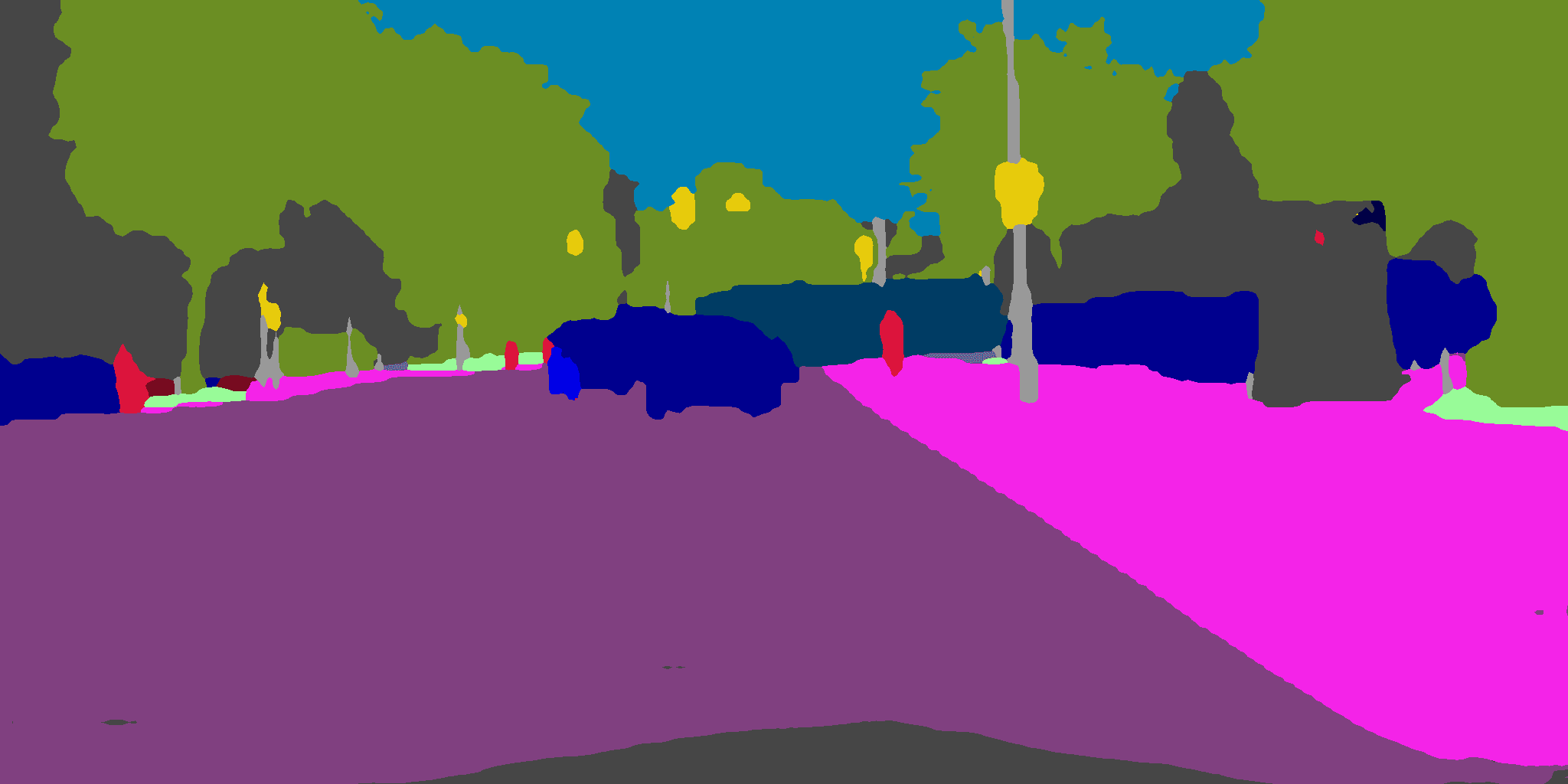}\\

        \includegraphics[width=0.25\columnwidth]{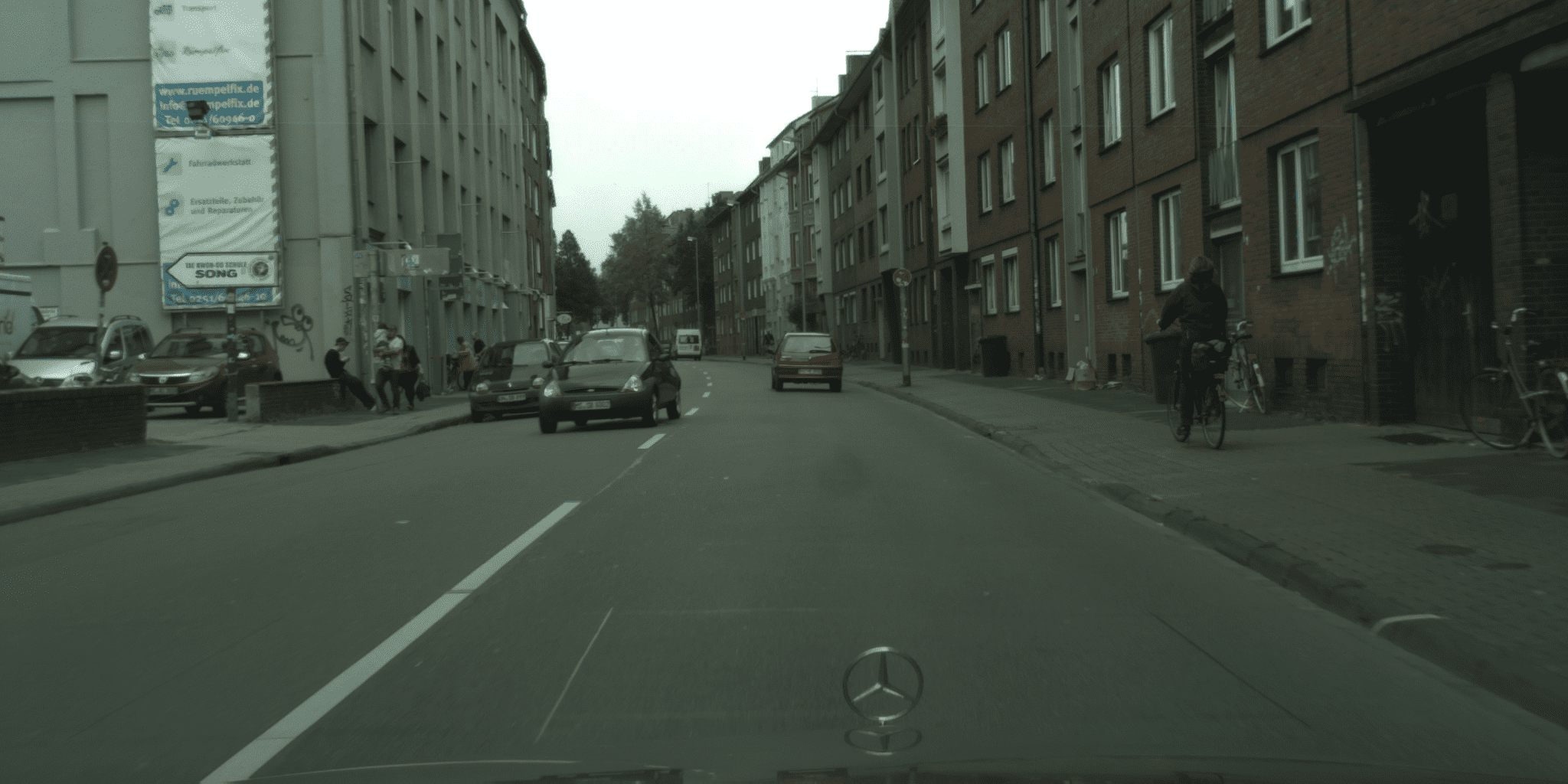}&
        \includegraphics[width=0.25\columnwidth]{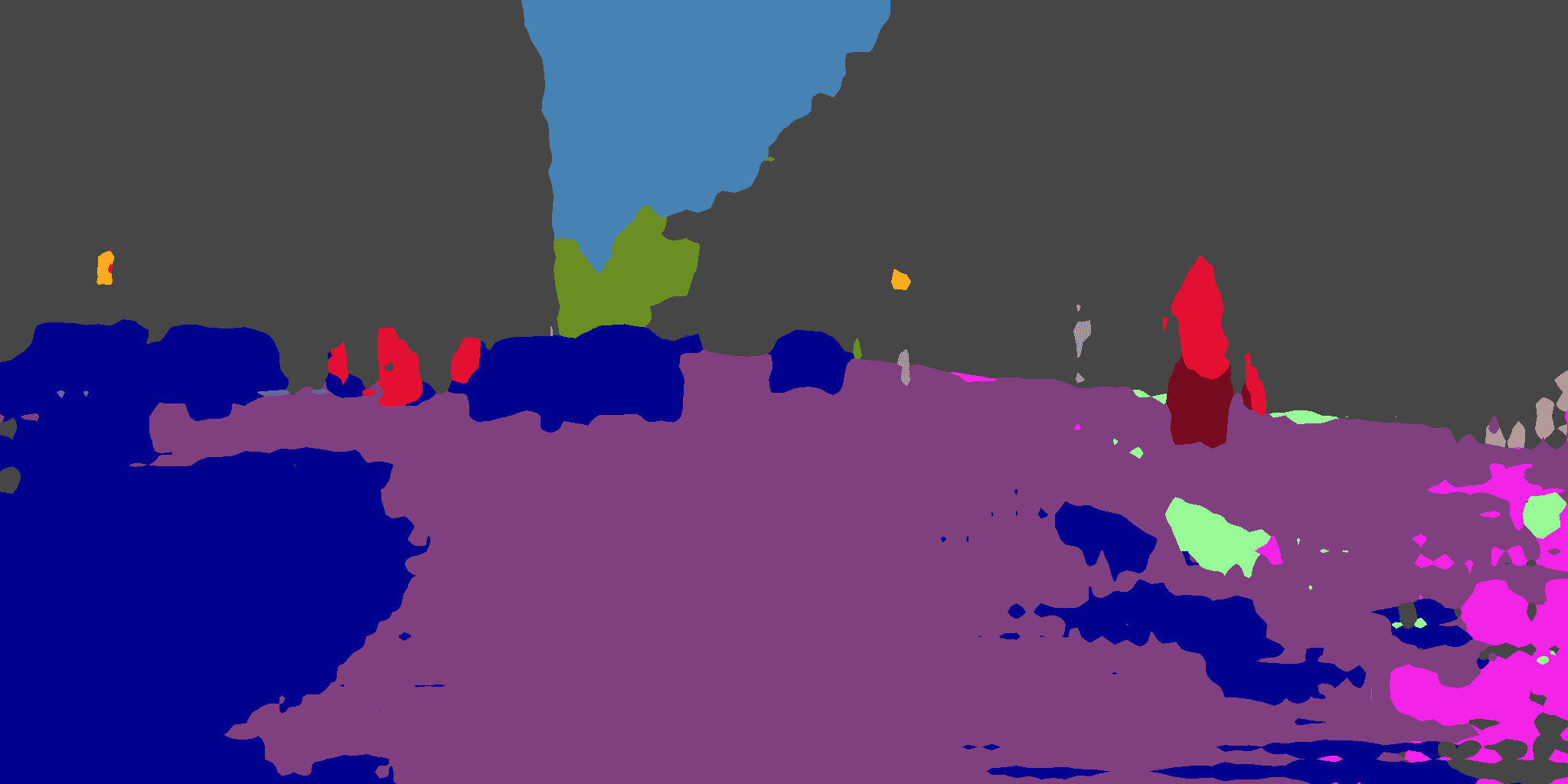}&
        \includegraphics[width=0.25\columnwidth]{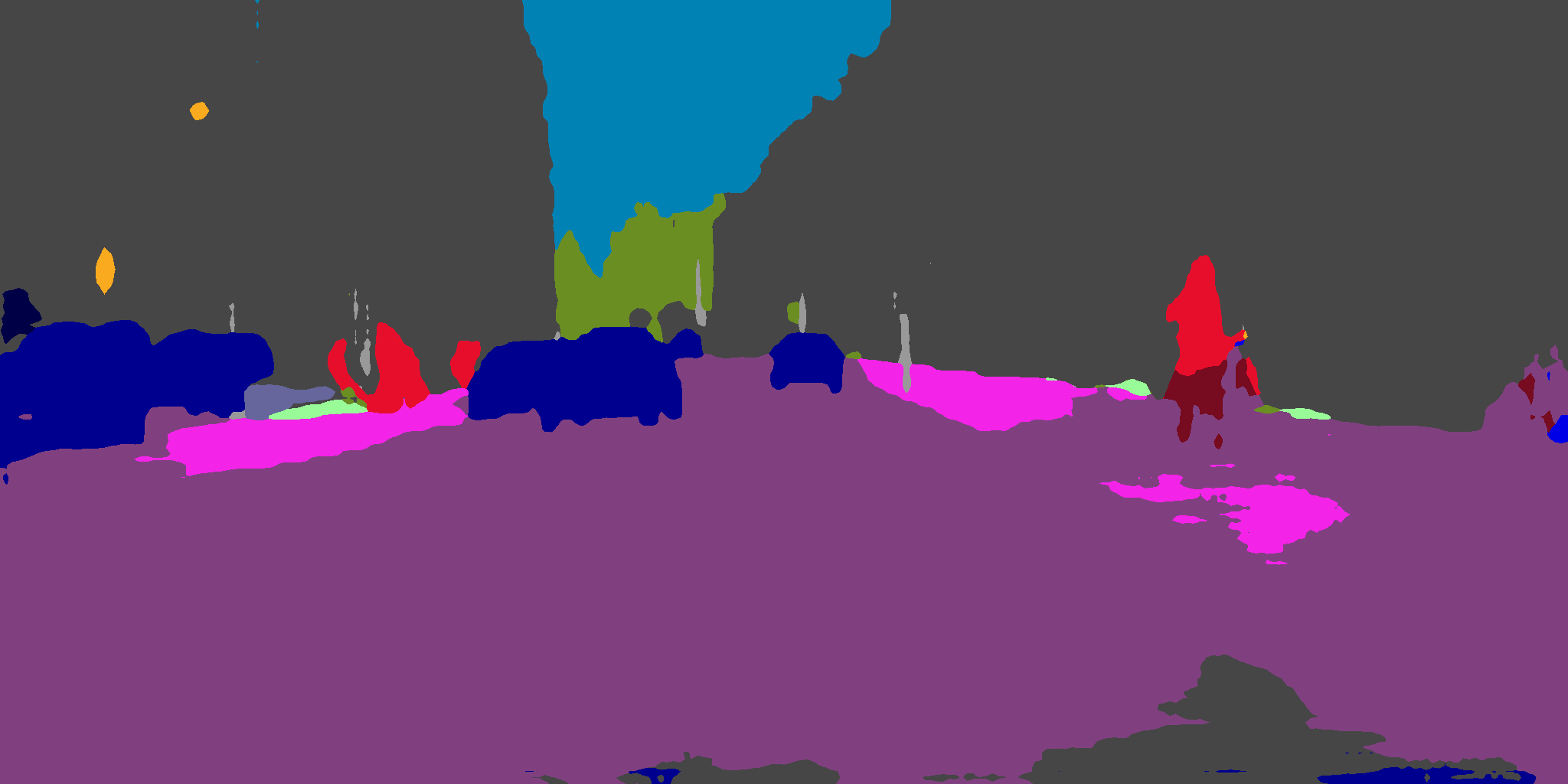}&
        \includegraphics[width=0.25\columnwidth]{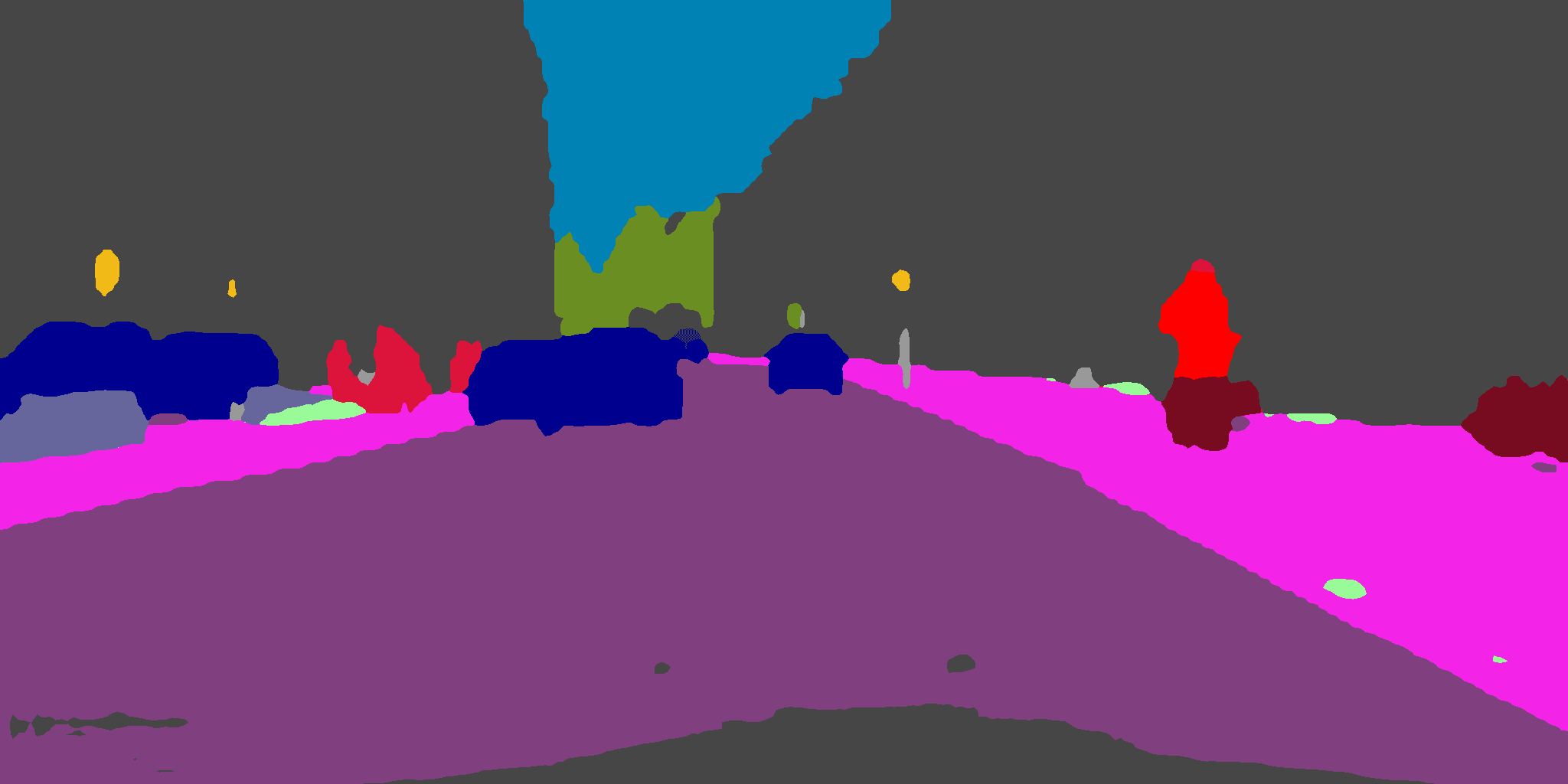}\\

        \includegraphics[width=0.25\columnwidth]{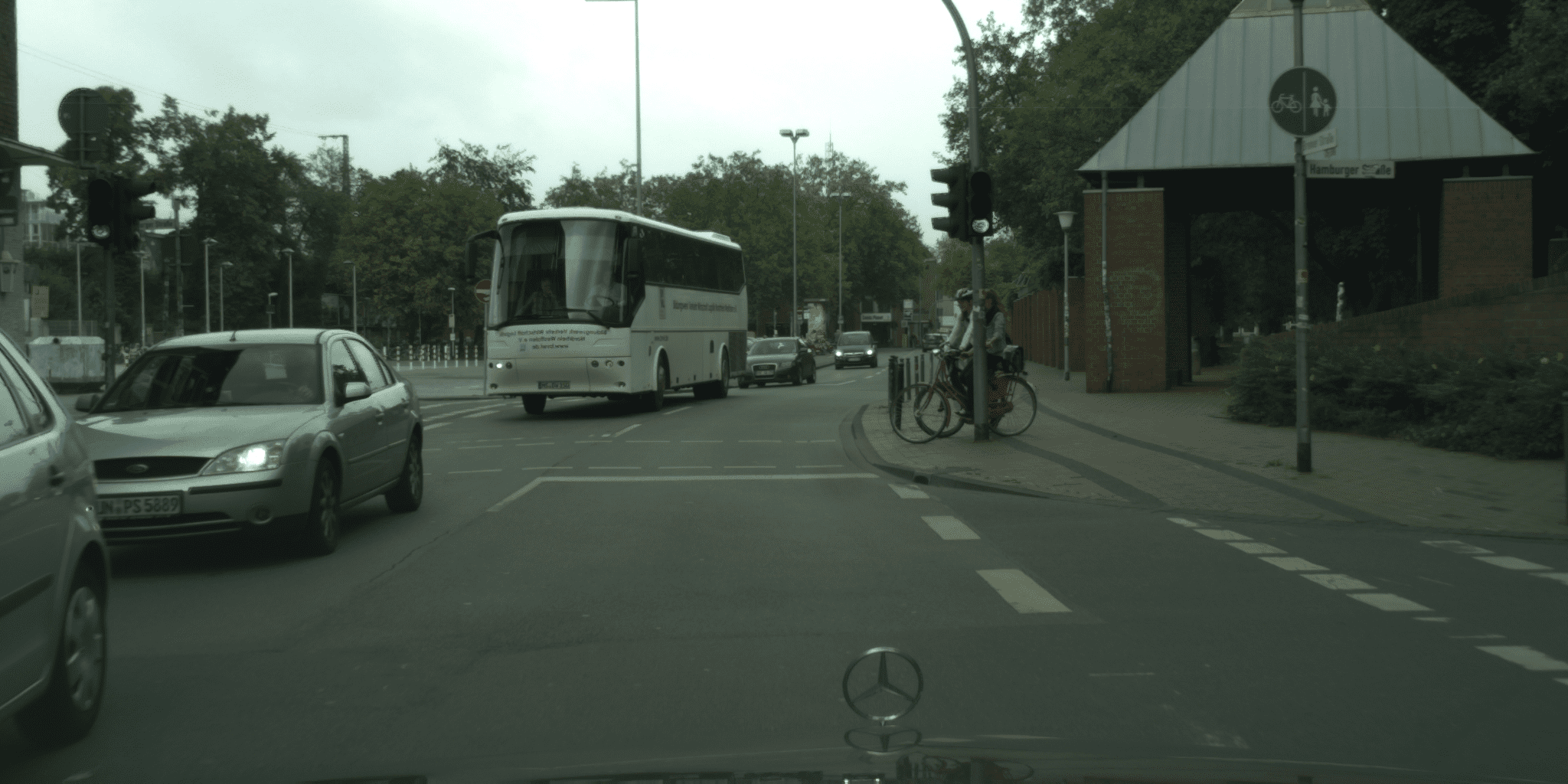}&
        \includegraphics[width=0.25\columnwidth]{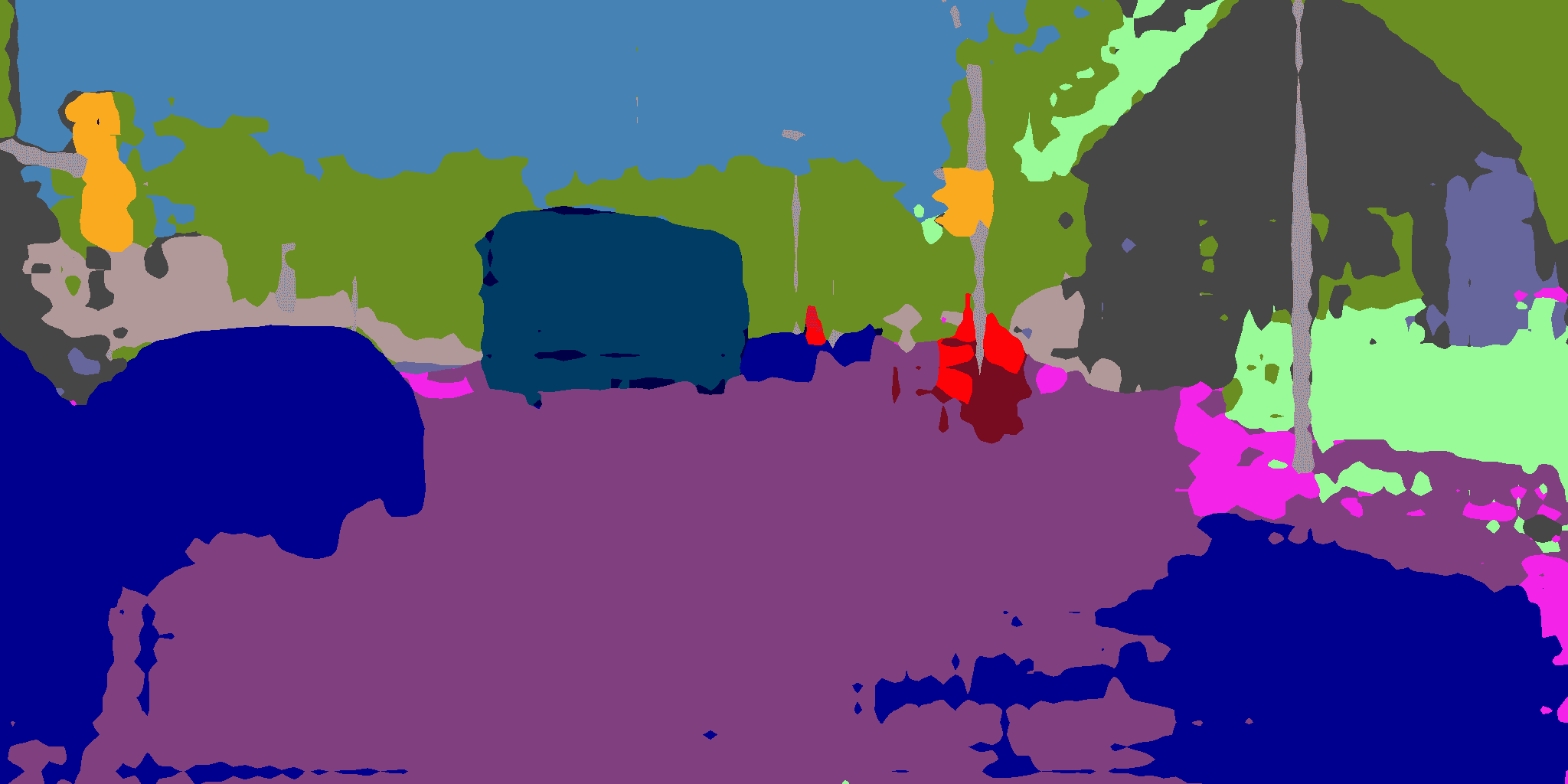}&
        \includegraphics[width=0.25\columnwidth]{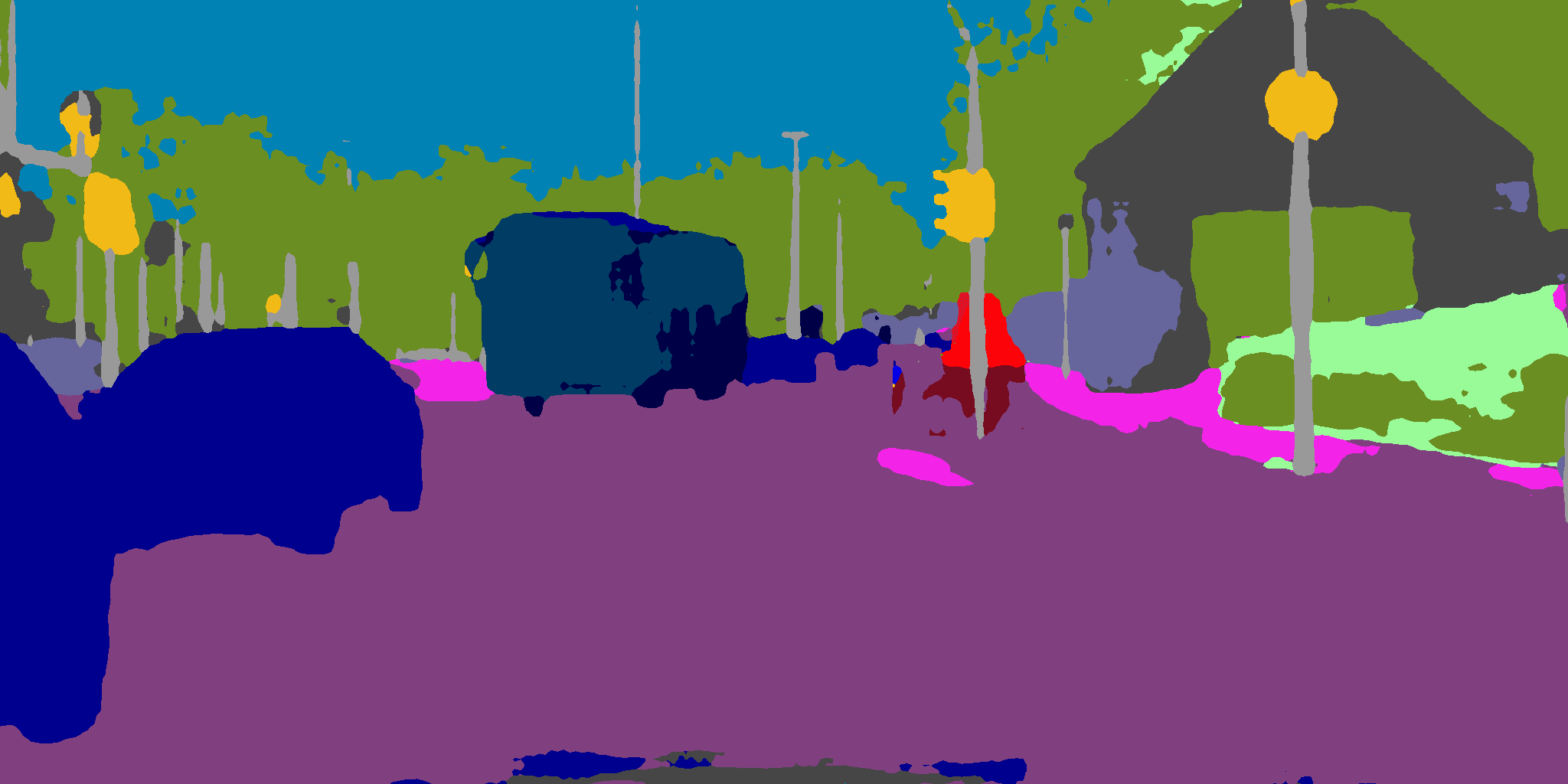}&
        \includegraphics[width=0.25\columnwidth]{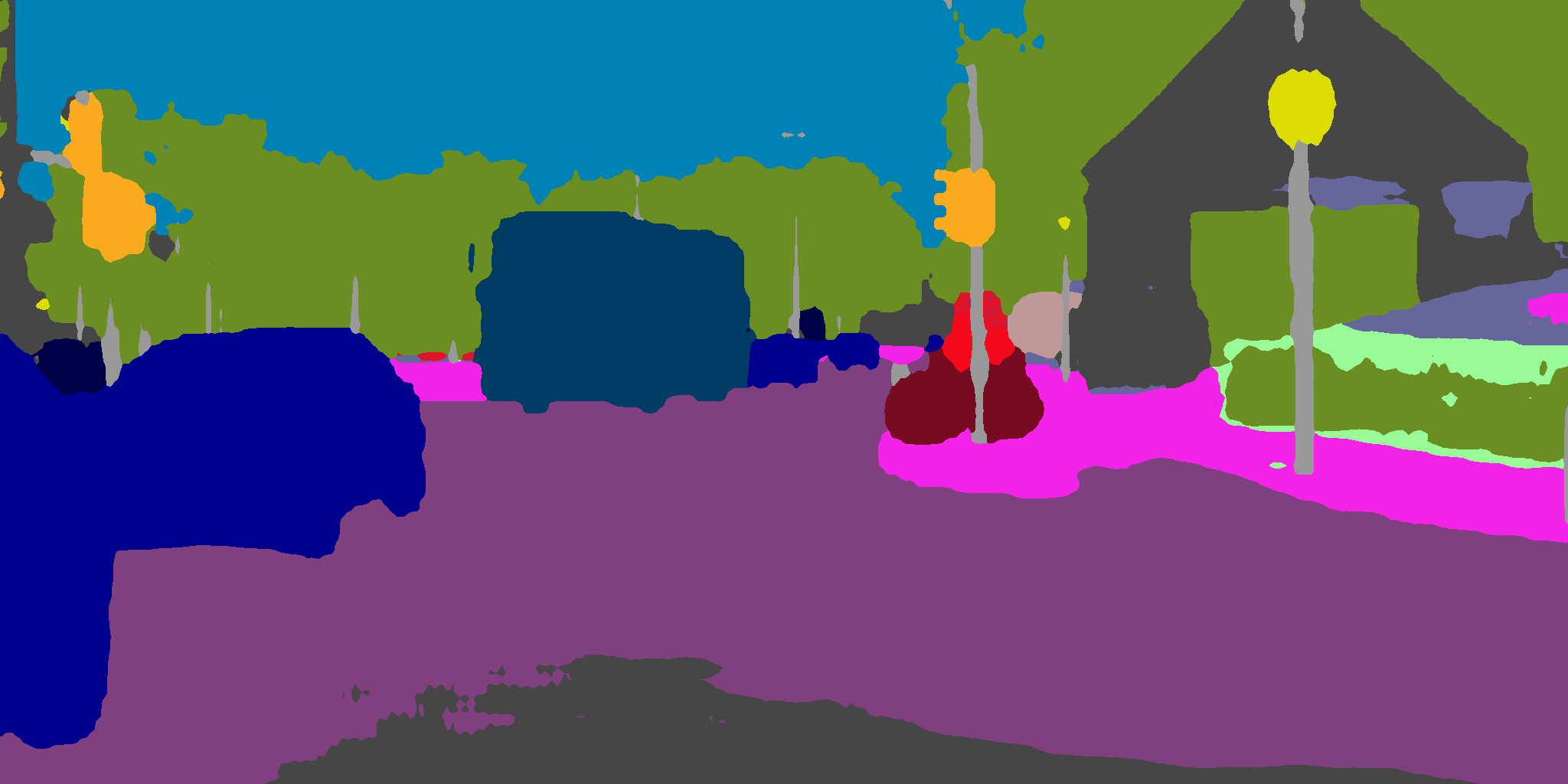}\\

        \includegraphics[width=0.25\columnwidth]{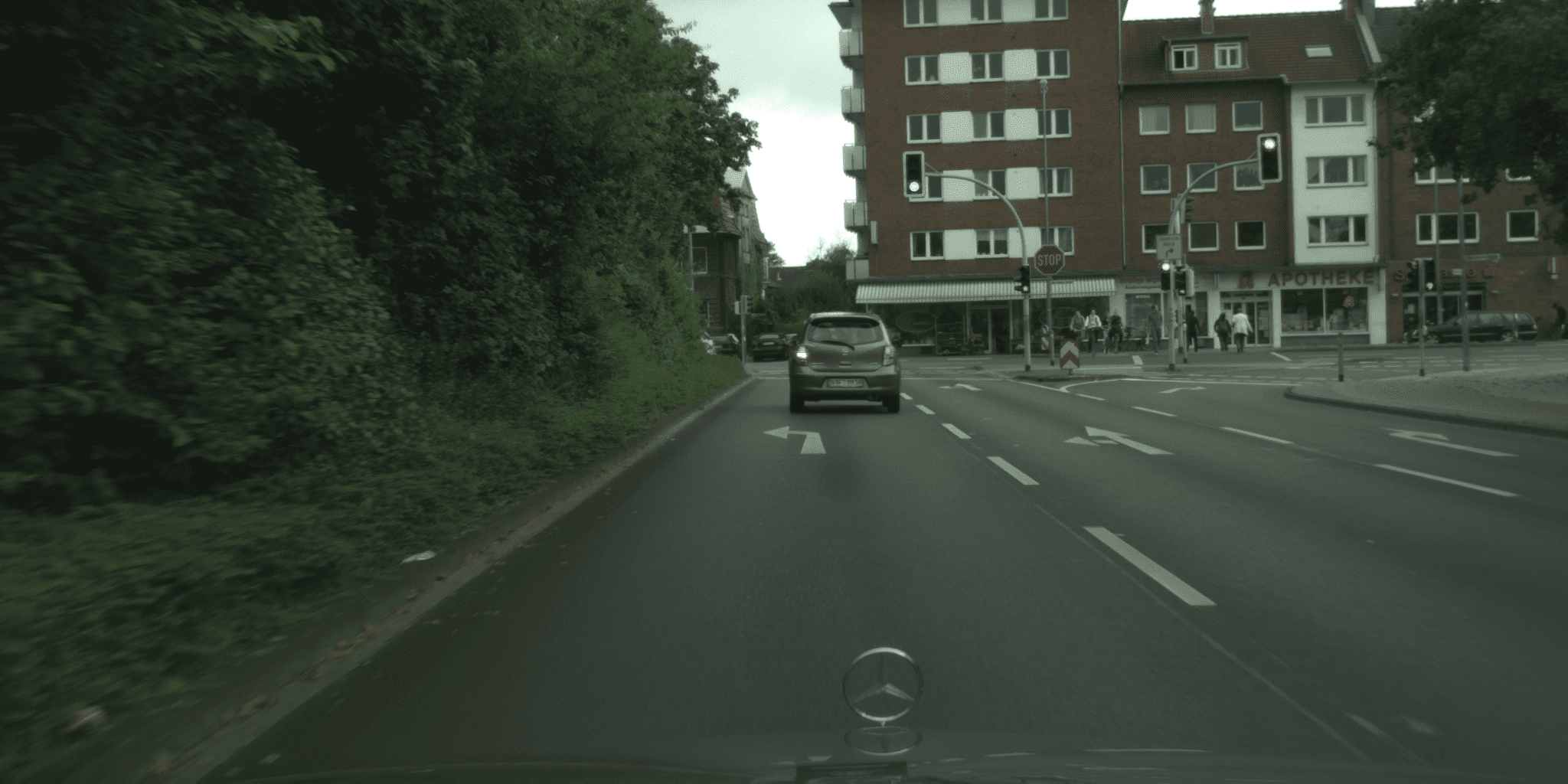}&
        \includegraphics[width=0.25\columnwidth]{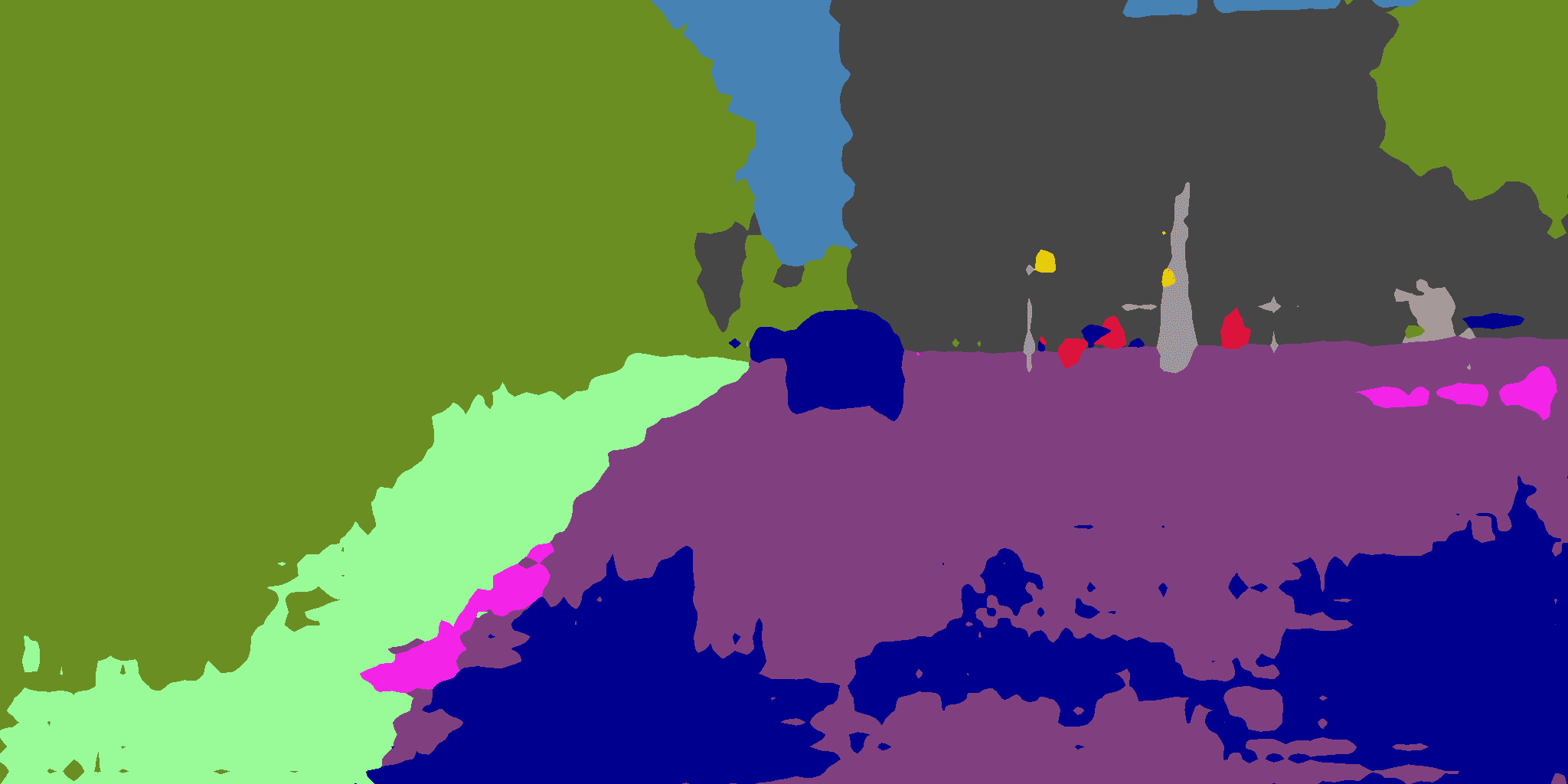}&
        \includegraphics[width=0.25\columnwidth]{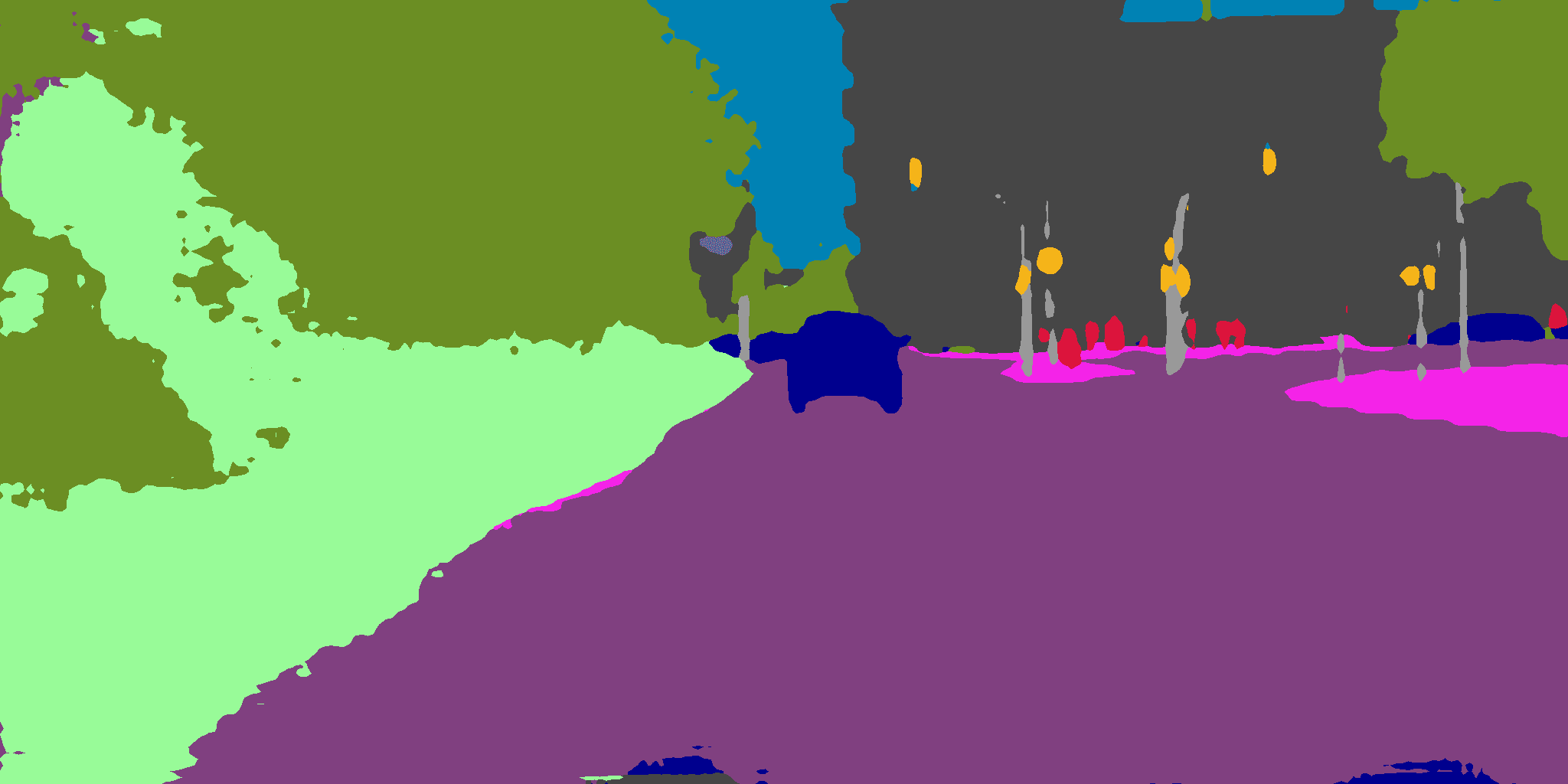}&
        \includegraphics[width=0.25\columnwidth]{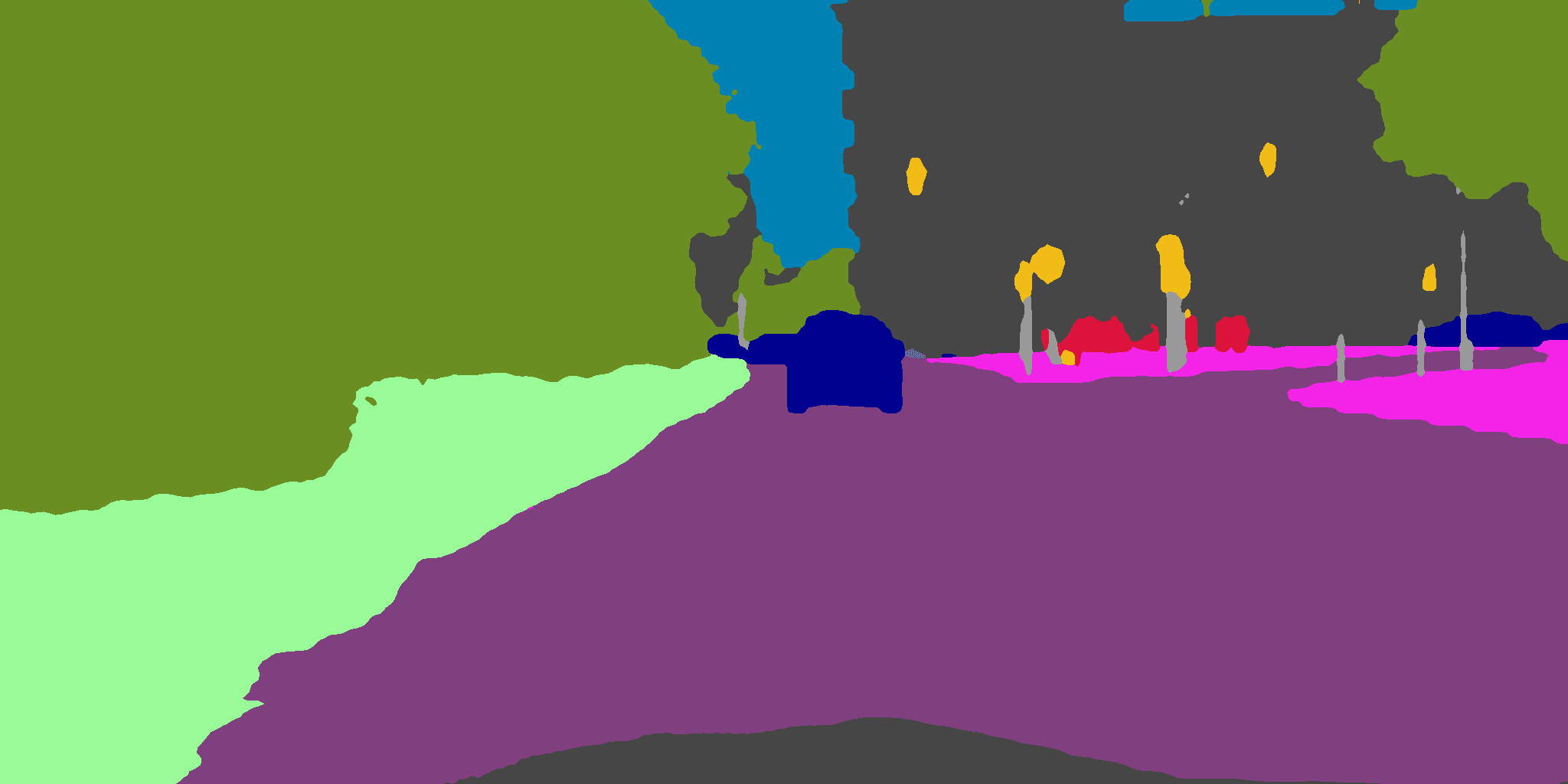}\\
        
        Input & Before adaptation & Conventional self-training & ProDA \\
    \end{tabular}
    }
    \caption{Qualitative results of semantic segmentation on the Cityscapes dataset. From left to right: input, before adaptation, conventional self-training, ProDA.}
    \label{figure:compare_baseline}
    \end{figure}

\begin{figure}[t]
    \center
    \small
    \setlength\tabcolsep{1pt}
    {
    
    \newcolumntype{P}[1]{>{\centering\arraybackslash}p{#1}}
    \begin{tabular}{@{}*{10}{P{0.0974\columnwidth}}@{}}
         {\cellcolor[rgb]{0.5,0.25,0.5}}\textcolor{white}{road} &{\cellcolor[rgb]{0.957,0.137,0.91}}sidewalk &{\cellcolor[rgb]{0.275,0.275,0.275}}\textcolor{white}{building} &{\cellcolor[rgb]{0.4,0.4,0.612}}\textcolor{white}{wall} &{\cellcolor[rgb]{0.745,0.6,0.6}}fence &{\cellcolor[rgb]{0.6,0.6,0.6}}pole &{\cellcolor[rgb]{0.98,0.667,0.118}}traffic light&{\cellcolor[rgb]{0.863,0.863,0}}traffic sign &{\cellcolor[rgb]{0.42,0.557,0.137}}vegetation & {\cellcolor[rgb]{0,0,0}}\textcolor{white}{n/a.}\\
         
         {\cellcolor[rgb]{0.596,0.984,0.596}}terrain &{\cellcolor[rgb]{0,0.51,0.706}}sky &{\cellcolor[rgb]{0.863,0.078,0.235}}\textcolor{white}{person} &{\cellcolor[rgb]{1,0,0}}\textcolor{white}{rider} &{\cellcolor[rgb]{0,0,0.557}}\textcolor{white}{car} &{\cellcolor[rgb]{0,0,0.275}}\textcolor{white}{truck} &{\cellcolor[rgb]{0,0.235,0.392}}\textcolor{white}{bus}&{\cellcolor[rgb]{0,0.314,0.392}}\textcolor{white}{train} &{\cellcolor[rgb]{0,0,0.902}}\textcolor{white}{motorcycle} & {\cellcolor[rgb]{0.467,0.043,0.125}}\textcolor{white}{bike}\\
    
    \end{tabular}
    
    \renewcommand{\arraystretch}{0.6}
    \begin{tabular}{@{}ccc@{}}
    %    \multicolumn{4}{c}{\includegraphics[width=2.088\columnwidth]{figure/class_color.png}}\\
    
        \includegraphics[width=0.335\columnwidth]{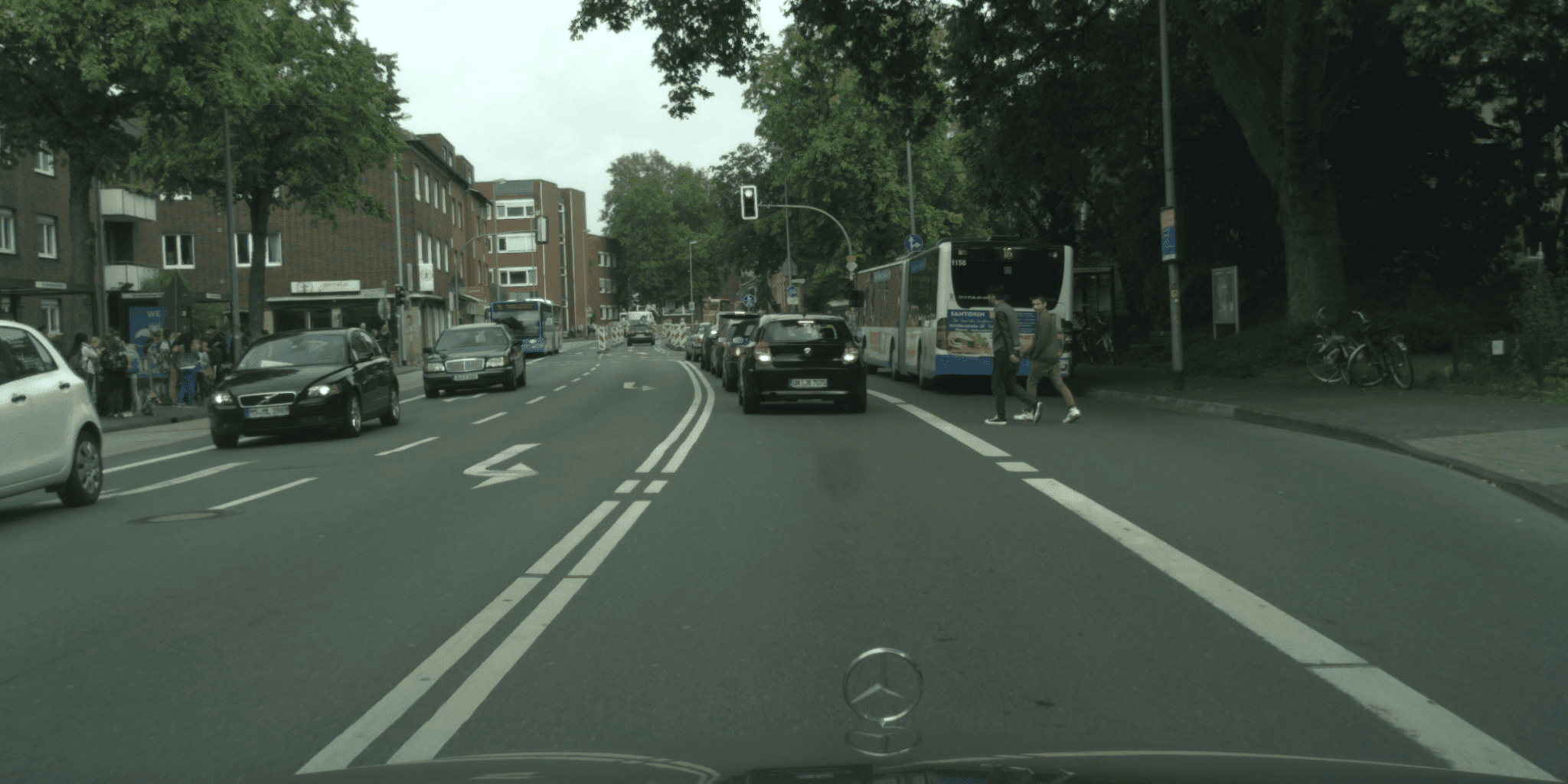}&
        \includegraphics[width=0.335\columnwidth]{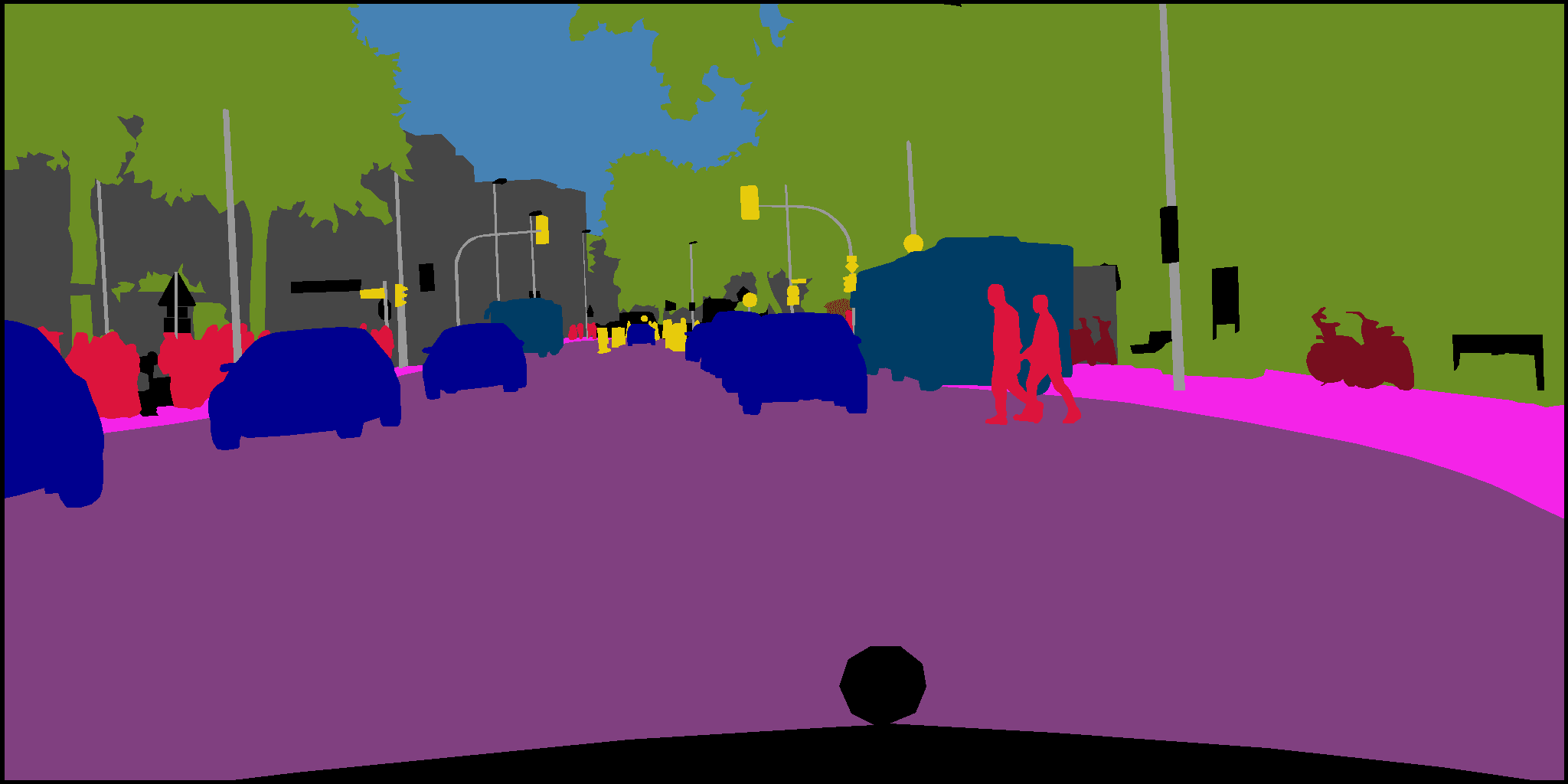}&
        \includegraphics[width=0.335\columnwidth]{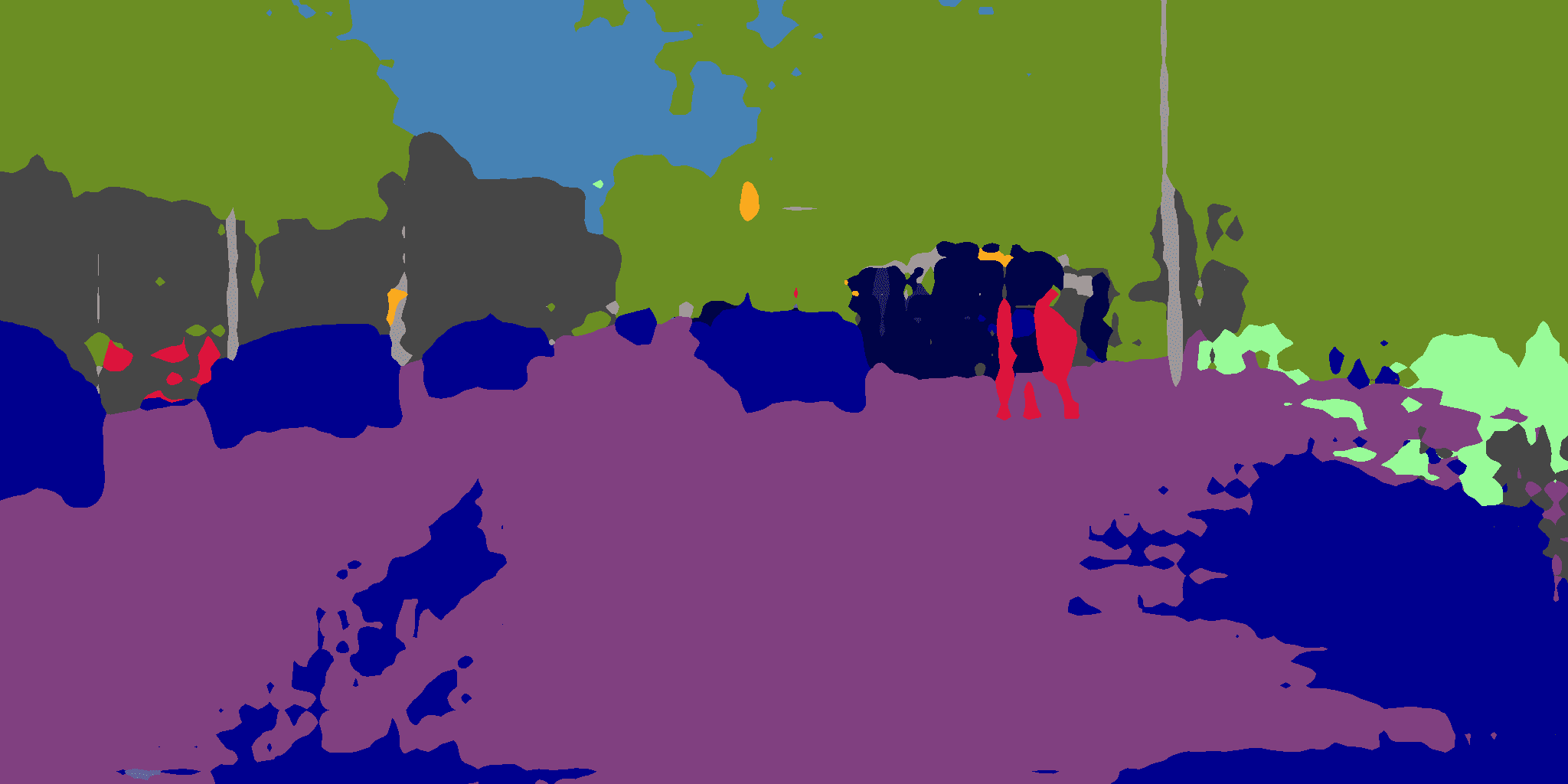}\\
        Input & Ground truth & Befor adaptation \\
        \includegraphics[width=0.335\columnwidth]{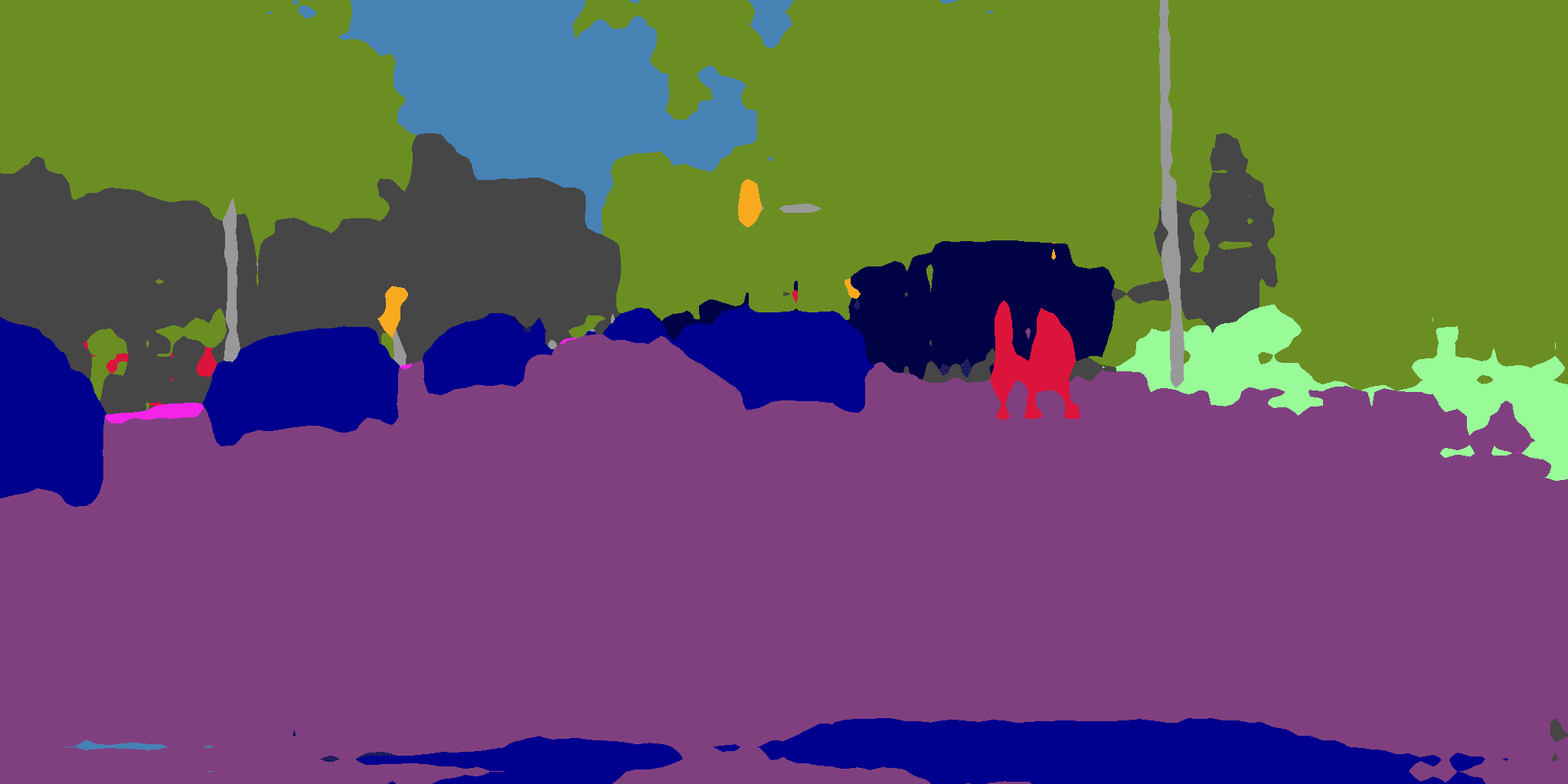}&
        \includegraphics[width=0.335\columnwidth]{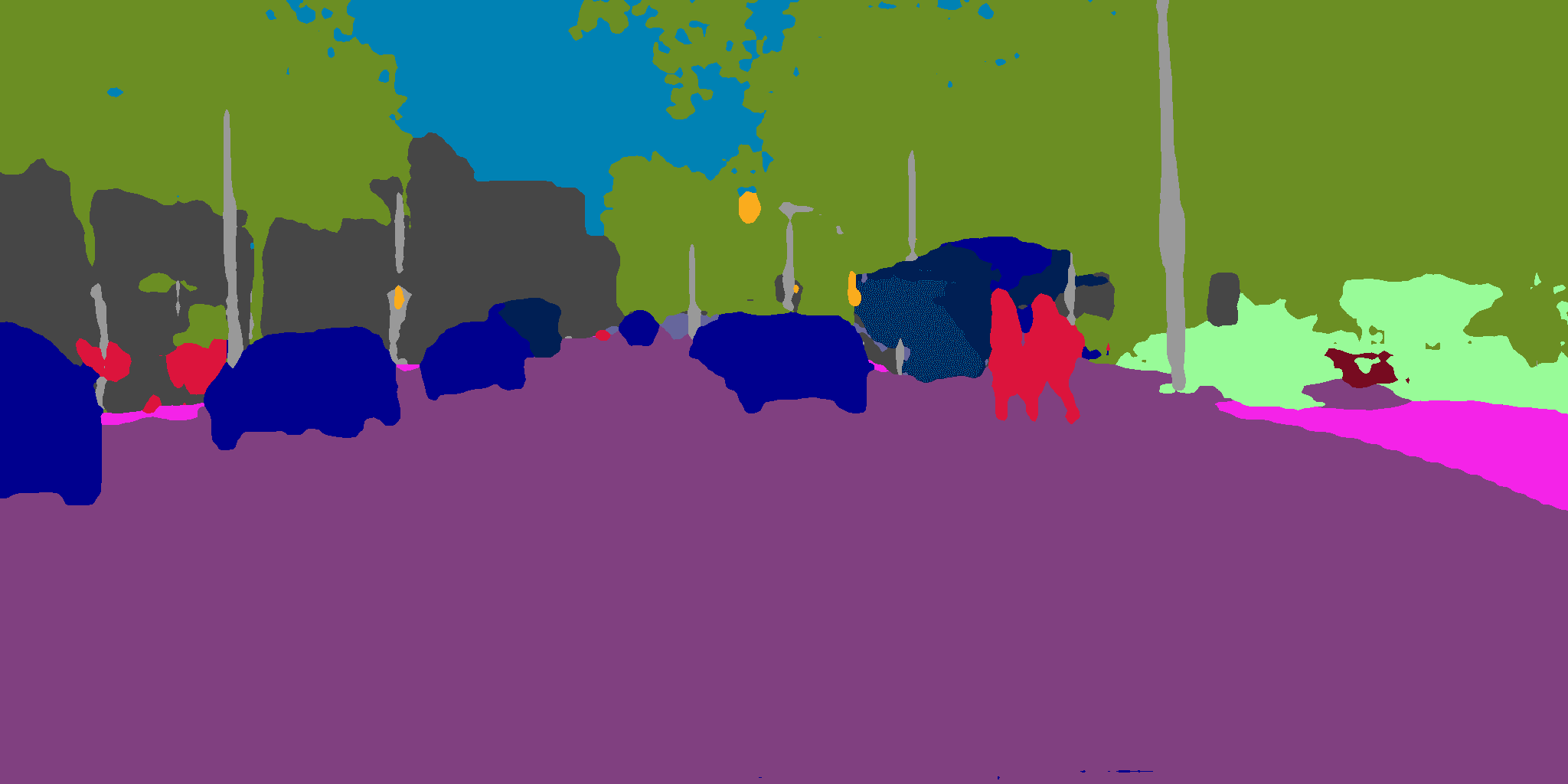}&
        \includegraphics[width=0.335\columnwidth]{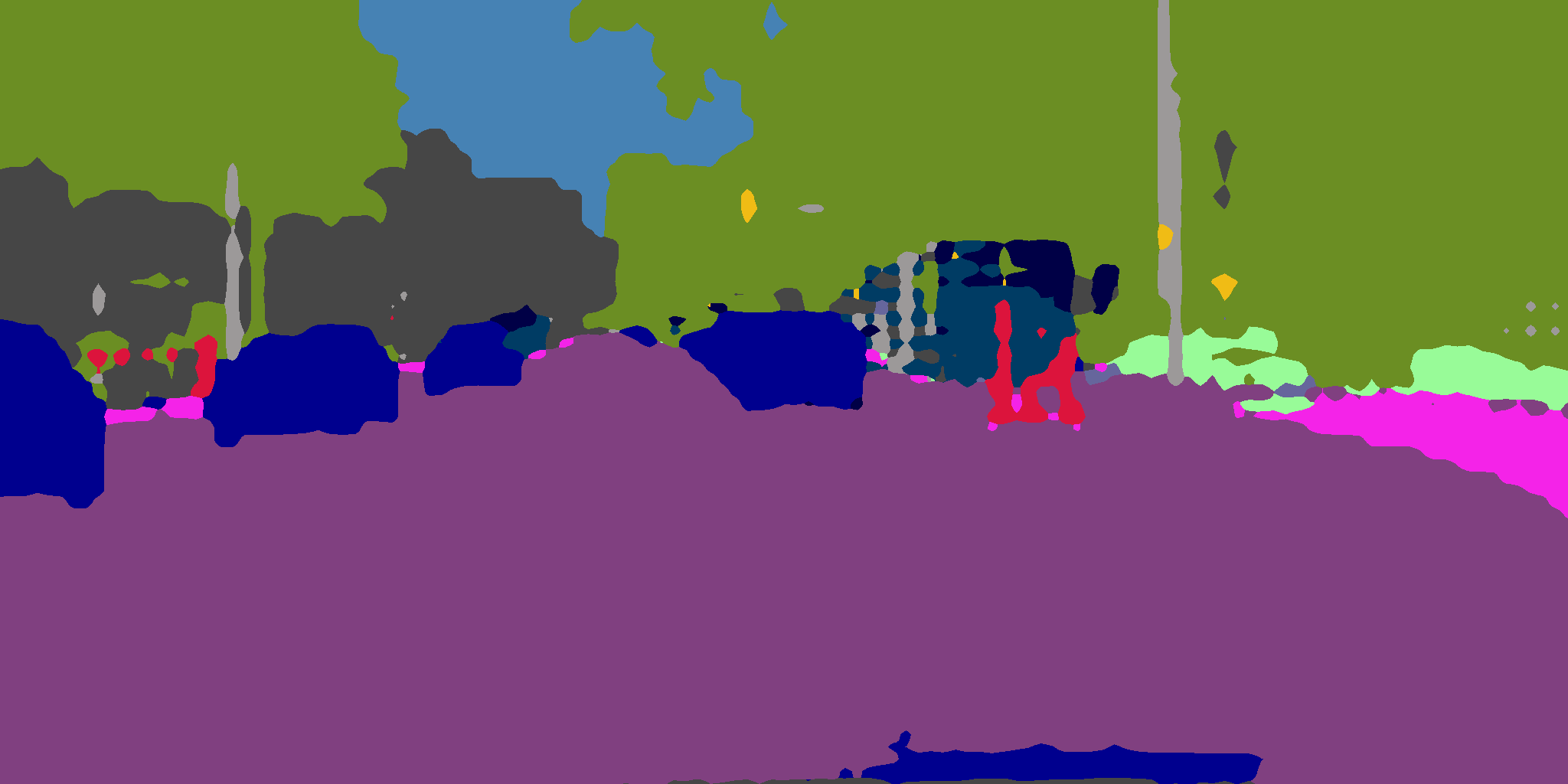}\\
        AdaptSeg~\cite{Tsai_adaptseg_2018} & CBST~\cite{zou2018unsupervised} & BDL~\cite{li2019bidirectional} \\
        \includegraphics[width=0.335\columnwidth]{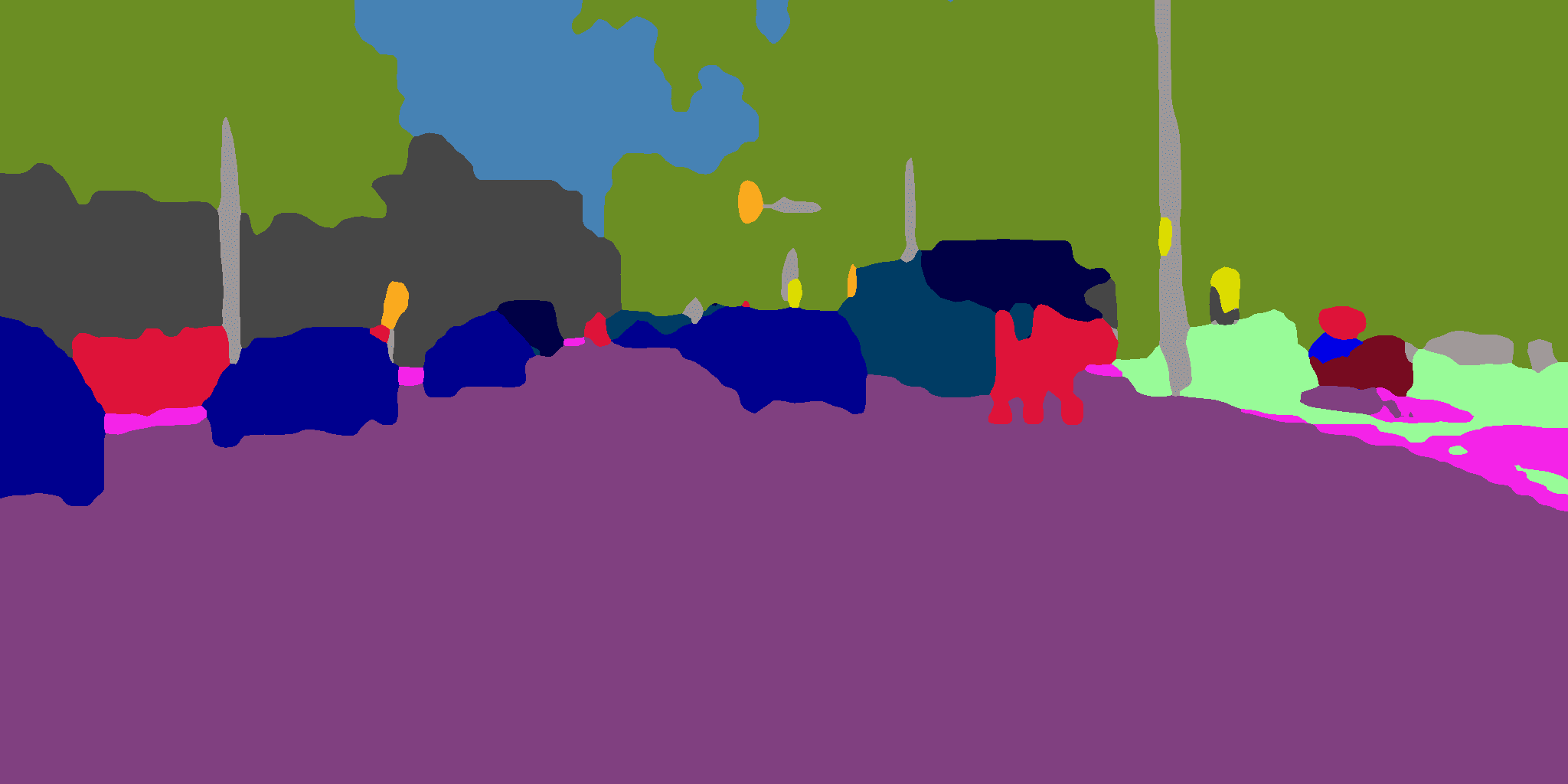}&
        \includegraphics[width=0.335\columnwidth]{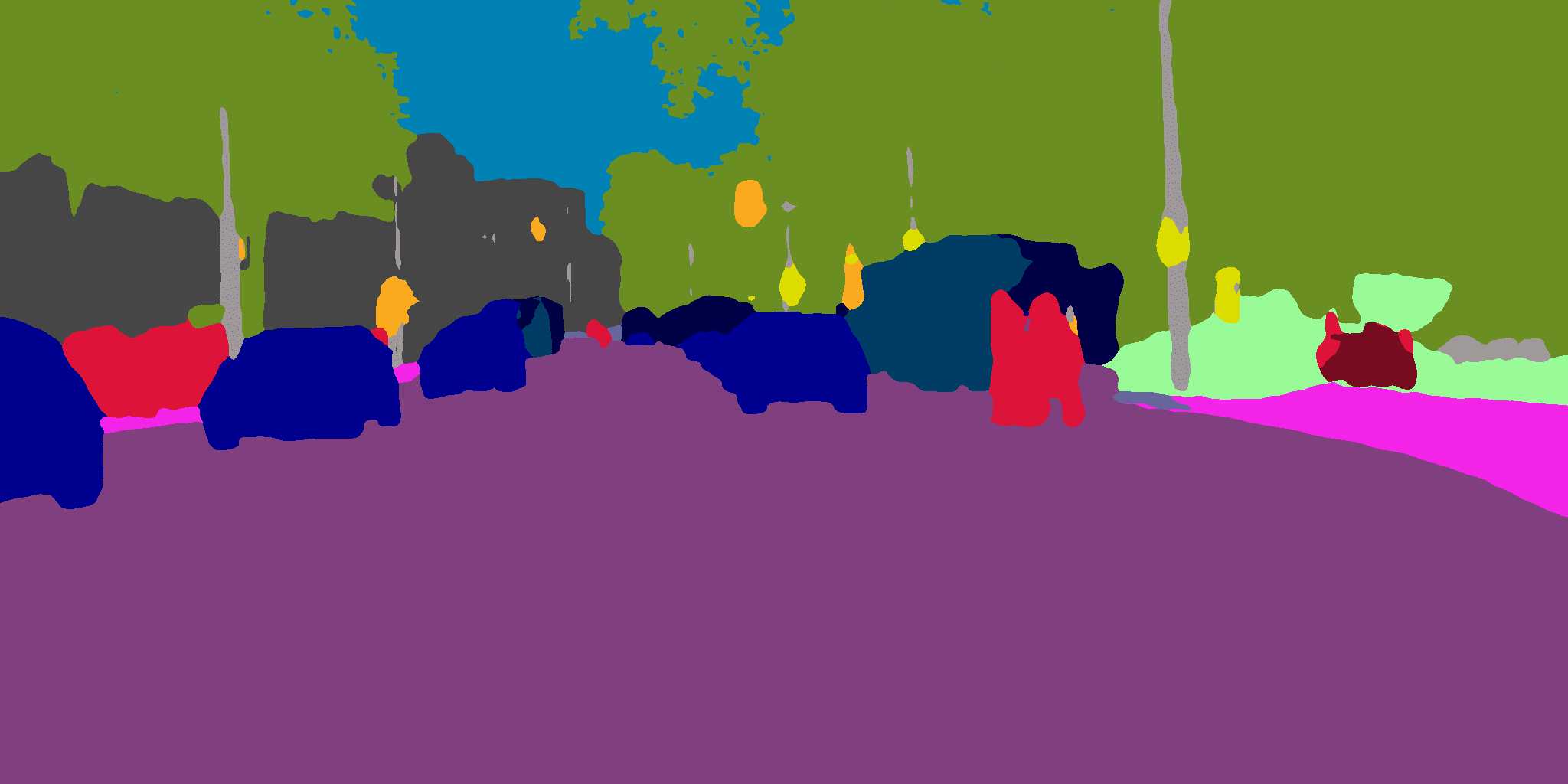}&
        \includegraphics[width=0.335\columnwidth]{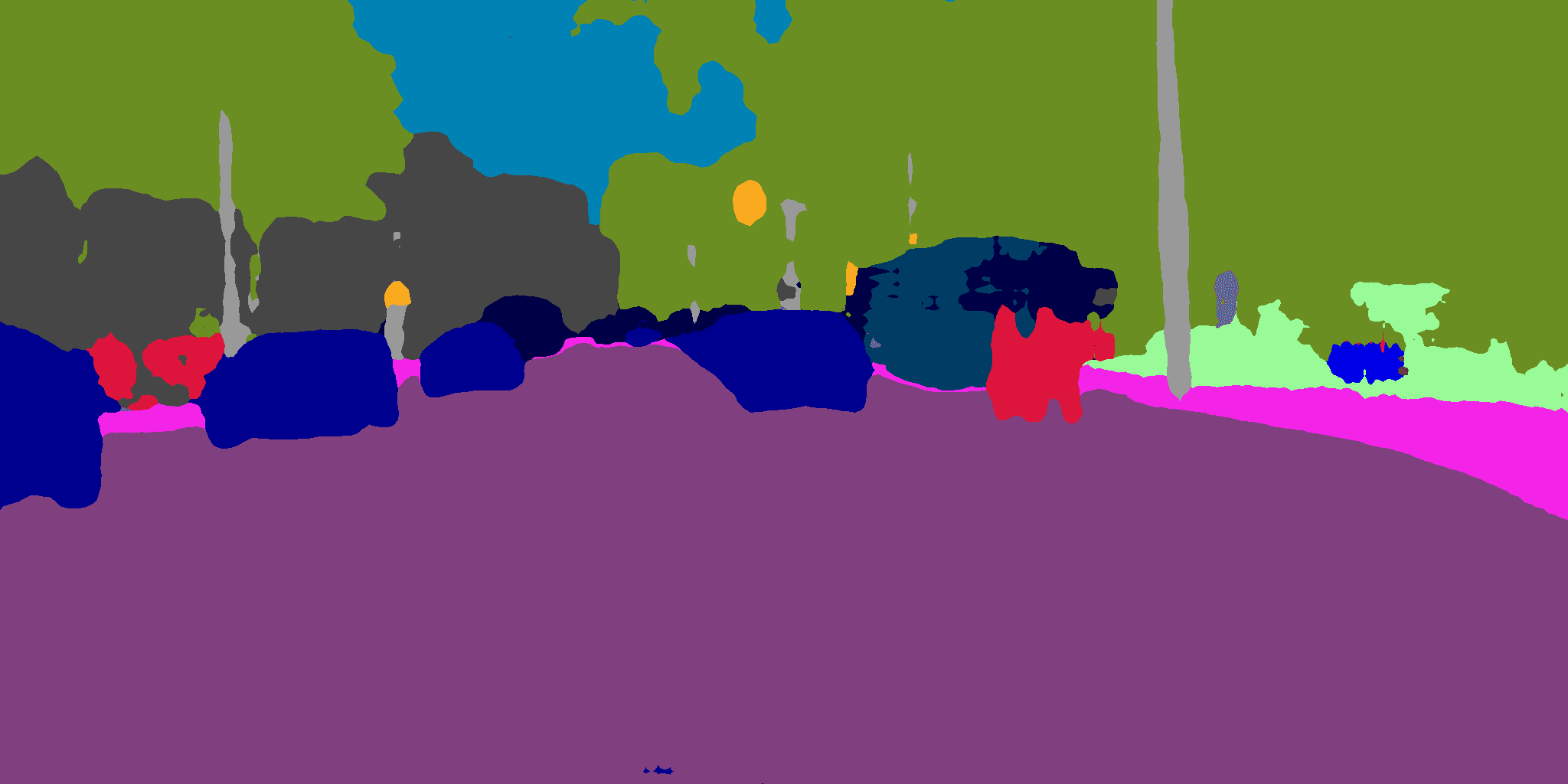}\\
        Seg-Uncertainty~\cite{zheng2020rectifying} & CAG\_UDA~\cite{zhang2019category} & FADA~\cite{Haoran_2020_ECCV} \\
        \includegraphics[width=0.335\columnwidth]{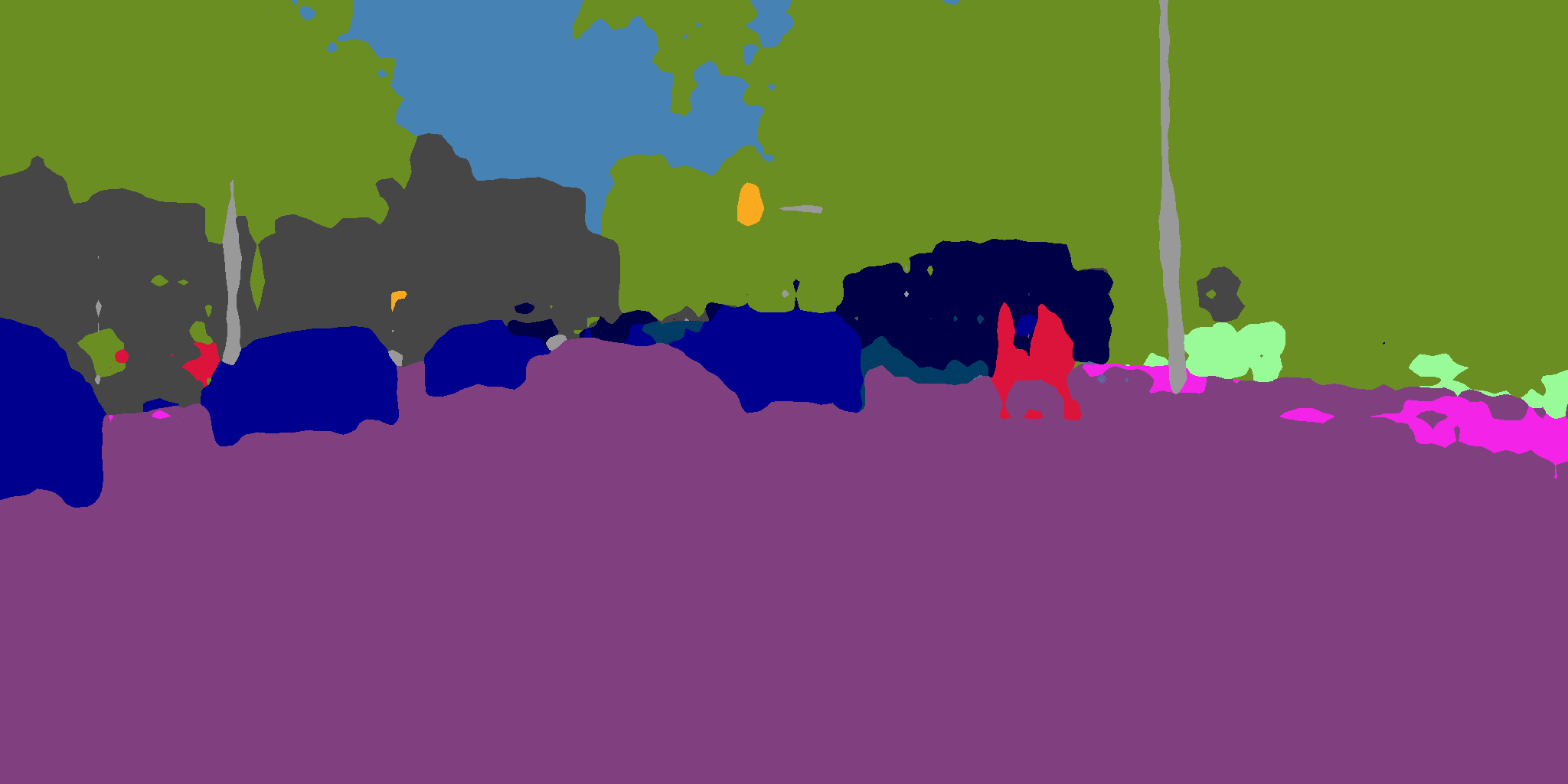}&
        \includegraphics[width=0.335\columnwidth]{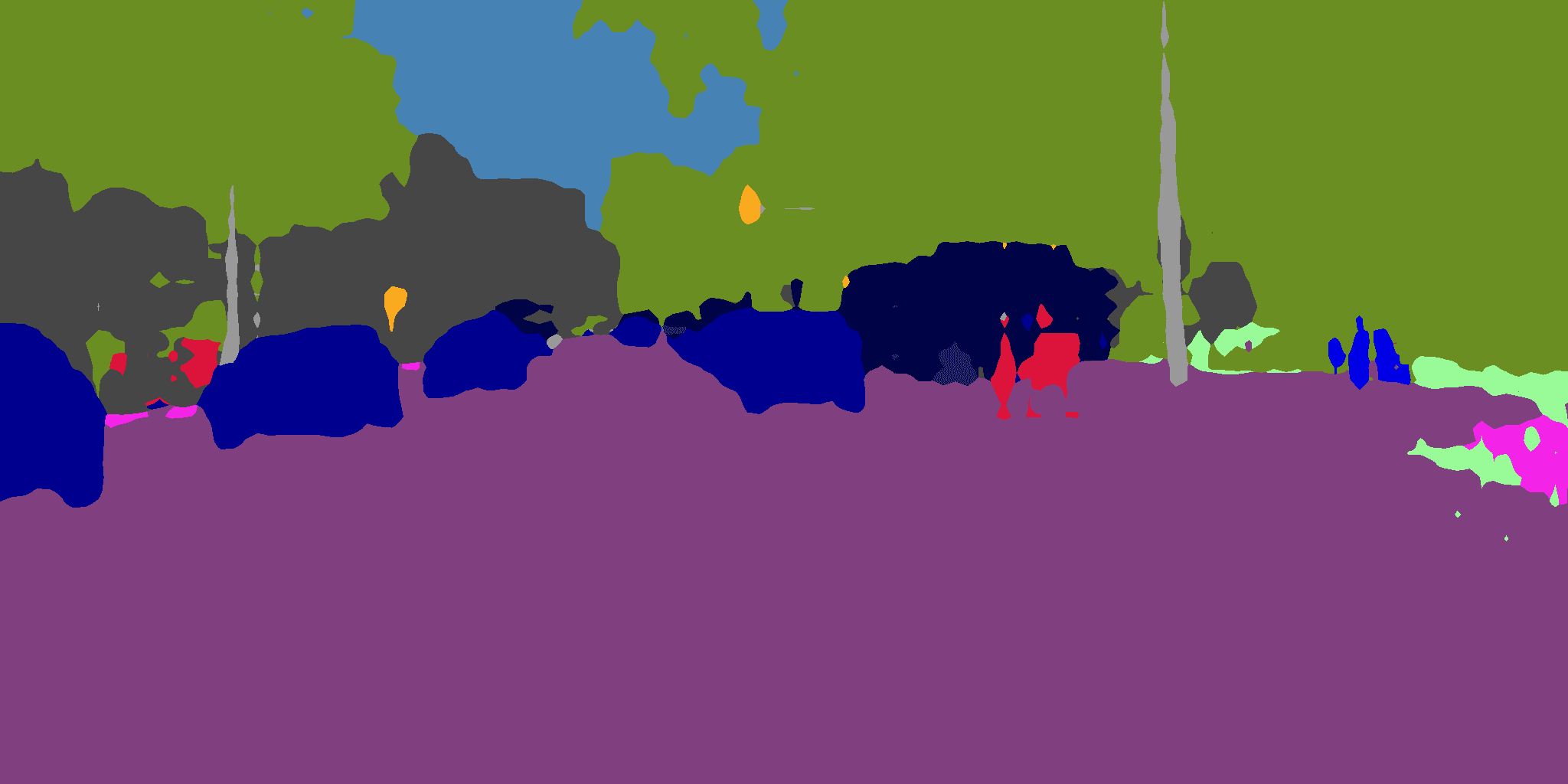}&
        \includegraphics[width=0.335\columnwidth]{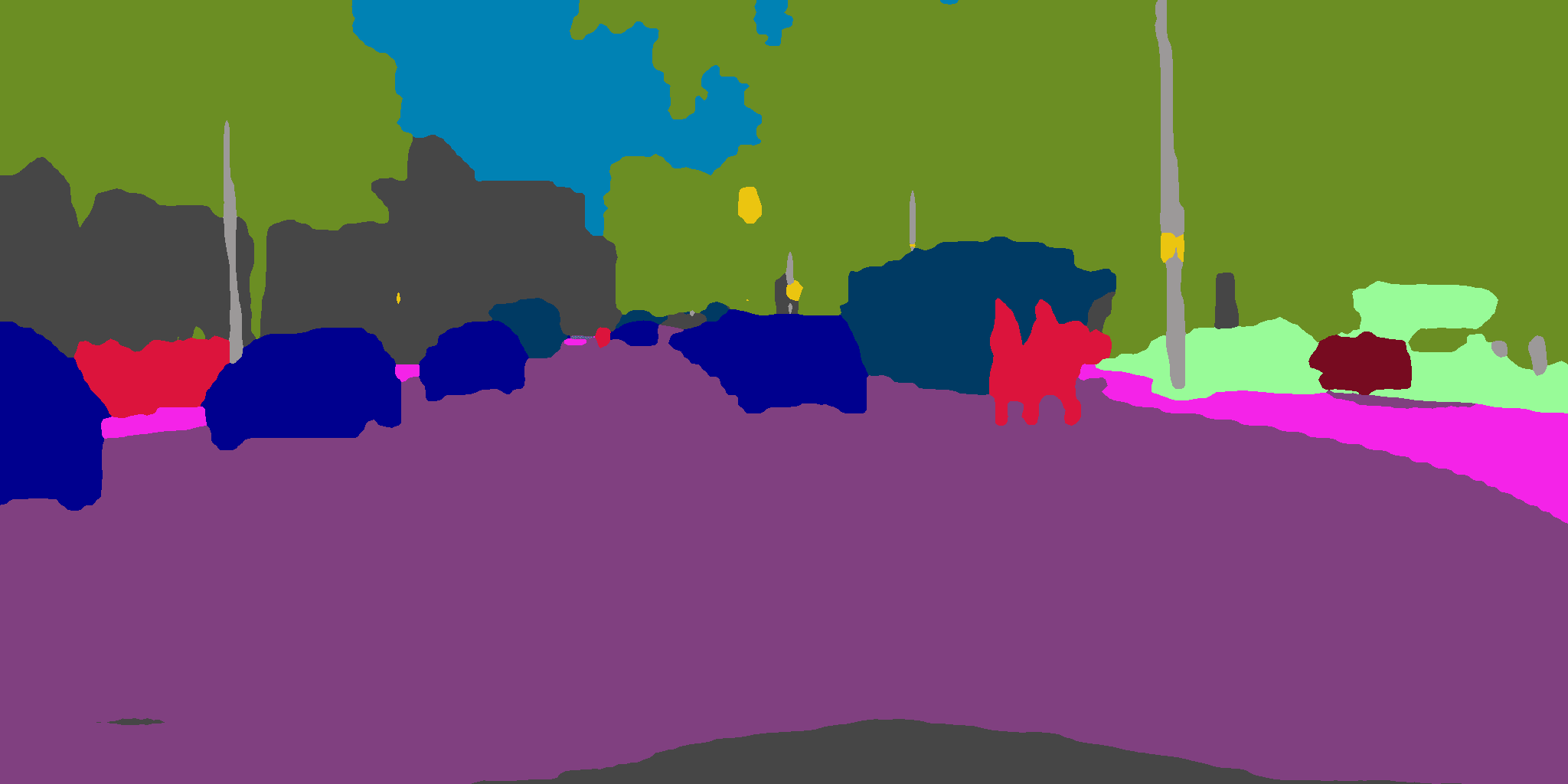}\\
        ADVENT~\cite{vu2019advent} & CLAN~\cite{luo2019taking} & ProDA \\

    \end{tabular}
    }
    \caption{Qualitative comparisons of different methods.}
    \label{figure:compare1}
    \end{figure}

\begin{figure}[t]
    \center
    \small
    \setlength\tabcolsep{1pt}
    {
    \newcolumntype{P}[1]{>{\centering\arraybackslash}p{#1}}
    \centering
    \begin{tabular}{@{}*{10}{P{0.0974\columnwidth}}@{}}
         {\cellcolor[rgb]{0.5,0.25,0.5}}\textcolor{white}{road} &{\cellcolor[rgb]{0.957,0.137,0.91}}sidewalk &{\cellcolor[rgb]{0.275,0.275,0.275}}\textcolor{white}{building} &{\cellcolor[rgb]{0.4,0.4,0.612}}\textcolor{white}{wall} &{\cellcolor[rgb]{0.745,0.6,0.6}}fence &{\cellcolor[rgb]{0.6,0.6,0.6}}pole &{\cellcolor[rgb]{0.98,0.667,0.118}}traffic light&{\cellcolor[rgb]{0.863,0.863,0}}traffic sign &{\cellcolor[rgb]{0.42,0.557,0.137}}vegetation & {\cellcolor[rgb]{0,0,0}}\textcolor{white}{n/a.}\\
         
         {\cellcolor[rgb]{0.596,0.984,0.596}}terrain &{\cellcolor[rgb]{0,0.51,0.706}}sky &{\cellcolor[rgb]{0.863,0.078,0.235}}\textcolor{white}{person} &{\cellcolor[rgb]{1,0,0}}\textcolor{white}{rider} &{\cellcolor[rgb]{0,0,0.557}}\textcolor{white}{car} &{\cellcolor[rgb]{0,0,0.275}}\textcolor{white}{truck} &{\cellcolor[rgb]{0,0.235,0.392}}\textcolor{white}{bus}&{\cellcolor[rgb]{0,0.314,0.392}}\textcolor{white}{train} &{\cellcolor[rgb]{0,0,0.902}}\textcolor{white}{motorcycle} & {\cellcolor[rgb]{0.467,0.043,0.125}}\textcolor{white}{bike}\\
    
    \end{tabular}
    
    \renewcommand{\arraystretch}{0.6}
    \begin{tabular}{@{}ccc@{}}
    %    \multicolumn{4}{c}{\includegraphics[width=2.088\columnwidth]{figure/class_color.png}}\\
    
        \includegraphics[width=0.335\columnwidth]{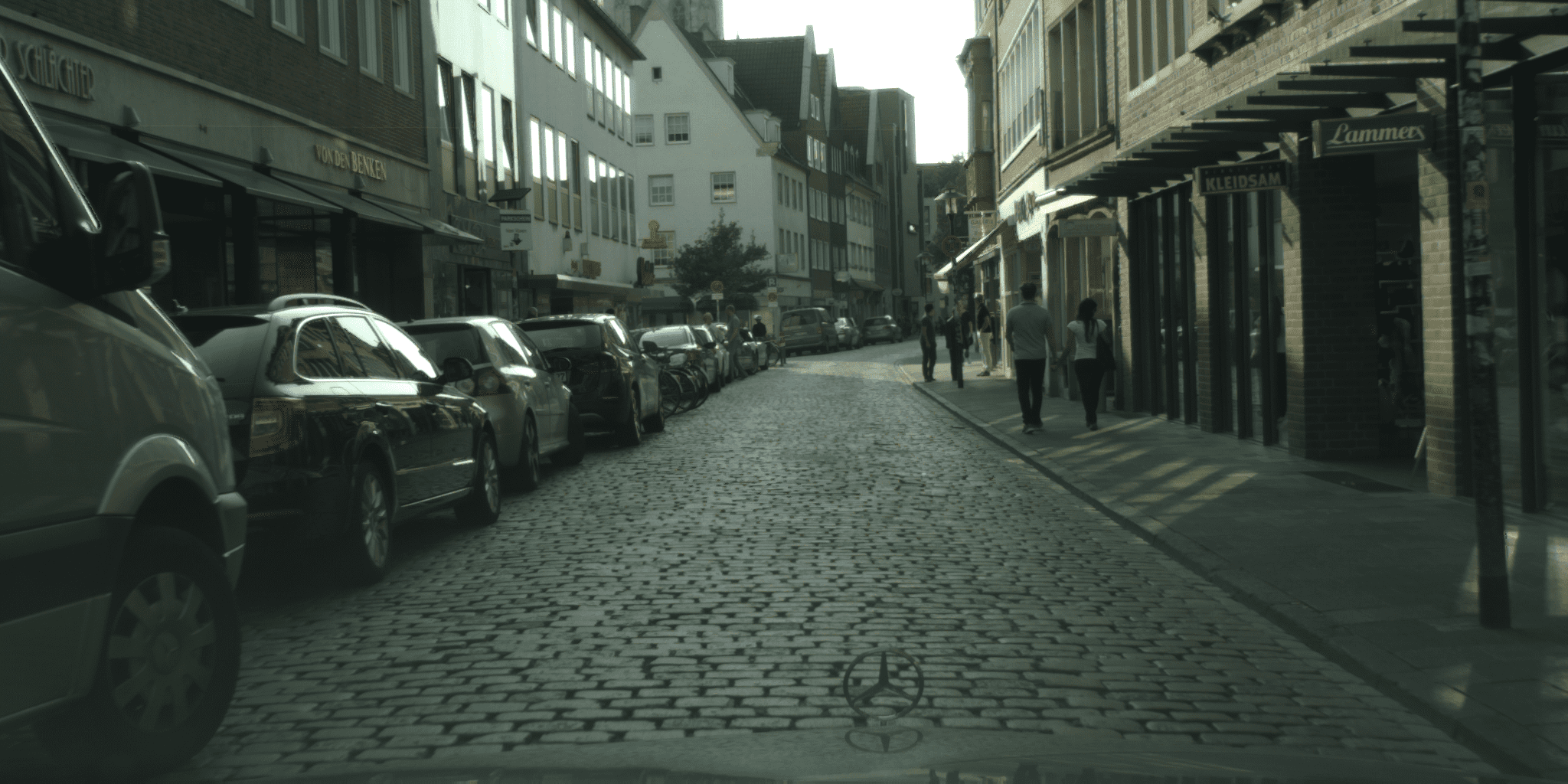}&
        \includegraphics[width=0.335\columnwidth]{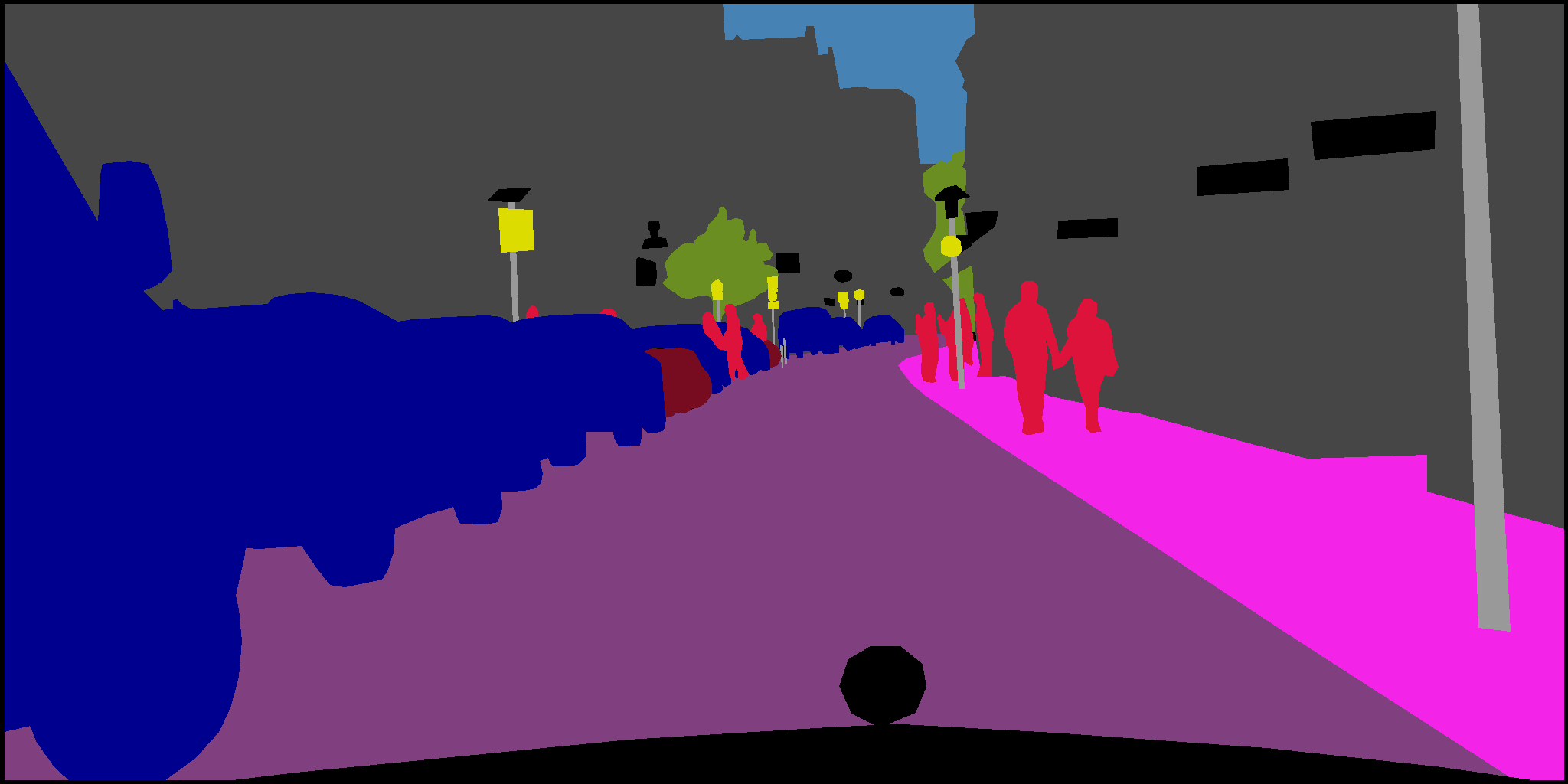}&
        \includegraphics[width=0.335\columnwidth]{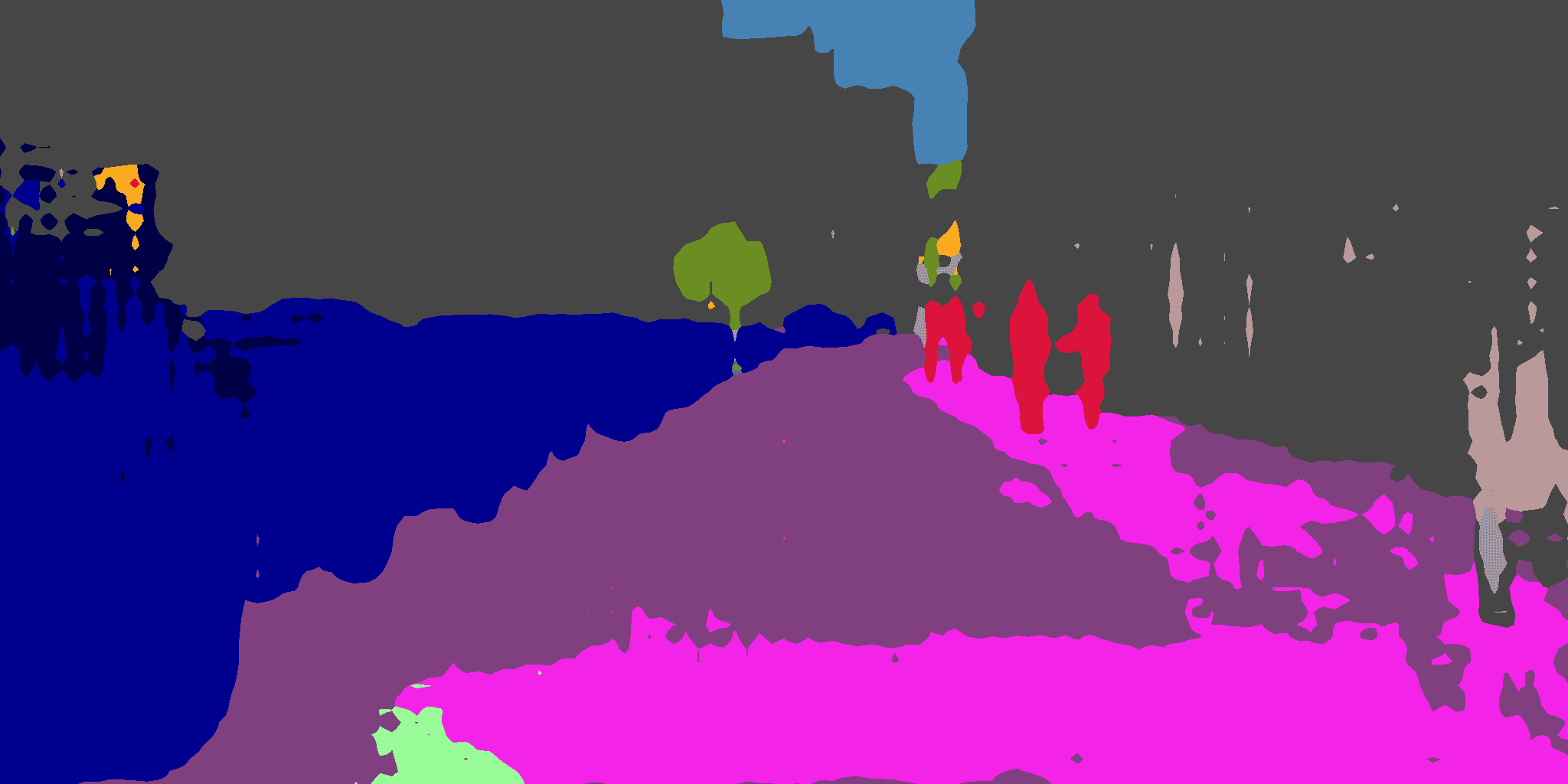}\\
        Input & Ground truth & Befor adaptation \\
        \includegraphics[width=0.335\columnwidth]{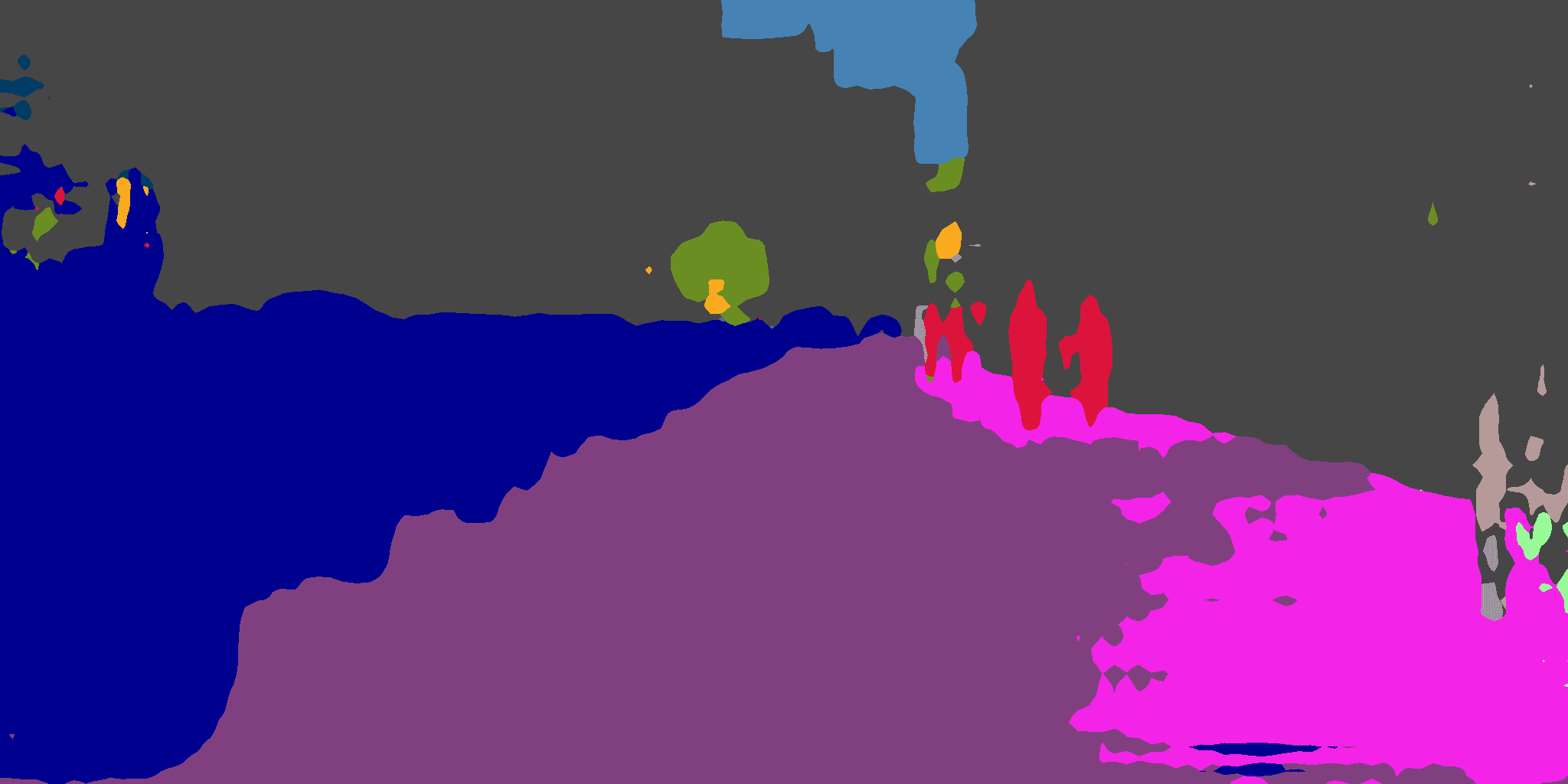}&
        \includegraphics[width=0.335\columnwidth]{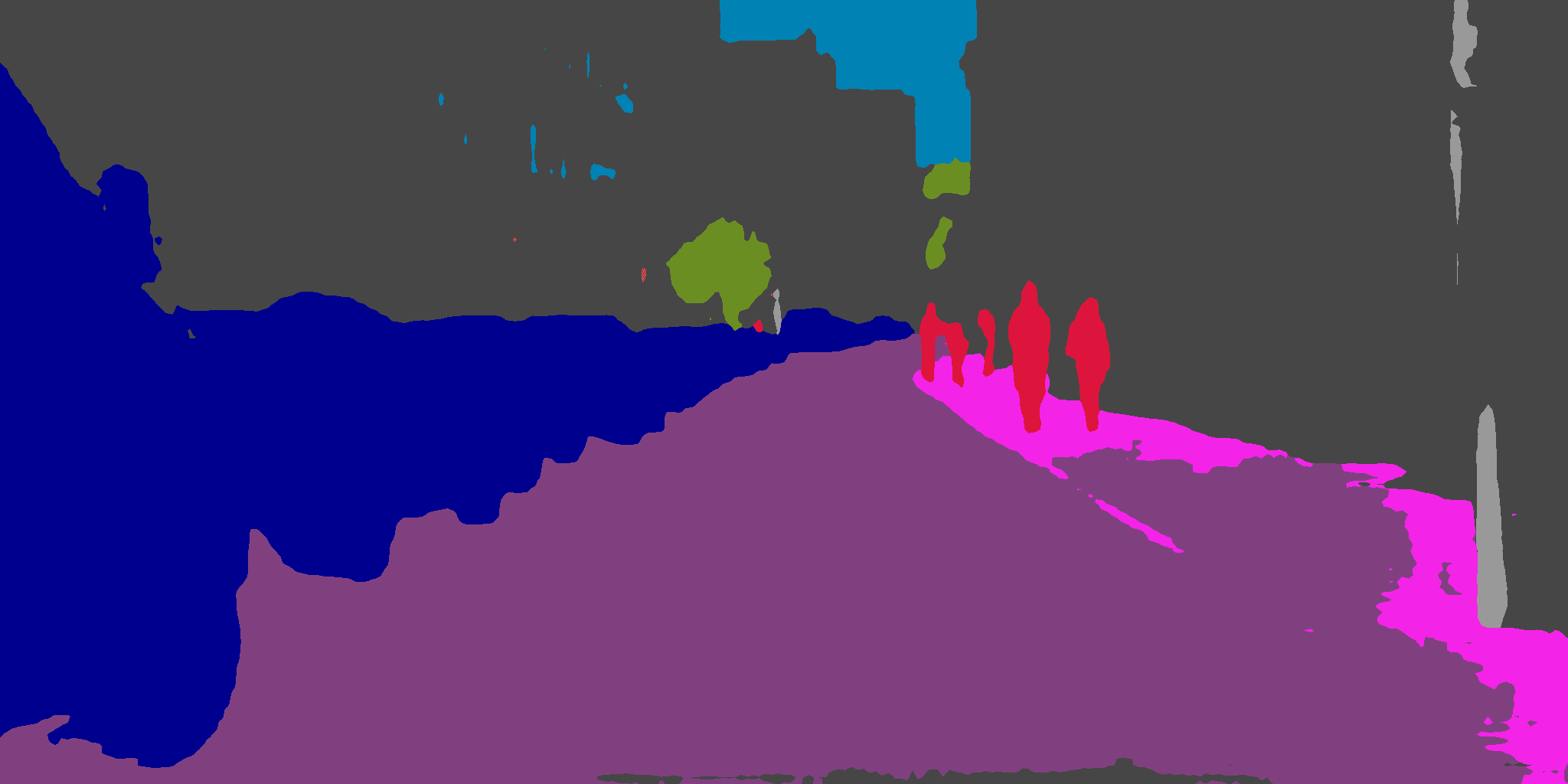}&
        \includegraphics[width=0.335\columnwidth]{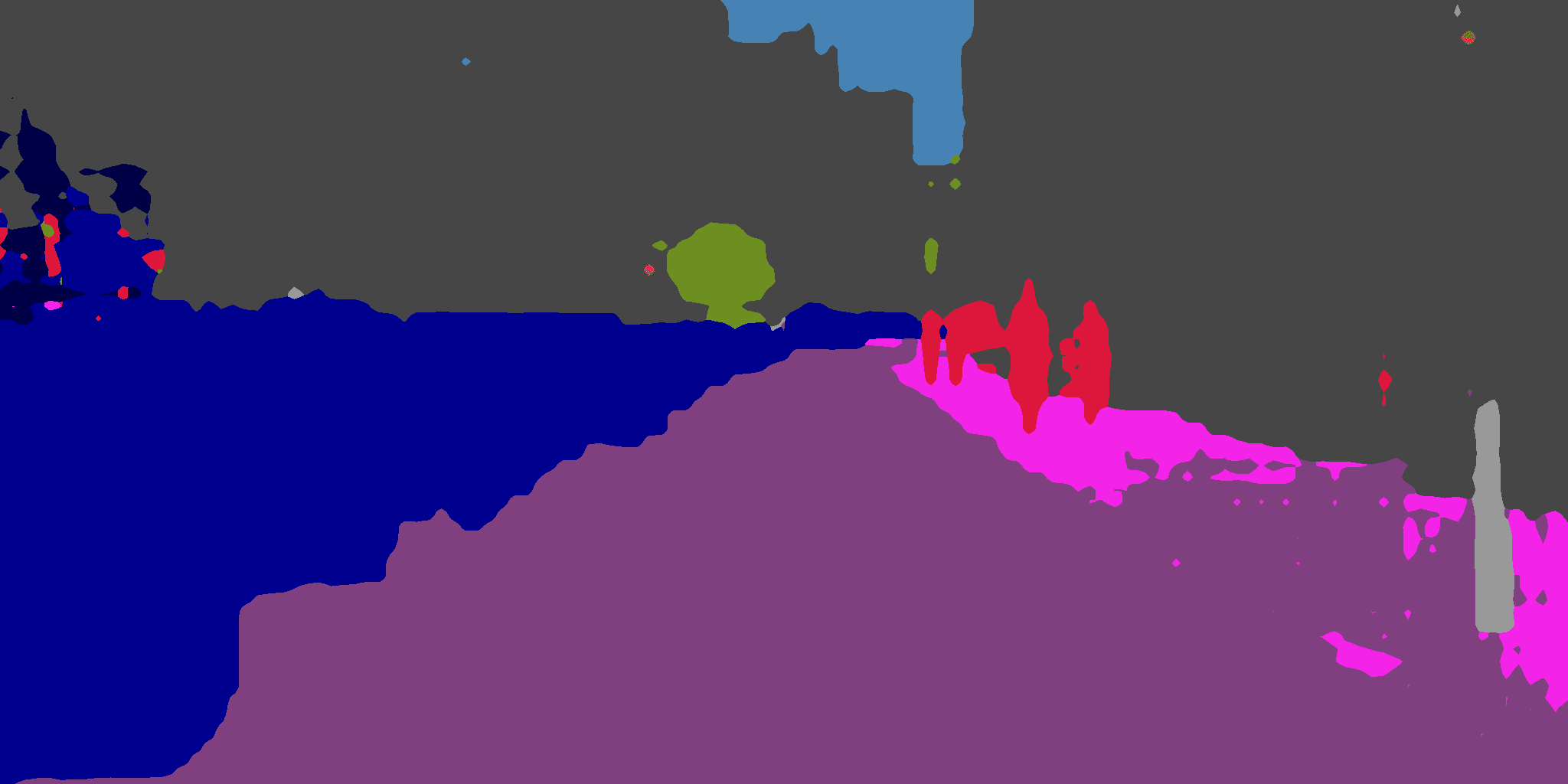}\\
        AdaptSeg~\cite{Tsai_adaptseg_2018} & CBST~\cite{zou2018unsupervised} & BDL~\cite{li2019bidirectional} \\
        \includegraphics[width=0.335\columnwidth]{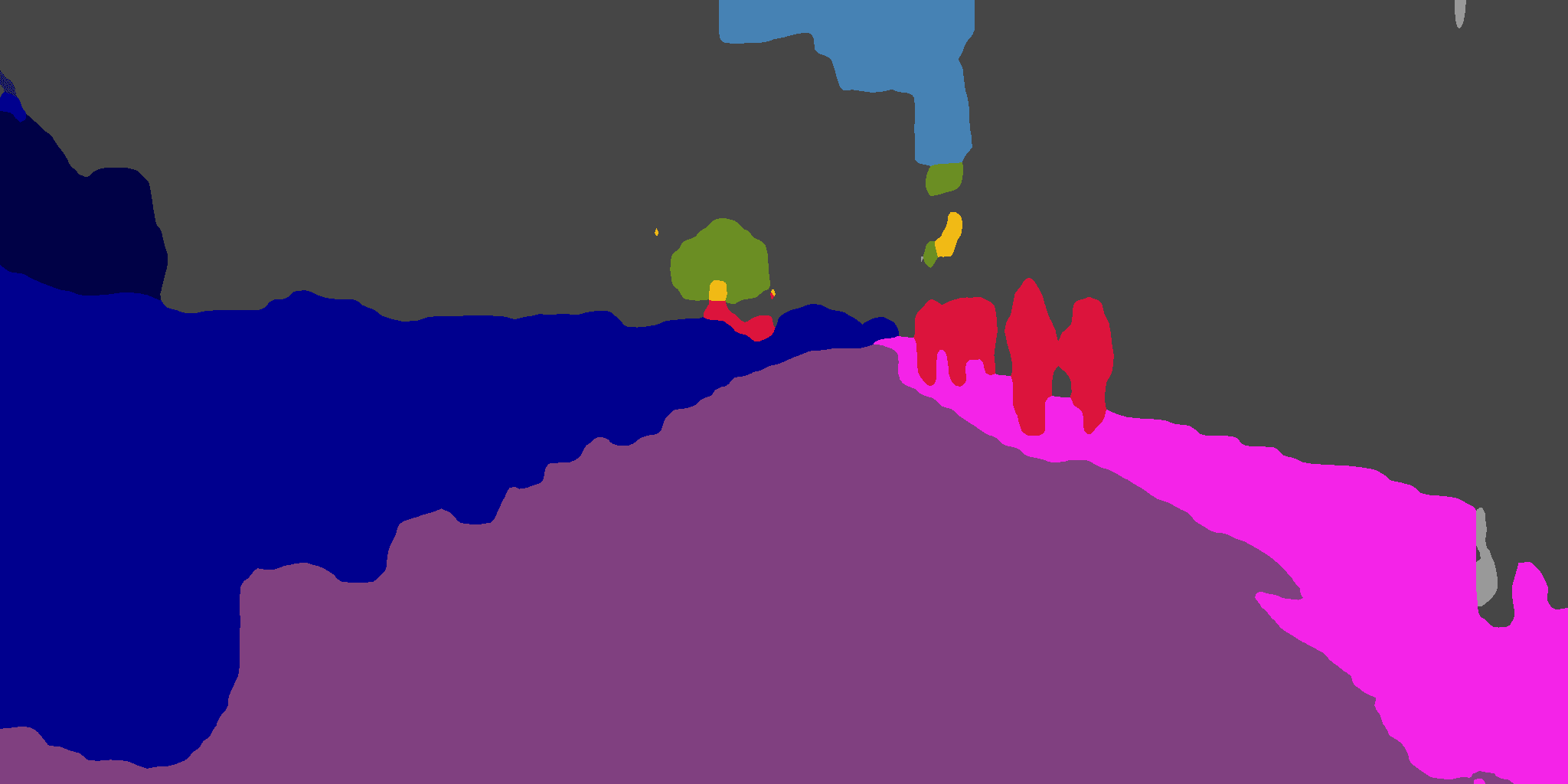}&
        \includegraphics[width=0.335\columnwidth]{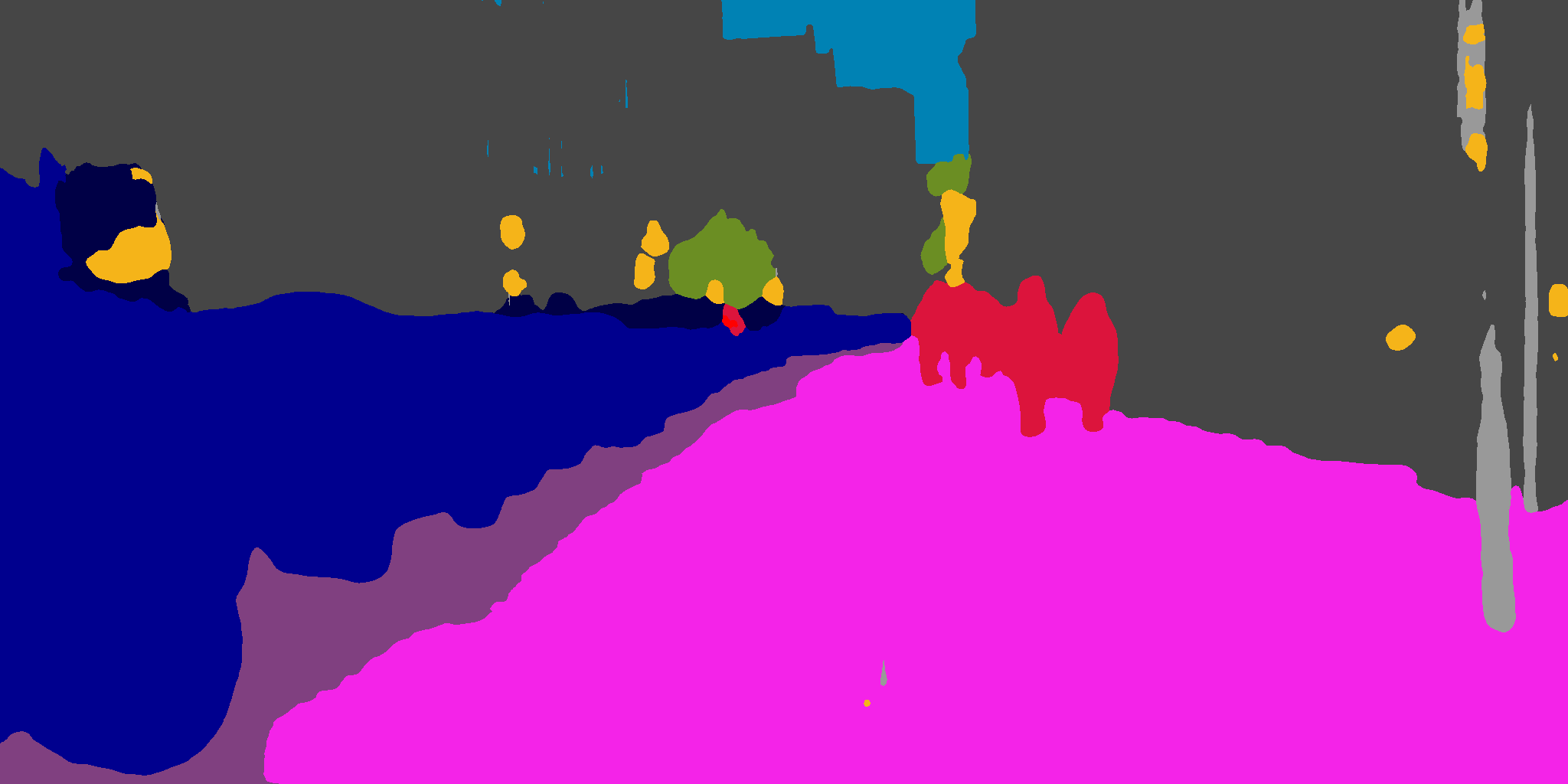}&
        \includegraphics[width=0.335\columnwidth]{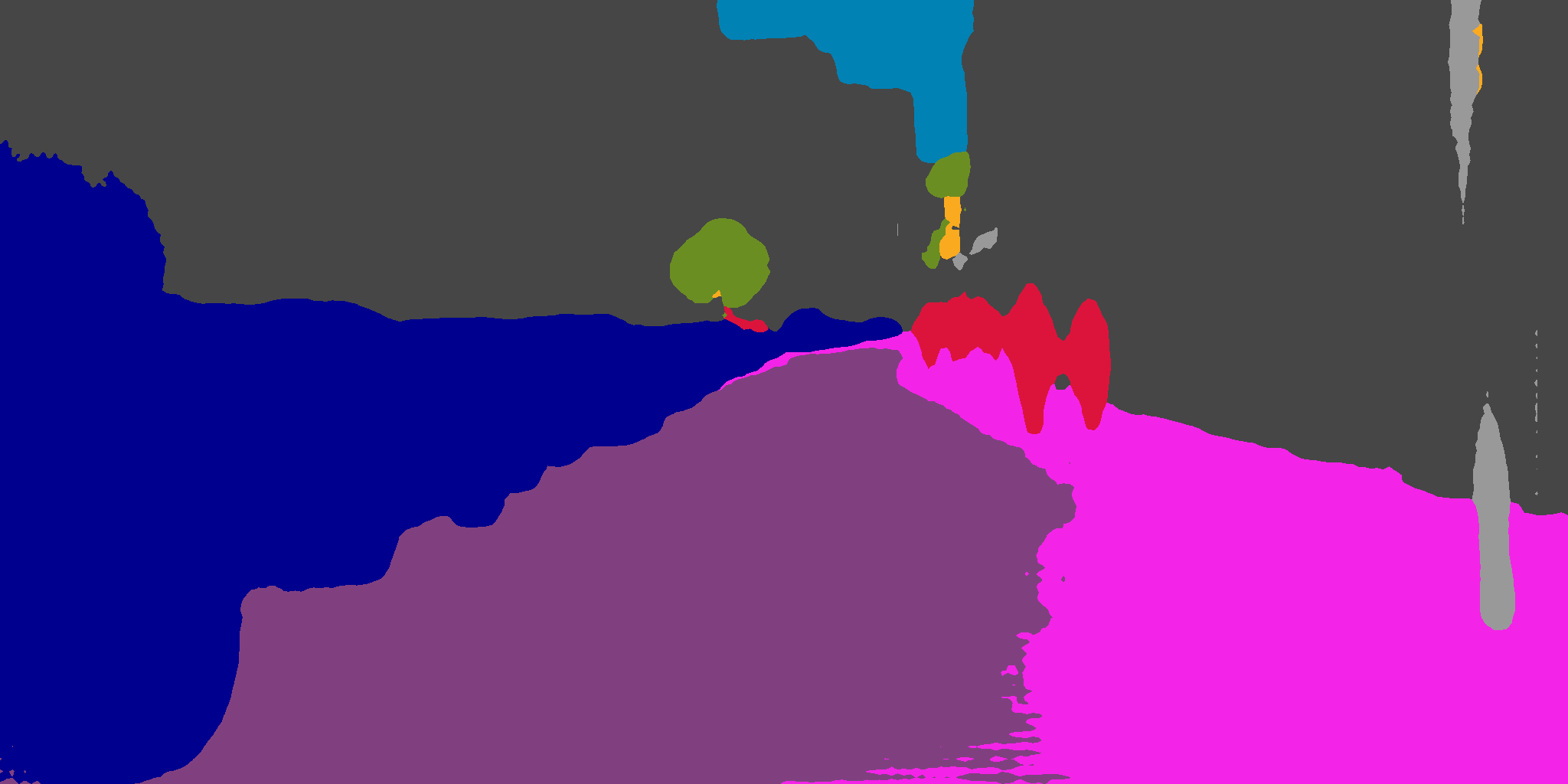}\\
        Seg-Uncertainty~\cite{zheng2020rectifying} & CAG\_UDA~\cite{zhang2019category} & FADA~\cite{Haoran_2020_ECCV} \\
        \includegraphics[width=0.335\columnwidth]{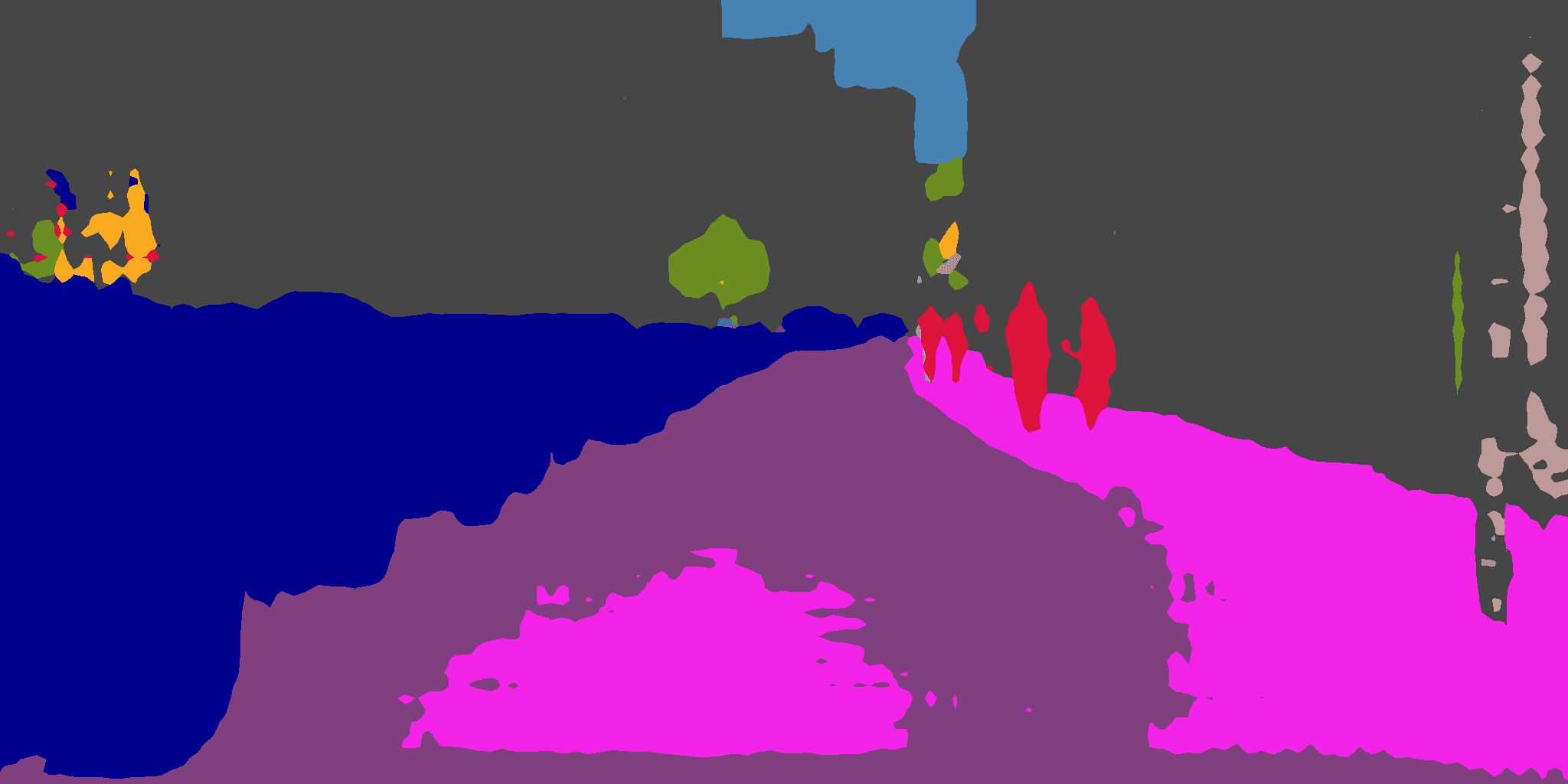}&
        \includegraphics[width=0.335\columnwidth]{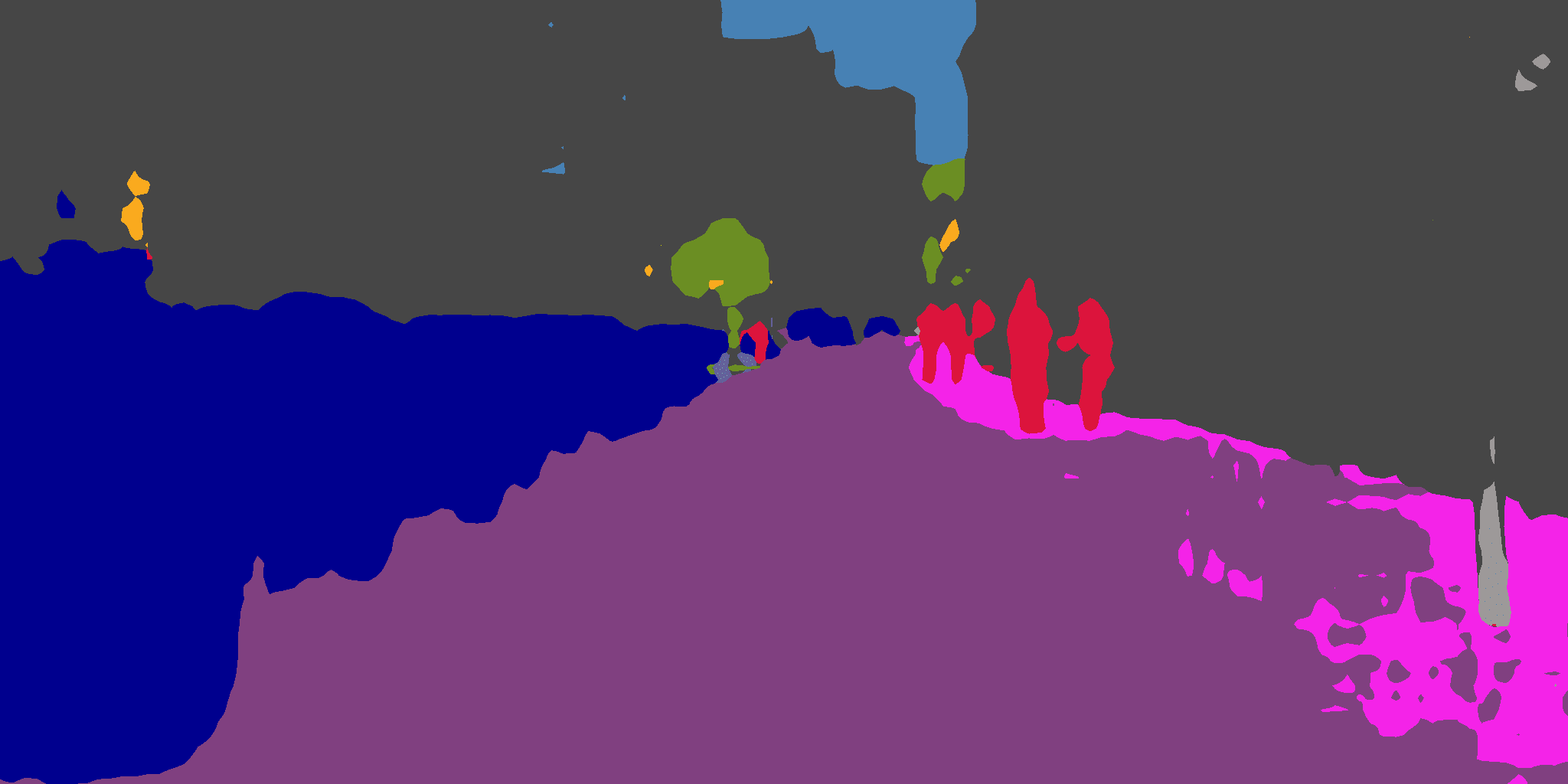}&
        \includegraphics[width=0.335\columnwidth]{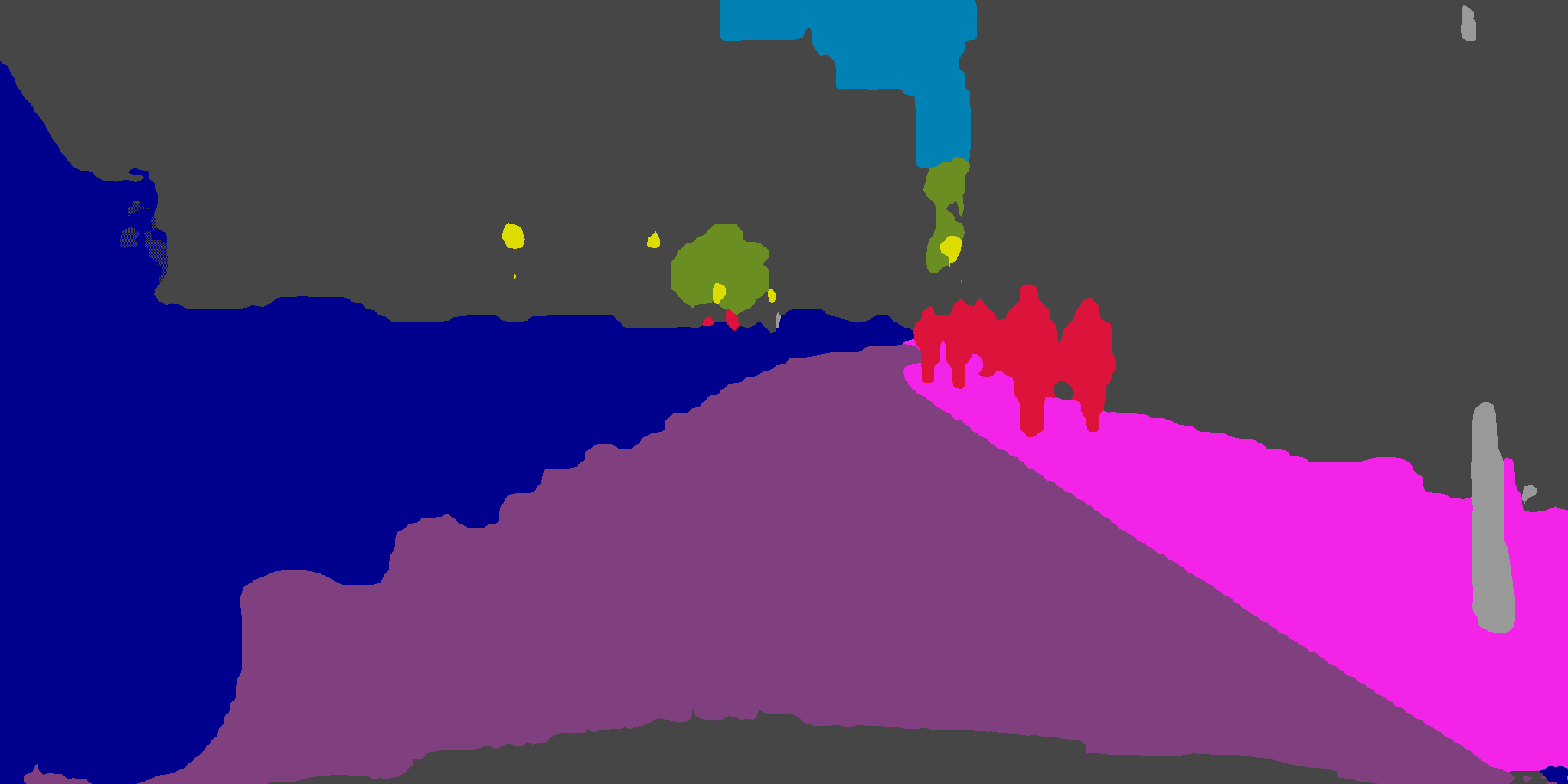}\\
        ADVENT~\cite{vu2019advent} & CLAN~\cite{luo2019taking} & ProDA \\

    \end{tabular}
    }
    \caption{Qualitative comparisons of different methods.}
    \label{figure:compare2}
    \end{figure}

\clearpage

\end{document}